\pdfoutput=1
\documentclass{article}

\usepackage[preprint]{neurips_2026}

\usepackage[utf8]{inputenc}
\usepackage[T1]{fontenc}
\usepackage{amsmath,amssymb,amsthm}
\usepackage{mathtools}
\usepackage{bm}
\usepackage{booktabs}
\usepackage{array}
\usepackage{tabularx}
\usepackage{multirow}
\usepackage{graphicx}
\usepackage{nicefrac}
\usepackage{microtype}
\usepackage{xcolor}
\usepackage{enumitem}
\usepackage{placeins}
\usepackage[colorlinks=true,citecolor=blue!55!black,linkcolor=blue!55!black,urlcolor=blue!55!black]{hyperref}
\usepackage{url}
\newtheorem{proposition}{Proposition}

\newcommand{\E}{\mathbb{E}}
\newcommand{\Cov}{\mathrm{Cov}}
\newcommand{\spanop}{\mathrm{span}}
\newcommand{\Dsub}{D_6}
\newcommand{\CD}{C_{D_6}}
\newcommand{\Eperp}{E_\perp}
\newcommand{\Dexec}{\Delta^{\mathrm{exec}}}
\newcommand{\cexec}{c^{\mathrm{exec}}}
\newcommand{\bexec}{b^{\mathrm{exec}}}
\newcommand{\mL}{m_{\mathcal{L}}}
\newcommand{\HTF}{H_{\mathrm{TF}}}
\newcommand{\boxedtok}{\texttt{\textbackslash boxed\{\}}}
\newcommand{\nothink}{\texttt{no-think}}
\newcommand{\think}{\texttt{think}}
\newcommand{\passk}[1]{\texttt{pass@#1}}
\newcommand{\cotpassk}[1]{\texttt{CoT-pass@#1}}

\setlist[itemize]{leftmargin=1.4em,itemsep=2pt,topsep=3pt}
\setlist[enumerate]{leftmargin=1.6em,itemsep=2pt,topsep=3pt}

\title{Steepest Paths First: The Initial Fisher Geometry of Reasoning RL}

\author{%
  Qihong Lin\thanks{Corresponding author.}\qquad Wenyi Fang\qquad Huabing Yan \\
  Huawei Technologies Co., Ltd. \\
  \texttt{\{linqihong1, fangwenyi3, yanhuabing\}@huawei.com}
}

\begin{document}

\maketitle

\begin{abstract}
Reinforcement learning with verifiable rewards (RLVR) often makes reasoning models both more accurate and much longer, even when longer responses are less likely to be correct within the same prompt on average. We investigate what drives this lengthening and separate latent habit release from new learning by decomposing RL weight displacements via the pre-RL trajectory Fisher information matrix into a six-dimensional leading Fisher subspace $\Dsub$ and its orthogonal complement $\Eperp$.

\textbf{(1)~Discovering a pre-RL length--accuracy knob.} From unlabeled base-model rollouts alone, we estimate the top Fisher direction $u_0$; this direction is highly concentrated, with roughly $10^{-5}$ of all parameters carrying half of its energy. A single BF16 weight shift along $u_0$, calibrated via a reference RL checkpoint, raises \texttt{Qwen3-0.6B}'s held-out MATH accuracy by \textbf{$\mathbf{13.77}$ percentage points} while mean response length expands $4.2\times$; analogous single-axis lengthening or format-tightening effects appear across the other tested model families and prompt settings.

\textbf{(2)~Separating mode release from additional learning.} Splitting 40-step GRPO weight displacements directly in BF16 shows that restoring $\Dsub$ alone reproduces \textbf{about $\mathbf{60\%}$} of the held-out accuracy gain, while adding $\Eperp$ recovers nearly all of the remainder. Mixed-precision rounding treats these two subspaces asymmetrically: because $\Dsub$ is highly concentrated whereas $\Eperp$ is diffuse, projecting out $\Dsub$ in FP32 before rounding to BF16 leaves a \textbf{reversed rounding bias} that drops ablated accuracy near the pre-RL baseline and masks the orthogonal gain.

\textbf{(3)~Tracing training dynamics from mode release to orthogonal refinement.} Early in training, cross-problem reproducible reward gradients concentrate in $\Dsub$ to release latent thinking or formatting habits in the base model (\textbf{base-model mode arbitrage}). Near step~20, these shared gradients weaken sharply alongside rising 4k truncation, and learning shifts toward diffuse updates in $\Eperp$ that improve accuracy and error recovery from corrupted prefixes (\textbf{orthogonal refinement}).
\end{abstract}

\section{Introduction}
\label{sec:intro}

\begin{figure}[t]
  \centering
  \includegraphics[width=\textwidth]{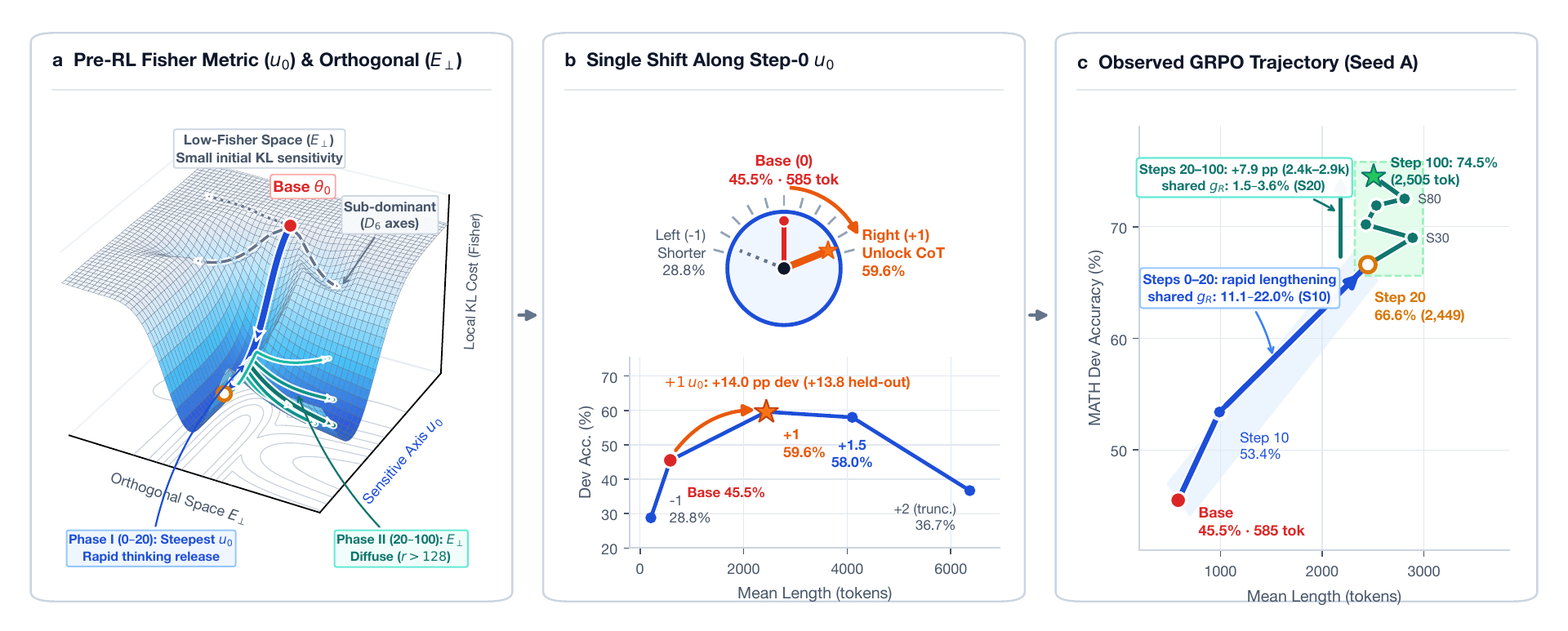}
  \caption{\textbf{Early RL gains combine pre-RL mode release along $u_0$ with post-step-20 refinement beyond lengthening.} \textbf{(a)} Local Fisher geometry separates the concentrated leading direction $u_0$ (unlocking existing output modes) from the Euclidean orthogonal complement $\Eperp$ of $\Dsub$ (carrying later refinement). \textbf{(b)} Without reward optimization, a single BF16 shift along unlabeled $u_0$ (reference-calibrated at step~40) lifts MATH dev accuracy by $14.0$ pp ($45.5\% \to 59.6\%$, plotted curve) and fixed-dose held-out accuracy by $13.77$ pp at $+1\,u_0$ before over-extension at $+2\,u_0$ triggers 8k truncation. \textbf{(c)} On Seed~A (all 8 checkpoints from Table~\ref{tab:c2}), rapid lengthening and most early accuracy growth finish by step~20, coinciding with a sharp 3-seed drop in cross-subset reproducible reward gradients ($11.1\%\text{--}22.0\% \to 1.5\%\text{--}3.6\%$; Table~\ref{tab:c4}) and rising 4k truncation; accuracy then climbs another $7.9$ pp through step~100 while mean length stays within $2.4\text{k}\text{--}2.9\text{k}$ tokens.}
  \label{fig:overview}
\end{figure}

On MATH, Group Relative Policy Optimization (GRPO) lengthens \texttt{Qwen3-0.6B}'s responses under the \nothink{} template (which inserts an empty thinking block) yet shortens them under a 32-token direct-answer prompt. In both settings, however, the within-prompt covariance between log response length and reward is negative on average; covariance sign alone therefore cannot tell whether early GRPO will lengthen or shorten outputs. Under \nothink{}, on-policy GRPO with answer-only rewards pushes mean length past 2,000 tokens by step~40 and lifts held-out accuracy by $26.4$ percentage points (pp, 3-seed mean), whereas direct-answer outputs fall below 10 tokens. Our step-0 Fisher-weighted response predicts this early length direction across all three tested settings (Table~\ref{tab:pred}). To measure how much of the early accuracy gain comes from unlocking pre-existing habits rather than learning new reasoning, we estimate a single leading Fisher direction $u_0$ from unlabeled pre-RL rollouts and calibrate its unit step to a reference step-40 RL checkpoint (\S\ref{sec:quadrants}). Without any RL updates, one BF16 shift along $u_0$ raises dev accuracy by $\mathbf{14.0}$ \textbf{pp} and held-out accuracy by $\mathbf{13.77}$ \textbf{pp} (Figure~\ref{fig:overview}b), showing that a large share of early RL gains reflects unlocking pre-existing habits.

We use the step-0 trajectory Fisher information matrix to define the local Fisher metric and decompose the trained weight displacement directly in BF16 execution precision into two complementary spatial components (Figure~\ref{fig:overview}). Inside the six-dimensional leading Fisher subspace $\Dsub$ (estimated from unlabeled pre-RL rollouts), small concentrated shifts amplify pre-existing thinking or formatting modes under the current prompt (\textbf{Base-Model Mode Arbitrage}); restoring this $\Dsub$ component ($\CD$) alone reproduces $15.7$ pp of the 40-step held-out gain and nearly all of the length growth. Adding the Euclidean orthogonal complement $\Eperp$ to $\CD$ adds another $8.3\text{--}11.7$ pp in accuracy, improves recovery from corrupted prefixes, and reduces truncation (\textbf{Orthogonal Refinement}).

Spatial splitting leaves open why rapid lengthening stalls near step~20 while accuracy keeps rising. Measuring the fraction of reward-gradient energy reproducible across disjoint problem subsets (\S\ref{sec:dynamics}) shows a sharp drop from $11.1\%\text{--}22.0\%$ at step~10 to $1.5\%\text{--}3.6\%$ at step~20, driven by a steep fall in the cross-subset numerator alongside rising 4k truncation. Shared lengthening therefore weakens sharply by step~20, separating two temporal regimes. Throughout the paper, \textbf{$\CD$ and $\Eperp$ denote spatial components} of the weight displacement at any checkpoint, whereas \textbf{Phase~I and Phase~II denote the temporal regimes} before and after this step-20 transition. We use the same decomposition to account for cross-model differences and template sensitivity (\S\ref{sec:puzzles}), out-of-domain transfer (\S\ref{sec:net-gain}), and finite-budget coverage (\S\ref{sec:coverage}).

We make three main contributions:
\begin{itemize}
  \item \textbf{Step-0 Fisher selection--response relation and bidirectional steering (\S\ref{sec:geometry}--\S\ref{sec:arbitrage}):} A Fisher-weighted selection--response relation predicts the sign of early GRPO length change across all three tested settings. Comparing the 8k chain-of-thought (CoT) and 32-token direct-answer settings inverts directional Fisher information from $8\times$ larger on the CoT axis to $60\times$ larger on the format axis, accounting for why prompt-and-budget shifts flip both training-free steering and early GRPO from lengthening to shortening. Across model families and prompts, single-axis BF16 steering accordingly falls into two regimes: lengthening reasoning or tightening answer formatting.
  \item \textbf{BF16 execution decomposition and post-step-20 dynamics (\S\ref{sec:bf16}--\S\ref{sec:dynamics}):} We prove when FP32-only projection creates a reversed BF16 rounding bias via the half-ULP dead zone (where shifts within half a BF16 grid step round back to the anchor; Proposition~\ref{prop:deadzone}), and eliminate it with iterative BF16 placement to separate $\CD$'s $15.7$ pp mode-release gain from the $8.3\text{--}11.7$ pp added by $\Eperp$ (with positive orthogonal gains replicating across external controls, \texttt{Qwen3-8B}, and \texttt{Qwen2.5-1.5B}). Between steps~10 and 20, cross-problem reproducible reward gradients drop sharply alongside rising 4k truncation even as directional Fisher curvature along $u_0$ rises; thereafter, $\Eperp$ drives standalone accuracy and error-recovery gains through sub-ULP FP32 accumulation (updates smaller than one BF16 grid step) on weights whose BF16 values have not yet shifted.
  \item \textbf{Behavioral improvements beyond mode release (\S\ref{sec:refinement}):} Beyond early lengthening and template release, RL yields two measurable improvements: stronger error-sensitive recovery from corrupted prefixes, and selective within-support probability reallocation that lifts small-$k$ hit rates and trims truncation. Targeted controls separate these gains from residual length, boilerplate or prefix-restart shortcuts, template-suppression coverage artifacts, and global sharpening (\S\ref{sec:refinement}).
\end{itemize}

\section{Setup and the Step-0 Fisher Selection--Response Relation}
\label{sec:geometry}

GRPO lengthens responses under \nothink{} yet shortens them under the 32-token direct-answer prompt, even though within-prompt length--reward covariance is negative on average in both settings. To account for these opposite early length changes, we construct the unlabeled Fisher subspace $\Dsub$ (\S\ref{sec:setup}), derive a step-0 Fisher-weighted selection--response relation, and test it via Fisher-metric interventions (\S\ref{sec:selection-response}).

\subsection{Experimental Protocol and the Unsupervised Subspace \texorpdfstring{$\Dsub$}{D6}}
\label{sec:setup}

\paragraph{Primary setting and mixed-precision convention.}
In our primary setup, we train instruction-tuned \texttt{Qwen3-0.6B} \citep{yang2025qwen3} on MATH \citep{hendrycks2021math} under the \nothink{} template \citep{ma2025without}. Optimization uses on-policy GRPO in its single-step REINFORCE form, without a Kullback--Leibler (KL) penalty or Proximal Policy Optimization (PPO) ratio clipping \citep{shao2024deepseekmath}. AdamW \citep{loshchilov2019adamw} maintains FP32 master weights $\theta$, which we round element-wise to nearest BF16 to obtain the \textbf{BF16 execution weights} $Q(\theta)$ used for all rollouts and evaluations. We partition MATH by problem-statement hash into four disjoint splits: training (6,400 problems), development (1,024), held-out (512), and lockbox (1,024; Table~\ref{tab:b2}). GRPO training rollouts (and gradient diagnostics) use a 4,096-token cap, whereas standard MATH evaluations use an 8,192-token cap. We use dev only to estimate unlabeled geometry and run diagnostics, held-out for our primary decompositions and behavioral evaluations, and lockbox strictly for late-stage blind synchronization checks (\S\ref{sec:unified}; apart from 128 problems used only for unsupervised \texttt{R1-Distill} axis extraction, Table~\ref{tab:b2}); none of the evaluation splits inform gradient updates or hyperparameter selection. Appendix~\ref{app:b1} details the full notation, dataset splits, and cross-model configurations (Tables~\ref{tab:b1}--\ref{tab:b2}). All accuracy differences are reported in percentage points (pp).

\paragraph{Step-0 trajectory Fisher information matrix and $\Dsub$.}
Let $\pi_{\theta_0}(y \mid x)$ denote the pre-RL base policy and $G(x, y) = \nabla_\theta \log \pi_{\theta_0}(y \mid x)$ its trajectory score on self-sampled outputs $y \sim \pi_{\theta_0}(\cdot \mid x)$. We define the step-0 \textbf{trajectory Fisher information matrix} from model rollouts rather than ground-truth labels \citep{kunstner2019empirical,martens2020new}:
\begin{equation}
  F_0 = \E_{x,\, y \sim \pi_{\theta_0}(\cdot \mid x)}\!\left[G(x, y)\,G(x, y)^\top\right],
  \qquad F_0 v_k = \mu_k v_k,
  \qquad \mu_0 \ge \mu_1 \ge \dots \ge 0.
  \label{eq:fisher}
\end{equation}
The matrix $F_0$ induces the local \textbf{Fisher metric} on parameter space: a small step $\varepsilon$ along a unit vector $v$ changes prompt-averaged policy KL divergence by $\tfrac{1}{2}\varepsilon^2 v^\top F_0 v$ to second order. Thus, $v^\top F_0 v$ is the directional Fisher information (directional curvature of the policy KL) along $v$, and each eigenvalue $\mu_k$ measures policy sensitivity along $v_k$. We partition the dev prompts into two disjoint halves and run Lanczos iteration \citep{lanczos1950iteration,ghorbani2019hessian} on unlabeled pre-RL rollouts from each half. Taking the top three Lanczos Ritz vectors from each half and applying Gram--Schmidt orthogonalization yields an orthonormal basis $U = [u_0, \dots, u_5]$ ($U^\top U = I_6$) spanning the six-dimensional Fisher subspace $\Dsub = \spanop(U)$ (Appendix~\ref{app:b1}). Even without answers or rewards, this subspace exhibits two key properties:
\begin{enumerate}
  \item \textbf{Cross-prompt stability and similar output effects:} Leading Ritz vectors (Lanczos eigenvector approximations) estimated independently on disjoint halves of dev prompts produce nearly identical shifts in token log-probabilities (cosine similarities of $0.98\text{--}0.99$) and retain high parameter-space inner products ($0.57\text{--}0.73$, compared to near-zero for random vectors in $5.96 \times 10^8$ dimensions; Appendix~\ref{app:b2}). Moving along $u_0$ raises transition and reflection token probabilities while lowering formatting tokens, \boxedtok{}, and the end-of-sequence token (\texttt{EOS}), showing that $u_0$ captures a problem-general behavioral mode rather than prompt-specific noise. In a separate normalization check, scaling per-sample scores by $1/\sqrt{L}$ or $1/L$ recovers essentially the same leading subspace on \texttt{Qwen3-0.6B}, ruling out sequence-length weighting as an artifact (principal cosines and outlier audits in Appendix~\ref{app:b2}).
  \item \textbf{Extreme coordinate concentration:} Roughly $10^{-5}$ of all parameters, predominantly in shallow MLP layers, carry half of the squared $\ell_2$ energy of $u_0$. This extreme sparsity causes a concentrated $\Dsub$ subtraction and a diffuse RL update to round differently in BF16 (\S\ref{sec:bf16}).
\end{enumerate}

\subsection{Selection--Response Relation and Causal Interventions}
\label{sec:selection-response}

Let $z(x,y)$ be a $\theta$-independent behavioral statistic such as log-length ($\log L$), with within-prompt centered version $\tilde{z} = z - \E[z \mid x]$, and let $R$ be reward (centered $\tilde{R} = R - \E[R \mid x]$). Denote the policy gradients of expected reward and expected $z$ at $\theta_0$ by $g_R = \nabla_\theta \E[R]$ and $g_z = \nabla_\theta \E[z]$, respectively. Write $z_k$ and $r_k$ for the projections of $\tilde{z}$ and $\tilde{R}$ onto the normalized score basis $\phi_k = v_k^\top G / \sqrt{\mu_k}$, with orthogonal residuals $z_{\mathrm{res}}$ and $R_{\mathrm{res}}$. Under Euclidean policy-gradient flow, the Fisher eigenvalues weight each mode unequally (proof in Appendix~\ref{app:a1}):

\vspace{3pt}
\begin{proposition}[Step-0 Fisher Selection--Response Relation]
\label{prop:sr}
Under ideal on-policy policy-gradient flow $\mathrm{d}\theta / \mathrm{d}\tau = g_R$, the within-prompt behavioral--reward covariance and the initial behavioral response at $\theta_0$ satisfy
\begin{equation}
  \E_x[\Cov(z, R \mid x)] = \sum_{\mu_k > 0} z_k r_k + \langle z_{\mathrm{res}}, R_{\mathrm{res}} \rangle,
  \qquad
  \left.\frac{\mathrm{d}\E[z]}{\mathrm{d}\tau}\right|_{\theta_0} = g_z^\top g_R = \sum_{\mu_k > 0} \mu_k z_k r_k.
  \label{eq:sr}
\end{equation}
\end{proposition}

Equation~\eqref{eq:sr} mirrors Lande's selection--response equation $\Delta \bar{z} = \mathbf{G}\boldsymbol{\beta}$ in quantitative genetics \citep{lande1979quantitative,lande1983measurement}. Sample covariance weights all score modes equally and also includes the score-orthogonal residual $\langle z_{\mathrm{res}}, R_{\mathrm{res}} \rangle$, whereas ideal Euclidean policy-gradient flow weights each score mode by $\mu_k$; the two quantities can therefore take opposite signs whenever leading high-$\mu_k$ modes have positive $z_k r_k$ while small-eigenvalue modes and residuals are negative overall. Empirically, $\mu_0$ is $10\text{--}100\times$ larger than bulk eigenvalues, and in our lengthening settings the positive leading contribution along $u_0$ outweighs negative off-axis terms, predicting lengthening despite negative covariance (consistent with the opposite-sign $\Dsub$ and $\Eperp$ terms in Table~\ref{tab:b3b}).

To test whether step-0 geometry predicts early length trajectories across tasks, we evaluate Equation~\eqref{eq:sr} on the pre-RL checkpoint alone across three distinct settings (Table~\ref{tab:pred}). On MATH \nothink{} and Knights \& Knaves \citep{xie2024memorization,xie2025logicrl}, Equation~\eqref{eq:sr} predicts lengthening despite negative within-prompt covariance, with the dominant axis in $\Dsub$ ($u_0$ on MATH \nothink{}; $u_1 \in \Dsub$ on Knights \& Knaves) supplying $95\%$ or more of the positive response; under the 32-token direct-answer prompt, it predicts shortening, and GRPO outputs indeed fall to $8.0$ tokens by step~5. The step-0 Fisher-weighted response predicts the sign of early length change in all three tested settings, matching the direction of the observed multi-step GRPO trajectories.

\begin{table}[t]
  \centering
  \caption{First-order step-0 response of mean log response length ($\mathrm{d}\E[\log L]/\mathrm{d}\tau$) evaluated on the pre-RL checkpoint versus observed early multi-step GRPO mean token counts (allowing single-step BF16 replay to differ, Table~\ref{tab:b3b}). Values separated by a slash are computed on disjoint dev halves. Dominant-axis shares above $100\%$ reflect opposing small-eigenvalue modes; the dominant axis is $u_0$ on MATH \nothink{}, $u_1 \in \Dsub$ on Knights \& Knaves, and the dedicated format axis on MATH Direct.}
  \label{tab:pred}
  \footnotesize
  \setlength{\tabcolsep}{4pt}
  \renewcommand{\arraystretch}{1.15}
  \begin{tabularx}{\textwidth}{@{}>{\raggedright\arraybackslash}p{3.45cm}ccc>{\raggedright\arraybackslash}X@{}}
    \toprule
    Setting \& Task & Within-Prompt Cov & Step-0 $\log L$ & Dominant Share & Observed Early GRPO (Mean Tok) \\
    \midrule
    \texttt{Qwen3-0.6B}, MATH \nothink{} & $-0.0042$ & Longer ($+$) & $95\%$ ($u_0$) & Step~10: $585 \to 858\text{--}1{,}044$ (mean 964) \\
    \texttt{Qwen3-0.6B}, Knights \& Knaves & $-0.0026$ / $-0.0056$ & Longer ($+$) & $103\text{--}104\%$ ($u_1$) & Length grows $3.2\times$ \\
    \texttt{Qwen3-0.6B}, MATH Direct (32~tok) & $-0.0070$ / $-0.0063$ & Shorter ($-$) & $96\text{--}97\%$ (format) & Step~5: $28.8 \to 8.0$ (train batch) \\
    \bottomrule
  \end{tabularx}
\end{table}

Practical GRPO adds group standardization, token-mean averaging, AdamW preconditioning, and BF16 rounding to the ideal Euclidean flow of Proposition~\ref{prop:sr}. Reconstructing Fully Sharded Data Parallel (FSDP) optimizer states (Appendix~\ref{app:a1}, Table~\ref{tab:c6}) and replaying exact step-1 updates (Appendix~\ref{app:b2}, Table~\ref{tab:b3b}) explains why step-0 geometry still governs early multi-step trajectories: on step~1 alone, $\Dsub$ and $\Eperp$ already push length in opposite directions, and even the raw GRPO gradient has a slightly negative net length response before AdamW and BF16 further compress $\Dsub$'s relative single-step weight (Table~\ref{tab:b3b}). Yet in the executed BF16 step, $\Dsub$ generates $+1.95\times 10^{-5}$ (more than half the magnitude of $\Eperp$'s $-3.69\times 10^{-5}$) from only $0.0115\%$ of the squared execution norm, roughly $4.6\times 10^3\times$ stronger length leverage per unit squared norm than $\Eperp$. Over steps~1--20, because forward length sensitivity $g_z$ retains its $\sqrt{\mu_k}$ scaling regardless of optimizer preconditioning, coherent $\Dsub$ updates add constructively across steps while diffuse $\Eperp$ length terms do not align consistently (Table~\ref{tab:c6}).

\begin{figure}[t]
  \centering
  \includegraphics[width=\textwidth]{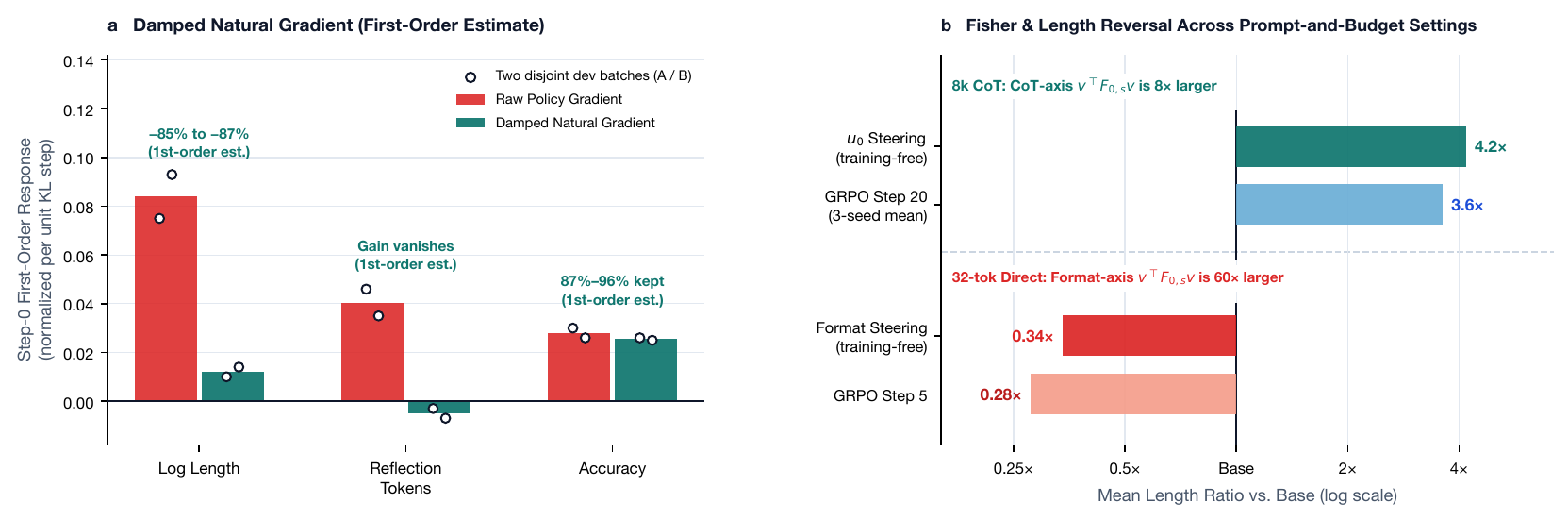}
  \caption{\textbf{Pre-RL Fisher weighting explains both why standard gradients favor lengthening and why prompt-and-budget shifts invert the length trend.} \textbf{(a)} Standard policy gradients amplify length growth via the leading eigenvalue $\mu_0$; dampening top Fisher modes with $(F_0 + \lambda I)^{-1} g_R$ at matched local KL removes $85\%\text{--}87\%$ of predicted log-length growth and eliminates reflection-token growth while retaining $87\%\text{--}96\%$ of the predicted accuracy gain (disjoint dev halves; Table~\ref{tab:b3}). \textbf{(b)} Switching from 8k CoT to 32-token direct answering flips dominant directional Fisher information ($v^\top F_{0,s} v$) from $8\times$ larger on the CoT axis to $60\times$ larger on the format axis, matching the sign reversal in both steering and early GRPO (Table~\ref{tab:b6}).}
  \label{fig:causal}
\end{figure}

\paragraph{Fisher-metric interventions (Figure~\ref{fig:causal}).}
Two interventions test the role of pre-RL Fisher weighting. First, to test whether early lengthening is required for reward gain or is instead amplified by large leading eigenvalues, we replace the step-0 gradient with the damped natural gradient $v_\lambda = (F_0 + \lambda I)^{-1} g_R$ \citep{kakade2001natural,martens2015kfac} at matched local KL step size (Appendix~\ref{app:a1}). Reducing the influence of large Fisher eigenvalues removes $85\%\text{--}87\%$ of the predicted first-order log-length response and eliminates reflection-token growth while retaining $87\%\text{--}96\%$ of the predicted first-order accuracy gain (Figure~\ref{fig:causal}a, Table~\ref{tab:b3}), showing that early lengthening is heavily amplified by leading Fisher modes rather than strictly required for local reward gain. Extending $\Dsub$ damping to online GRPO exposes the practical trade-off under \nothink{} suppression: mild damping trims length with little accuracy loss, whereas strong damping cuts length sharply at a substantial accuracy cost because blocking $\Dsub$ online forfeits early mode-release gains (Table~\ref{tab:d2}).

Second, to understand why the same pre-RL checkpoint reverses its length trajectory under the 32-token direct-answer prompt (where prompt wording and token cap change together), we compare directional Fisher information ($v^\top F_{0,s} v$, with $s$ indexing the setting) on the CoT and format axes across both settings. Under 8k CoT, directional Fisher information is $8\times$ larger on the CoT axis than on the format axis; in the 32-token direct-answer setting, it is $60\times$ larger on the format axis, a $480\times$ reversal (Table~\ref{tab:b6}). Under this combined prompt-and-budget shift, early answer formatting becomes far more KL-sensitive than extended derivation, so the dominant sensitive axis switches from CoT length to formatting, explaining why both training-free steering and early GRPO invert from lengthening to shortening (Figure~\ref{fig:causal}b).

\section{Pre-RL Mode Arbitrage: Steering Regimes and Token Anatomy}
\label{sec:arbitrage}

Using the unlabeled pre-RL leading direction $u_0$, we next test how much accuracy improves from a single BF16 weight shift along $u_0$ (\textbf{single-axis steering}). Across model families and prompts, this shift releases whichever latent habit carries the largest Fisher sensitivity under the current prompt, either expanding pre-answer derivation or tightening answer formatting (\S\ref{sec:quadrants}); inspecting individual $\Dsub$ axes and token timing then explains why weight steering succeeds where decoding-time \texttt{EOS} suppression fails (\S\ref{sec:anatomy}), and clarifies two active RLVR debates: the Qwen--Llama gap and template sensitivity in R1-Zero-style RL (\S\ref{sec:puzzles}).

\subsection{Two Steering Regimes Across Models and Prompts}
\label{sec:quadrants}

To test whether single-axis steering generalizes beyond one model or always improves accuracy by lengthening outputs, we sweep BF16 weight shifts in both directions along each setting's step-0 axis across five model--prompt pairs (Figure~\ref{fig:quadrants}, Table~\ref{tab:quadrants}; full dose curves and 8B length-normalized Lanczos details in Appendix~\ref{app:b2}, Table~\ref{tab:b7}, Figure~\ref{fig:b2}). Every main and format axis is oriented so that a positive dose $\alpha > 0$ lengthens responses. On \texttt{Qwen3-0.6B}, the unit dose $\alpha = 1$ is calibrated to Seed~A's step-40 BF16 $u_0$ displacement ($s = 0.0248$; per-seed coordinates in Appendix~\ref{app:b1}), so the direction is unsupervised while this unit step is reference-calibrated. Across the five settings, accuracy improves in two opposite regimes, determined by which latent habit sits on the top Fisher axis:

\begin{itemize}
  \item \textbf{Longer reasoning \& higher accuracy:} When the prompt suppresses latent long-CoT habits, positive steering unlocks pre-answer derivation and raises accuracy across scales and families (Table~\ref{tab:quadrants}). On \texttt{Qwen3-0.6B}, $+1\,u_0$ raises dev accuracy by $\mathbf{14.0}$ \textbf{pp} ($45.5\% \to 59.6\%$) and held-out accuracy at the same fixed dose by $\mathbf{13.77}$ \textbf{pp} ($4.2\times$ length), before over-steering hits the 8k cap and lowers accuracy through truncation (Table~\ref{tab:b12}). Replicating on larger and distilled models confirms that mode release is not confined to small models: steering adds $\mathbf{5.7}$ \textbf{pp} on \texttt{Qwen3-8B} with low truncation and $\mathbf{8.35}$ \textbf{pp} on \texttt{DeepSeek-R1-Distill-Llama-8B} under \nothink{} as reflection tokens increase (Table~\ref{tab:quadrants}). Matched-norm random directions shift accuracy by only ${\sim}1$ pp, ruling out generic perturbation effects.
  \item \textbf{Shorter outputs \& tighter formatting:} Where long-CoT habits are absent or blocked by a short token cap, accuracy improves in the opposite direction. Native \texttt{Llama-3.2-1B-Instruct} has a top Fisher eigenvalue comparable to \texttt{Qwen3-0.6B} ($\mu_0 = 2.7\times 10^5$ vs.\ $1.7\text{--}2.3\times 10^5$; Appendix~\ref{app:b2}), yet responds with the opposite sign (Table~\ref{tab:quadrants}, Appendix~\ref{app:b2}): negative steering shortens outputs, boosts \boxedtok{} compliance, and adds $\mathbf{3.5}$ \textbf{pp}, whereas positive steering lengthens outputs while breaking \boxedtok{} formatting. Similarly, under 32-token direct answering on \texttt{Qwen3-0.6B}, format-axis steering sharply shortens outputs, brings \boxedtok{} usage near ceiling, and adds $\mathbf{4.61}$ \textbf{pp}. Together, these cases show that a large top eigenvalue sets behavioral leverage rather than direction: without latent long CoT, the top sensitive habit is format compliance, so tightening rather than lengthening pays off.
\end{itemize}

\begin{figure}[t]
  \centering
  \includegraphics[width=\textwidth]{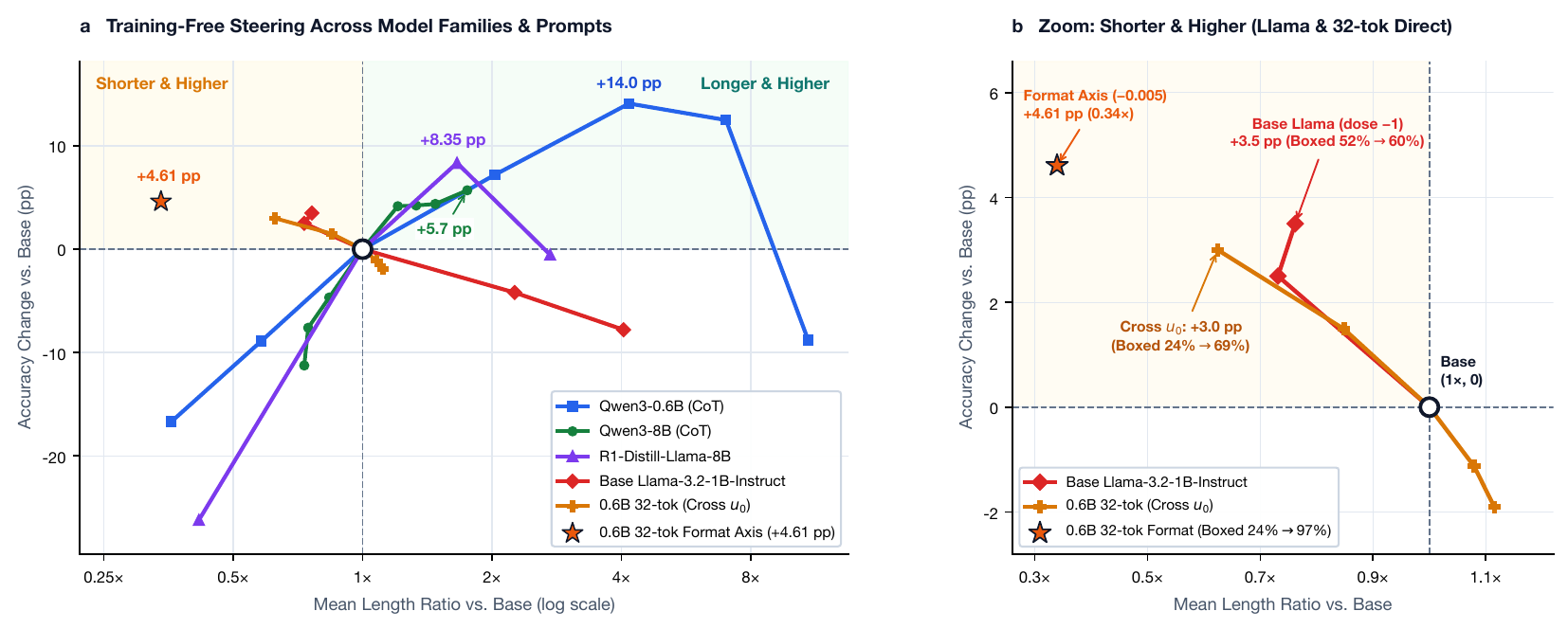}
  \caption{\textbf{Single-axis steering unlocks whichever latent habit (longer derivation or tighter formatting) carries the top Fisher sensitivity under the prompt.} \textbf{(a)} Relative length versus accuracy change across five model--prompt settings (full curves in Table~\ref{tab:b7} and Appendix~\ref{app:b2}). Models with suppressed long-CoT priors (\texttt{Qwen3}, \texttt{R1-Distill}) gain by moving right (longer reasoning), whereas settings lacking latent long CoT or bounded by a 32-token cap (\texttt{Llama-3.2-1B}, direct answering) gain by moving left (shorter, better-formatted answers). Held-out \texttt{Qwen3-0.6B} uses a fixed dose; other sweeps report evaluation-split doses. \textbf{(b)} Zooming into the format-tightening regime shows that the CoT axis $u_0$ transfers only weakly to direct answering, whereas steering along the prompt's own dedicated format axis aligns with the dominant sensitivity, shortening outputs more cleanly and yielding larger gains.}
  \label{fig:quadrants}
\end{figure}

\begin{table}[t]
  \centering
  \caption{Representative single-axis BF16 steering across model families and prompts. In the top three settings, steering improves accuracy as reasoning lengthens; in the bottom two, it improves accuracy as outputs shorten and \boxedtok{} answers become more common. Applied doses are $+1$, $+0.10$, $+0.04$, $-1$, and $-0.005$. For \texttt{Qwen3-0.6B}, the held-out transfer uses a fixed dose ($+13.77\,[11.13, 16.41]$ pp). For the other sweeps, we report the best-performing dose on the evaluation split, so those gains are upper-bound estimates (Appendix~\ref{app:b2}).}
  \label{tab:quadrants}
  \footnotesize
  \resizebox{\textwidth}{!}{%
  \begin{tabular}{@{}lccc@{}}
    \toprule
    Model \& Prompt Setting & Base Acc / Len & Steered $\Delta$Acc / Len & Primary Behavioral Shift \\
    \midrule
    \texttt{Qwen3-0.6B} (MATH \nothink{}) & $45.5\%$ / 585 & $\mathbf{+14.0}$ \textbf{pp} / 2,438 & Pre-box derivation (held-out fixed dose $+13.77$ pp) \\
    \texttt{Qwen3-8B} (MATH \nothink{}) & $81.7\%$ / 820 & $\mathbf{+5.7}$ \textbf{pp} / 1,436 & Monotonic lengthening; $1.6\%$ truncation \\
    \texttt{R1-Distill-Llama-8B} (MATH \nothink{}) & $72.95\%$ / 1,115 & $\mathbf{+8.35}$ \textbf{pp} / 1,849 & Reflection tokens $2.5\times$; thinking reopened \\
    Native \texttt{Llama-3.2-1B-Instruct} (MATH) & $11.3\%$ / 752 & $\mathbf{+3.5}$ \textbf{pp} / 573 & Shorter; \boxedtok{} rate $52\% \to 60\%$ \\
    \texttt{Qwen3-0.6B} (MATH Direct, 32 tok) & $3.64\%$ / 27.68 & $\mathbf{+4.61}$ \textbf{pp} / 9.40 & Earlier \boxedtok{} ($24\% \to 97\%$; dev) \\
    \bottomrule
  \end{tabular}}
\end{table}

\subsection{Subspace Specialization and Pre-Answer Expansion}
\label{sec:anatomy}

\begin{figure}[t]
  \centering
  \includegraphics[width=\textwidth]{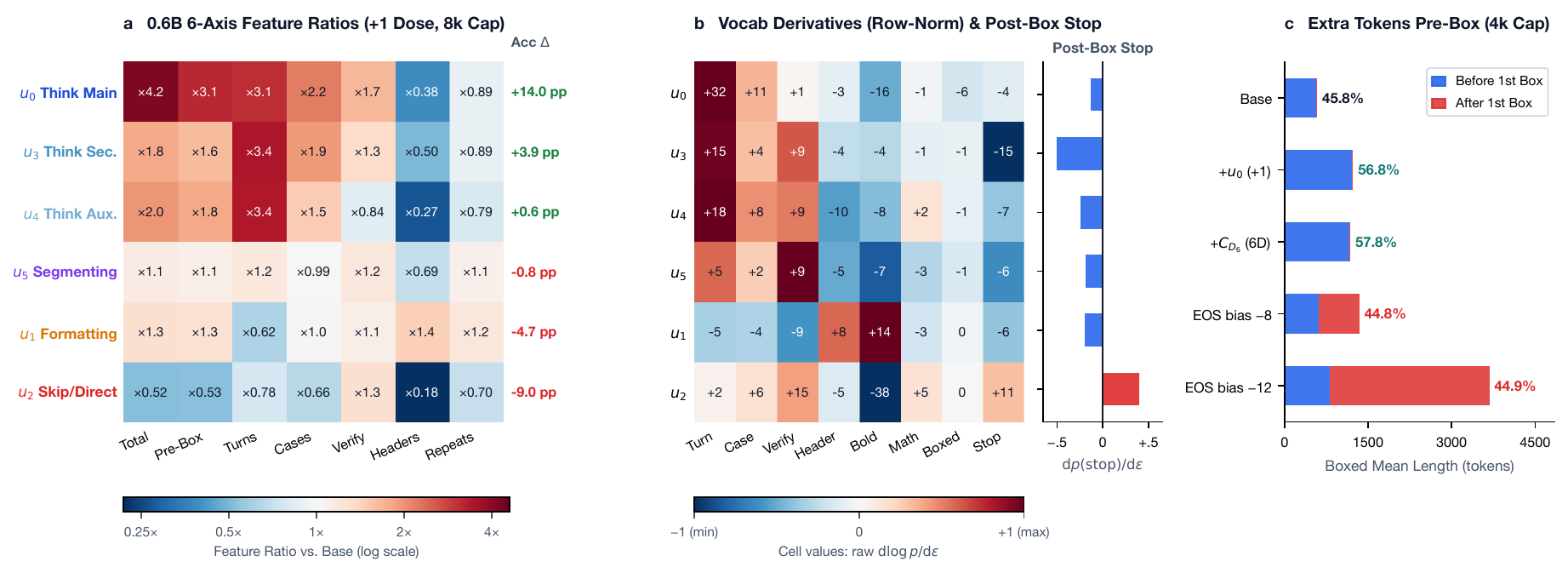}
  \caption{\textbf{$\Dsub$ is functionally specialized rather than a uniform length knob, and useful steering expands derivation before answer commitment (\texttt{Qwen3-0.6B}).} \textbf{(a)} $+1$-dose profiles across all six $\Dsub$ axes (8k cap; Table~\ref{tab:b5}) separate accuracy-lifting ``think-more'' axes ($u_0, u_3, u_4$) from format-clutter ($u_1$) and early-stopping ($u_2$) axes. \textbf{(b)} Token log-probability directional derivatives (row-normalized; raw values annotated) show why: $u_0$ jointly raises reflection tokens while suppressing \boxedtok{} and \texttt{EOS}, delaying answer commitment rather than merely blocking termination. \textbf{(c)} Under a matched 4k cap (Table~\ref{tab:b4}), $u_0$ and $\CD$ expand tokens \emph{before} the first \boxedtok{} and add $11\text{--}12$ pp, whereas decoding \texttt{EOS} suppression piles extra generation primarily \emph{after} the first box with no accuracy gain.}
  \label{fig:anatomy}
\end{figure}

\paragraph{Subspace specialization and pre-\boxedtok{} expansion (Figure~\ref{fig:anatomy}, Tables~\ref{tab:b4}--\ref{tab:b5}).}
To test whether $\Dsub$ is merely a crude length or anti-stopping knob equivalent to suppressing \texttt{EOS}, we perturb each axis individually and compare weight steering against decoding-time \texttt{EOS} logit suppression under a matched 4k cap. On \texttt{Qwen3-0.6B}, the six $\Dsub$ axes specialize cleanly (Table~\ref{tab:b5}): three ``think-more'' axes ($u_0, u_3, u_4$) suppress \texttt{EOS}, increase reflection tokens, and raise accuracy, whereas $u_1$ adds Markdown headings while lowering accuracy and $u_2$ triggers early stopping right after \boxedtok{}; matched-norm random or shuffled directions leave accuracy flat (Table~\ref{tab:b4}). More importantly, where extra tokens land explains why weight steering helps while \texttt{EOS} suppression does not (Figure~\ref{fig:anatomy}b--c): because $u_0$ and $\CD$ jointly boost reflection tokens while suppressing both \boxedtok{} and \texttt{EOS}, they delay answer commitment, more than doubling tokens \emph{before} the first \boxedtok{} and raising accuracy by $11\text{--}12$ pp while still stopping soon after the box. By contrast, decoding-time \texttt{EOS} suppression places extra generation primarily \emph{after} the first \boxedtok{} (post-box mean length surges to $725\text{--}2{,}871$ tokens versus a modest pre-box increase; Table~\ref{tab:b4}), piling repetitive rewrites after the initial answer with no accuracy gain.

\subsection{Implications for Two RLVR Puzzles: The Qwen--Llama Gap and Template Sensitivity}
\label{sec:puzzles}

Two recurring puzzles in the RLVR literature turn out to be direct consequences of which behavioral habit occupies the leading pre-RL Fisher axis and how strongly the prompt suppresses it at step~0.

\paragraph{Why Qwen models surge under math RLVR while native Llama models plateau.}
In R1-Zero-style RLVR (applying verifiable-reward RL directly without a cold-start long-CoT supervised fine-tuning stage), \texttt{Qwen} models reliably lengthen derivations, exhibit reflection and backtracking, and post large accuracy gains, whereas native \texttt{Llama} models barely lengthen and plateau early; \citet{gandhi2025cognitive} link this gap behaviorally to pretraining cognitive habits. Our step-0 spectrum locates the mechanism in parameter geometry. Native \texttt{Llama-3.2-1B-Instruct} has a leading Fisher eigenvalue comparable to \texttt{Qwen3-0.6B} ($\mu_0 = 2.7\times 10^5$ vs.\ $1.7\text{--}2.3\times 10^5$; Appendix~\ref{app:b2}), so it does not lack a steep low-dimensional direction; rather, that direction controls \boxedtok{} formatting instead of extended derivation (\S\ref{sec:quadrants}). Under GRPO, native Llama quickly drives \boxedtok{} compliance near ceiling ($99.8\%$; Appendix~\ref{app:b2}) while accuracy stalls near $30\%$ even at matched learning rates (Table~\ref{tab:d2}), leaving no remaining low-dimensional reasoning mode to arbitrage. By contrast, \texttt{DeepSeek-R1-Distill-Llama-8B} shares the Llama architecture family yet carries distilled long-CoT priors, and its leading axis lengthens derivation (Table~\ref{tab:quadrants}). This contrast points to pre-existing reasoning habits rather than architecture family alone, though distillation and model scale also vary together (\S\ref{sec:limitations}).

\paragraph{Why R1-Zero-style RL is highly sensitive to prompt templates.}
A second widely debated finding is that R1-Zero-style RL produces dramatic accuracy and length jumps under some prompt templates yet modest gains under others \citep{liu2025understanding}. Our Fisher-geometry analysis traces this template sensitivity to two step-0 factors: which behavioral axis (CoT length or answer formatting) carries the largest directional Fisher information $v^\top F_{0,s} v$ under the prompt and token budget (\S\ref{sec:selection-response}, Table~\ref{tab:b6}), and how strongly the template suppresses that behavior before training. Templates that suppress latent reasoning at step~0 depress the pre-RL baseline and create a larger apparent RL jump; in our BF16 decomposition, most of that extra template-induced gain sits in the low-dimensional $\CD$ component rather than in $\Eperp$ (\S\ref{sec:net-gain}, \S\ref{sec:coverage}).

\section{Orthogonal Decomposition in BF16 Execution Weights}
\label{sec:bf16}

Because training-free steering already recovers a substantial portion of the RL gain, measuring what RL adds beyond mode release requires removing the $\Dsub$ displacement from trained checkpoints. This removal must act directly on the BF16 execution weights used in forward passes: projecting out $\Dsub$ only in FP32 master weights before rounding to BF16 leaves a reversed numerical bias large enough to mask orthogonal gains (\S\ref{sec:deadzone}). Correcting this bias isolates persistent orthogonal net gains across our primary split, out-of-domain transfer, external controls, and larger models (\S\ref{sec:net-gain}).

\subsection{The Half-ULP Dead Zone and Iterative BF16 Placement}
\label{sec:deadzone}

For any checkpoint pair $(\theta_a, \theta_b)$, write the BF16 execution displacement as $\Dexec_{a \to b} = Q(\theta_b) - Q(\theta_a)$. With $P = UU^\top$, its \textbf{algebraic orthogonal split} is
\begin{equation}
  \Dexec_{a \to b}
  = \underbrace{P \Dexec_{a \to b}}_{\CD^{a \to b}}
  + \underbrace{(I - P)\Dexec_{a \to b}}_{\Eperp^{a \to b}},
  \qquad b = U^\top \Dexec_{a \to b}.
  \label{eq:split}
\end{equation}
The algebra in Equation~\eqref{eq:split} defines the continuous projection targets. Because BF16 weights lie on a discrete grid, $\CD^{a \to b}$ and $\Eperp^{a \to b}$ cannot both be realized with zero quantization spillover; the \textbf{BF16 execution realizations} instead evaluate achieved BF16 placements that match those target $\Dsub$ coordinates within tolerance via post-hoc $2 \times 2$ placement or online $\Dsub$-locked training, with read-back residuals audited separately (Appendix~\ref{app:b1}).

Subtracting the projection $cu$ along a sensitive unit vector $u$ in FP32 ($\delta'_i = \Delta_i - c u_i$, $c = u^\top \Delta$) and then rounding to BF16 generally leaves a nonzero component along $u$. Here, one \textbf{ULP} (\emph{unit in the last place}) denotes the local spacing between adjacent representable BF16 values at coordinate $\theta_{a,i}$; an update smaller than one local grid step is a \textbf{sub-ULP} change, and any value strictly inside the rounding cell bounded by the midpoints to the adjacent BF16 values rounds back to $\theta_{a,i}$ (approximately $\pm\tfrac{1}{2}\mathrm{ULP}$ away from powers of two; exact boundaries and tie-to-even in Appendix~\ref{app:a2}). We refer to this no-change cell as the \textbf{half-ULP dead zone}. As a result, the executed coordinate along $u$ can reverse sign under the conditions below (proof in Appendix~\ref{app:a2}):

\vspace{3pt}
\begin{proposition}[Reversed Execution Offset from the BF16 Half-ULP Dead Zone]
\label{prop:deadzone}
Suppose the anchor $\theta_a$ lies on the BF16 grid, $u$ is a unit vector, and $c = u^\top \Delta \ne 0$. Let $\mathcal{L} = \{i : Q(\theta_{a,i} + \delta'_i) \ne \theta_{a,i}\}$ be the active set of coordinates that cross a BF16 grid boundary after FP32 subtraction. Define the signed projection ratio of $\Delta$ on $\mathcal{L}$ by $f = \frac{1}{c}\sum_{i \in \mathcal{L}} u_i \Delta_i \in \mathbb{R}$, and the active energy of $u$ on $\mathcal{L}$ by $\mL = \sum_{i \in \mathcal{L}} u_i^2 \in [0,1]$. Let $\varepsilon_Q = \sum_{i \in \mathcal{L}} u_i\bigl[Q(\theta_{a,i} + \delta'_i) - (\theta_{a,i} + \delta'_i)\bigr]$ denote the projected rounding remainder. Then the BF16 execution coordinate along $u$, $\bexec = u^\top\bigl[Q(\theta_a + \delta') - Q(\theta_a)\bigr]$, satisfies
\begin{equation}
  \bexec = (f - \mL)\,c + \varepsilon_Q.
  \label{eq:deadzone}
\end{equation}
In particular, whenever $f < \mL$ and $|\varepsilon_Q| < (\mL - f)|c|$, $\bexec$ reverses sign relative to $c$.
\end{proposition}

\begin{figure}[tb]
  \centering
  \includegraphics[width=\textwidth]{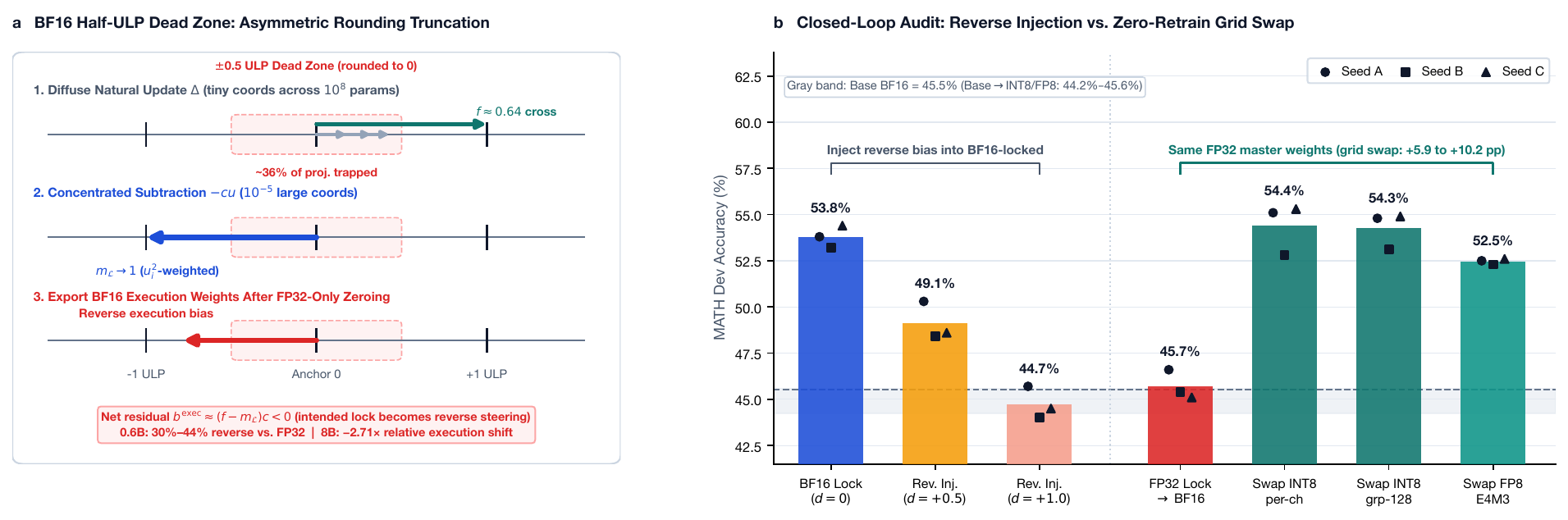}
  \caption{\textbf{FP32-only projection hides orthogonal gains because BF16 rounding asymmetrically erases diffuse updates while preserving concentrated subtraction.} \textbf{(a)} Reversed-offset mechanism (Proposition~\ref{prop:deadzone}): rounding diffuse updates back to the anchor drops ${\approx}\,36\%$ of the original signed projection ($f \approx 0.64$), whereas the sparse subtraction $-cu$ crosses BF16 boundaries almost intact ($\mL \to 1$), leaving a negative reverse-steering residual. \textbf{(b)} Closed-loop 3-seed checks on MATH dev (Table~\ref{tab:a1}): injecting the predicted reversed offset reproduces the full drop (left, sufficiency), while re-quantizing the exact same FP32-locked weights onto misaligned INT8/FP8 grids breaks BF16 dead-zone alignment and recovers accuracy to $52.5\%\text{--}54.4\%$ without retraining while the pre-RL base stays flat (right).}
  \label{fig:deadzone}
\end{figure}

Zeroing $\Dsub$ only in FP32 can therefore turn into reverse steering after BF16 rounding (Figure~\ref{fig:deadzone}a). Because $u$ concentrates on a tiny set of high-leverage coordinates, the subtracted term $-c u_i$ moves nearly all of those coordinates across BF16 grid boundaries ($\mL \to 1$), whereas many coordinates of the diffuse update $\Delta$ fall inside their BF16 rounding cells (the half-ULP dead zone) and drop out upon rounding ($f < \mL$). On \texttt{Qwen3-0.6B}, FP32-only zeroing leaves a reversed offset of $-30\%$ to $-44\%$ relative to the FP32 projection $c$. On \texttt{Qwen3-8B}, where far fewer diffuse weights cross a BF16 step ($6.72\%$), both $f$ and the unprojected BF16 execution coordinate $\cexec = u^\top\Dexec$ shrink while the concentrated subtraction still crosses, magnifying the reversed overshoot to $\mathbf{-2.71\times}$ relative to $\cexec$ (Appendix~\ref{app:a2}, Table~\ref{tab:d1}).

To test whether this rounding offset, rather than an ineffective orthogonal component or quantization in general, causes the apparent collapse, we run two complementary closed-loop checks across 3 seeds on MATH dev (Figure~\ref{fig:deadzone}b, Table~\ref{tab:a1}). While BF16 zeroing within tolerance scores $\mathbf{53.8\%}$, FP32-only zeroing exported to BF16 drops to $\mathbf{45.7\%}$, nearly matching the pre-RL base ($45.5\%$). First, injecting the predicted reversed offset into BF16-zeroed weights reproduces the full accuracy drop, showing that the offset alone suffices to explain the collapse. Second, re-quantizing the exact same FP32-locked master weights onto misaligned INT8 or FP8 grids shifts quantization boundaries away from the BF16 anchor without retraining; diffuse updates trapped in the BF16 dead zone are no longer identically zeroed, recovering the accuracy loss (3-seed means $52.5\%\text{--}54.4\%$, matching or exceeding the $53.8\%$ BF16-zeroed baseline on INT8; Table~\ref{tab:a1}) while pre-RL base weights on the same grids remain flat. Together, these checks show that the orthogonal gains remained intact in FP32 all along and were hidden specifically by BF16 dead-zone alignment.

\paragraph{Iterative BF16 placement and per-arm initializations.}
To remove this execution bias, we use \textbf{iterative BF16 placement} for all steering, post-hoc splitting, and online locking. Starting from an arm-specific FP32 displacement $d^{(0)}$, we iteratively adjust the $\Dsub$ coordinates toward a target vector $b^\star$:
\begin{equation}
  d^{(t+1)} = d^{(t)} + U\Bigl(b^\star - U^\top\bigl[Q(\theta_a + d^{(t)}) - Q(\theta_a)\bigr]\Bigr).
  \label{eq:placer}
\end{equation}
We iterate until $\max_j |b^\star_j - u_j^\top[Q(\theta_a + d^{(t)}) - Q(\theta_a)]|$ falls below tolerance or reaches the iteration cap (Appendix~\ref{app:b1}). We initialize Equation~\eqref{eq:placer} for the three post-hoc $2 \times 2$ arms as follows:
\begin{itemize}
  \item \textbf{$\CD$-only arm (and steering):} $d^{(0)} = 0$ with target $b^\star = b$ (or $b^\star = \alpha s e_k$ for steering along $u_k$, where $s$ is the calibrated unit-dose scale and $e_k$ is the $k$-th standard basis vector);
  \item \textbf{$\Eperp$-only arm:} $d^{(0)} = (I - P)\Dexec_{a \to b}$ (FP32 algebraic orthogonal component) with target $b^\star = 0$, converging to the placed displacement $d_E$;
  \item \textbf{Combined ($\CD + \Eperp$) arm:} $d^{(0)} = Q(\theta_a + d_E) - Q(\theta_a)$ (realized BF16 $\Eperp$-only displacement) with target $b^\star = b$.
\end{itemize}
For \textbf{online $\Dsub$-locked training}, each step first runs standard AdamW on the FP32 master weights (updating optimizer moments as usual), and then applies Equation~\eqref{eq:placer} to those master weights with $b^\star = 0$ without modifying the moments, stopping once the read-back error falls below tolerance or hits the step cap (Appendix~\ref{app:b1}). Appendix~\ref{app:b1} also audits per-arm BF16 spillover norms and verifies agreement with stochastic rounding. Every evaluation below runs strictly on read-back BF16 execution weights.

\subsection{Net Gains Across Tasks, External Controls, and Model Scales}
\label{sec:net-gain}

\paragraph{Primary BF16 $2 \times 2$ decomposition (Table~\ref{tab:split}).}
Using iterative BF16 placement to split 40-step GRPO ($26.4$ pp mean held-out gain) cleanly separates mode release from orthogonal improvement (Table~\ref{tab:split}). Restoring $\CD$ alone contributes $\mathbf{15.7}$ \textbf{pp} (about $60\%$ of the total gain) and drives nearly all length growth, whereas \textbf{restoring $\bm{\Eperp}$ alone improves accuracy in every seed, averaging $\mathbf{+7.5}$ pp} with only modest length growth. Recombining $\CD + \Eperp$ recovers nearly the entire unconstrained gain and beats $\CD$ alone by $8.3\text{--}11.7$ pp across seeds (symmetric Shapley shares: $66\%$ $\CD$ vs.\ $34\%$ $\Eperp$).

Three controls narrow what drives this orthogonal gain and how it interacts with $\Dsub$. First, to test additivity versus synergy, we measure the four-arm interaction $A_{CE} - A_C - A_E + A_0$ (where $A_0$, $A_C$, $A_E$, and $A_{CE}$ are the accuracies of the pre-RL base, $\CD$-only, $\Eperp$-only, and combined arms): it averages positive on accuracy yet varies across seeds (Table~\ref{tab:split}), whereas the interaction on error recovery is consistently positive (Table~\ref{tab:b10}), showing that overall accuracy is roughly additive while error recovery depends on both components cooperating. Second, to rule out generic orthogonal perturbation or regularization, we replace verifiable-reward $\Eperp$ with an orthogonal component trained under shuffled rewards; accuracy drops $\mathbf{4.7\text{--}7.2}$ \textbf{pp} below $\CD$ alone, indicating that $\Eperp$'s benefit is reward-directed. Third, to test whether online RL can bypass $\Dsub$ during training, we run online $\Dsub$-locked GRPO with BF16 displacement in $\Dsub$ held near zero within tolerance; it still gains $15.6$ pp on average and lengthens outputs (Table~\ref{tab:split}). When the steepest 6D path is blocked online, optimization reroutes part of the lengthening and mode-release pressure into secondary directions inside $\Eperp$, explaining why post-hoc BF16 splitting is essential to isolate orthogonal gains without $4\times$ lengthening.

\begin{table}[t]
  \centering
  \caption{Online $\Dsub$-locking and BF16 $2 \times 2$ decomposition across 3 seeds on \texttt{Qwen3-0.6B} (MATH \nothink{}, 512 held-out problems, $n=4$, step~40). A paired McNemar test comparing $\CD + \Eperp$ against $\CD$ gives $p \le 5.6\times 10^{-15}$. The 4-arm interaction $A_{CE}-A_C-A_E+A_0$ averages $+2.39$ pp (Seeds A/B/C: $-0.05$ / $+4.64$ / $+2.59$ pp). In \S\ref{sec:coverage}--\S\ref{sec:sharpening}, the pre-RL base mean is $42.78\%$ under a separate $n=64$ sampling run.}
  \label{tab:split}
  \footnotesize
  \renewcommand{\arraystretch}{1.16}
  \resizebox{\textwidth}{!}{%
  \begin{tabular}{@{}lccc@{}}
    \toprule
    Configuration & Accuracy (Seeds A / B / C; Mean) & $\Delta$ vs.\ Base & Mean Length (tok) \\
    \midrule
    Pre-RL base & $44.0\%$ (Seed-A batch $43.7\%$) & --- & 577--586 \\
    Base $+1.0\,u_0$ (ref-scaled fixed dose) & $57.8\%$ & $+13.77\,[11.13, 16.41]$ & 2,465 \\
    Unconstrained GRPO (step 40) & $70.4\%$ / $69.6\%$ / $71.3\%$ (\textbf{70.4\%}) & $\mathbf{+26.4}$ & 2,395--2,964 \\
    Online $\Dsub$-locked GRPO (step 40) & $56.4\%$--$61.2\%$ ($59.6\%$) & $+15.6$ & 1,070--2,320 \\
    Post-hoc: $\CD^{0\to 40}$ only & $60.2\%$ / $56.8\%$ / $62.2\%$ (\textbf{59.7\%}) & $\mathbf{+15.7}$ & 2,347--3,626 \\
    \textbf{Post-hoc: $\bm{\Eperp^{0\to 40}}$ only} & $53.8\%$ / $51.1\%$ / $49.7\%$ (\textbf{51.5\%}) & $\mathbf{+7.5}$ & 699--970 \\
    Post-hoc: $\CD^{0\to 40} + \Eperp^{0\to 40}$ & $70.2\%$ / $68.5\%$ / $70.5\%$ (\textbf{69.7\%}) & $\mathbf{+25.7}$ & 2,397--3,032 \\
    $\CD^{0\to 40}$ + shuffled-reward $\Eperp$ & $4.7\text{--}7.2$ pp below $\CD$ only & --- & --- \\
    \bottomrule
  \end{tabular}}
\end{table}

\paragraph{Out-of-domain transfer to MMLU and GSM8K (Table~\ref{tab:b8}).}
To test whether $\CD$ and $\Eperp$ reflect task-bound length habits or transferable improvements, we evaluate the Seed-A MATH BF16 split zero-shot on unseen MMLU \citep{hendrycks2021mmlu}, GSM8K \citep{cobbe2021gsm8k} (under both lenient and strict \texttt{\#\#\#\#} scoring), and \texttt{wikitext-2} \citep{merity2017wikitext} negative log-likelihood (NLL; Table~\ref{tab:b8}). Restoring $\CD$ (or steering along $u_0$) acts as an indiscriminate lengthening switch: it lifts MMLU by $3.8\text{--}5.0$ pp and boosts lenient GSM8K scoring, yet triples response length and disrupts the required \texttt{\#\#\#\#} format, cutting strict GSM8K accuracy nearly in half. By contrast, \textbf{$\bm{\Eperp}$ alone} keeps response length and \texttt{wikitext-2} NLL nearly unchanged while raising MMLU by $3.7$ pp and strict GSM8K accuracy by $\mathbf{6.9}$ \textbf{pp} (with \texttt{\#\#\#\#} compliance rising from $26.3\%$ to $38.4\%$). Combining $\CD + \Eperp$ matches full RL on both transfer tasks. Thus, $\CD$ blindly triggers verbose derivations that can break new format constraints, whereas $\Eperp$ transfers across benchmarks without inflating length or hurting general language modeling.

\paragraph{Accuracy gains under template, format, and scratchpad controls (Table~\ref{tab:controls}, Tables~\ref{tab:b13}--\ref{tab:b14}).}
To test whether RL gains survive once surface shortcuts are closed outside our subspace split, we design three external controls that each remove a different way length or formatting can inflate apparent RL gains (Table~\ref{tab:controls}). Control~1 eliminates template suppression by evaluating our trained RL checkpoint under the unsuppressed native \think{} template (a separate native-\think{} RL training run is compared in \S\ref{sec:coverage}). Control~2 removes format gaps in 32-token direct answering via \boxedtok{} prefilling or format-axis steering so that box rates approach ceiling. Control~3 removes written scratchpads entirely, evaluating \texttt{Qwen3-8B} on GSM8K prefilled with \texttt{"\#\#\#\# "} for numeric-only outputs \citep{deng2023implicit,pfau2024dot}. Across the primary split and all three external controls (even in $3.5$-token numeric-only outputs with no written CoT), net single-sample RL gains persist at $\mathbf{5.5}$ \textbf{to} $\mathbf{9.7}$ \textbf{pp} (Table~\ref{tab:controls}), confirming that single-sample accuracy gains survive once these surface shortcuts are closed.

\begin{table}[t]
  \centering
  \caption{Net single-sample RL gains under the BF16 subspace split and the template, format, and scratchpad controls (in pp).}
  \label{tab:controls}
  \footnotesize
  \setlength{\tabcolsep}{4pt}
  \renewcommand{\arraystretch}{1.12}
  \begin{tabularx}{\textwidth}{@{}>{\raggedright\arraybackslash}p{3.45cm}>{\raggedright\arraybackslash}Xccc@{}}
    \toprule
    Setting \& Control & Comparison & Base & After RL & Net Gain \\
    \midrule
    Primary: \texttt{0.6B} MATH \nothink{} & Post-hoc $\Eperp^{0\to 40}$ vs.\ base (512 held-out, 3 seeds) & $44.0\%$ & $51.5\%$ & $\mathbf{+7.5}$ ($+5.7$ to $+9.8$) \\
    \addlinespace[3pt]
    Control 1: \texttt{0.6B} MATH \think{} (16k) & Step-100 RL in \think{} vs.\ base \think{} (1,024 dev, 1 seed) & $71.4\%$ & $77.4\%$ & $\mathbf{+6.0}$ \\
    \addlinespace[3pt]
    Control 2: \texttt{0.6B} MATH Direct (32 tok) & Prefill \boxedtok{} / format-steered vs.\ RL (1,024 dev, 1 seed) & $8.11\text{--}8.25\%$ & $13.77\text{--}13.87\%$ & $\mathbf{+5.5}$ to $\mathbf{+5.8}$ \\
    \addlinespace[3pt]
    Control 3: \texttt{Qwen3-8B} GSM8K Mental (64 tok) & 3-seed RL vs.\ base, both with \texttt{"\#\#\#\# "} prefill (1,319 test) & $17.68\%$ & $27.39\%$ (3.5 tok) & $\mathbf{+9.71}$ \\
    \bottomrule
  \end{tabularx}
\end{table}

\paragraph{Cross-scale and cross-family replication (Table~\ref{tab:d1}).}
To test whether the half-ULP dead zone and orthogonal gains are specific to \texttt{Qwen3-0.6B} or the \nothink{} template, we repeat the dead-zone audit, BF16 $2 \times 2$ split, and 8-subset gradient decomposition (detailed in \S\ref{sec:step20}) on \texttt{Qwen3-8B} and on \texttt{Qwen2.5-1.5B} under a standard prompt with no \think{}/\nothink{} mode switch (Table~\ref{tab:d1}). All three main findings replicate. First, FP32-only zeroing drops both models below their pre-RL held-out baselines, confirming that the dead-zone trap persists across scales. Second, BF16 placement within tolerance isolates positive $\Eperp$ gains on both models, and $\Eperp$ outweighs $\CD$ on both because weaker or absent template suppression leaves less latent lengthening for $\CD$ to unlock (\S\ref{sec:puzzles}): on \texttt{Qwen2.5-1.5B} (2 seeds), $\Eperp$ adds $+3.3\text{--}4.4$ pp on held-out vs.\ $+1.3\text{--}2.4$ pp for $\CD$, and on \texttt{Qwen3-8B} (1 seed), $\Eperp$ adds $+3.61$ pp on dev and $+1.76$ pp on held-out (sampling $\mathrm{SE}\approx 1.15$ pp; Table~\ref{tab:d1}). Third, across sampled checkpoints, the cosine similarity $\cos(g_R^\perp, g_L^\perp)$ between orthogonal reward and log-length gradients stays near zero or negative (Table~\ref{tab:d1}), indicating that reward optimization outside $\Dsub$ does not systematically push for longer outputs and that lengthening concentrates in $\Dsub$.

\section{Training Dynamics: Step-20 Transition and Mixed-Precision Geometry}
\label{sec:dynamics}

Sections~\ref{sec:arbitrage}--\ref{sec:bf16} decomposed fixed checkpoints spatially into concentrated mode release ($\CD$) and diffuse orthogonal updates ($\Eperp$). Turning from static snapshots to training trajectories over steps 0--100 raises two coupled mechanistic questions: \textbf{(1)~Why does rapid lengthening stall near step~20 while accuracy keeps rising, and how do component roles shift afterward?} (\S\ref{sec:step20}); and \textbf{(2)~Once optimization shifts toward diffuse late updates, how do sub-ULP changes accumulate across BF16 execution precision?} (\S\ref{sec:unified}). Near step~20, shared reward gradients drop sharply alongside rising 4k truncation; afterward, $\Eperp$ drives standalone accuracy and error-recovery gains through sub-ULP FP32 accumulation on weights whose BF16 values remain temporarily static (full mixed-precision synthesis in Appendix~\ref{app:c2}, Table~\ref{tab:unified}).

\subsection{The Step-20 Transition Window: Weakening Shared Gradients and Post-Transition Component Roles}
\label{sec:step20}

\begin{figure}[t]
  \centering
  \includegraphics[width=\textwidth]{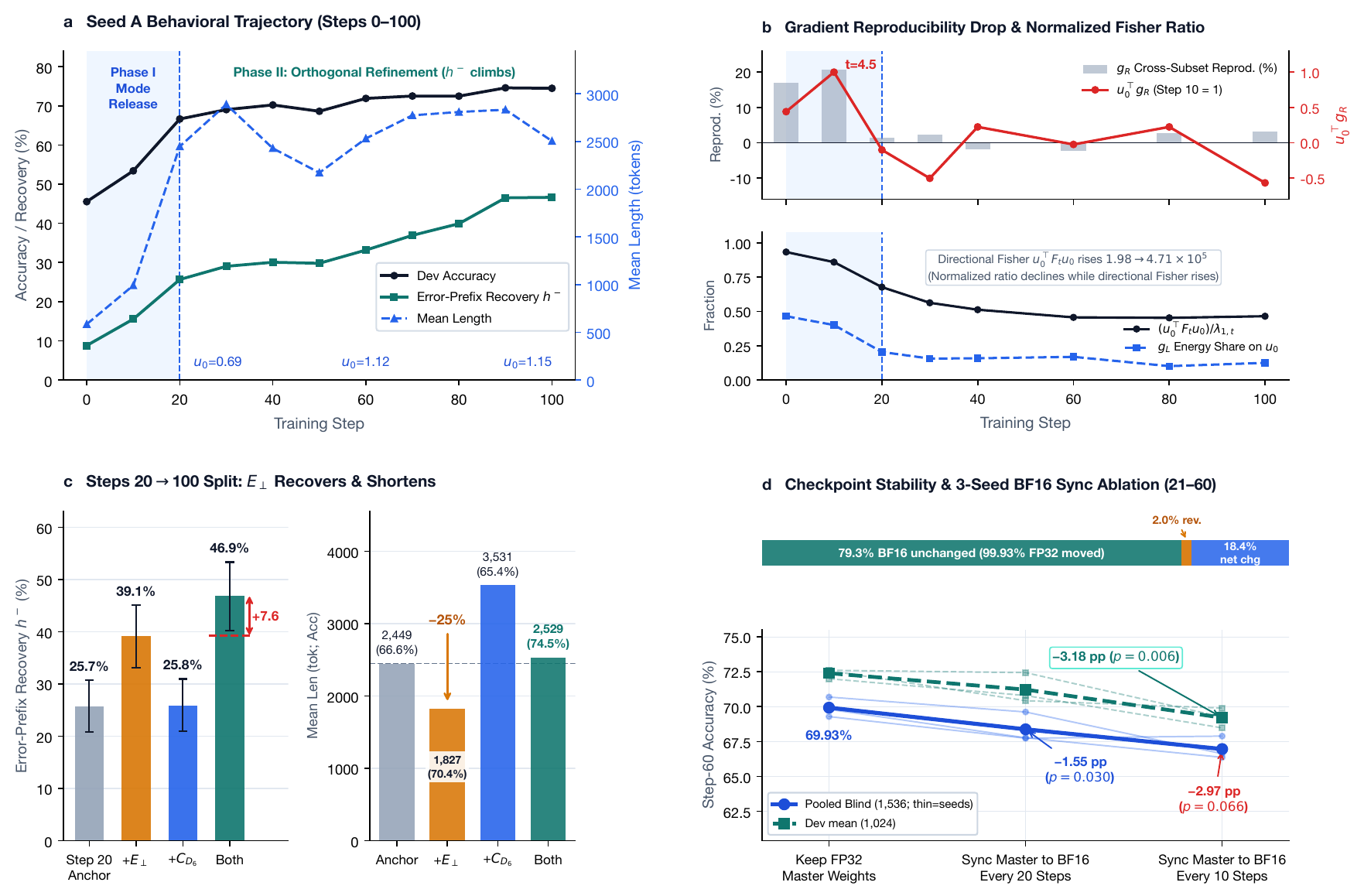}
  \caption{\textbf{Training transitions near step~20 from shared mode release to diffuse refinement relying on sub-ULP FP32 accumulation.} \textbf{(a)} Seed~A trajectories (Table~\ref{tab:c2}) show $u_0$ displacement and mean length leveling off near step~20 while dev accuracy and corrupted-prefix recovery ($h^-$; \S\ref{sec:recovery}) keep rising through step~100. \textbf{(b)} Seed~A cross-subset gradient diagnostics (all 3 seeds in Tables~\ref{tab:c1}, \ref{tab:c4}) show reproducible reward-gradient energy and $u_0^\top g_R$ dropping sharply by step~20 (top), while the normalized Fisher ratio $(u_0^\top F_t u_0)/\lambda_{1,t}$ and $\eta_{u_0}(g_L)$ decline even as raw directional Fisher curvature $u_0^\top F_t u_0$ rises (bottom curves and inset), ruling out parameter flattening alongside rising 4k truncation. \textbf{(c)} Anchoring at step~20 ($20\to 100$; Table~\ref{tab:b10}) separates late roles: $\Eperp^{20\to 100}$ shortens length $25\%$ while lifting accuracy and $h^-$, residual $\CD^{20\to 100}$ stretches outputs without aiding recovery, and combining both restores length and yields a $7.6$ pp super-additive recovery gain. \textbf{(d)} Over steps 20--60, $79.3\%$ of weights keep identical BF16 values across checkpoints while $99.93\%$ shift in FP32 (top); clearing those sub-ULP residuals via periodic BF16 sync lowers 3-seed pooled blind and dev accuracy (bottom; lockbox and Seed-A masked splits in Tables~\ref{tab:c7}--\ref{tab:unified}).}
  \label{fig:dynamics}
\end{figure}

\paragraph{Front-loaded lengthening and the cross-subset gradient test (Figure~\ref{fig:dynamics}a--b, Tables~\ref{tab:c1}--\ref{tab:c6}).}
Near step~20, mean response length levels off while dev accuracy and corrupted-prefix recovery $h^-$ (the fraction of continuations recovering the correct answer after one intermediate digit is altered; \S\ref{sec:recovery}) keep climbing through step~100 (Figure~\ref{fig:dynamics}a, Tables~\ref{tab:c2}--\ref{tab:c3}). Most early growth occurs within the first 20 steps: on Seed~A, step~20 already captures ${\sim}70\%$ of the eventual dev gain, ${\sim}85\%$ of the step-100 \passk{64} gain (probability of $\ge 1$ correct answer in 64 samples; \S\ref{sec:coverage}, Table~\ref{tab:b15a}), and $60\%$ of the final BF16 $u_0$ displacement ($50\%\text{--}61\%$ across seeds in BF16; $70\%$ in FP32 on Seed~C; Tables~\ref{tab:c2}--\ref{tab:c3}, \ref{tab:c6}).

This divergence raises a direct diagnostic question: does progression along $u_0$ slow after step~20 because the cross-problem shared reward signal weakens, because the 4,096-token cap intervenes, or because parameter curvature along $u_0$ flattens? To separate cross-problem reward gradients from rollout noise, we partition the dev problems into 8 disjoint subsets at each checkpoint across all three seeds (Appendix~\ref{app:c1}). Letting $g_R^{(i)}$ denote the reward gradient on subset~$i$ and $g_R = \sum_{i=1}^8 g_R^{(i)}$ the aggregated dev reward gradient, we measure the cross-subset reproducible gradient energy $\widehat{s}_R = \sum_{i \ne j}\langle g_R^{(i)}, g_R^{(j)} \rangle / \|g_R\|_2^2$ (Figure~\ref{fig:dynamics}b, Tables~\ref{tab:c1}, \ref{tab:c4}--\ref{tab:c5}).

Evaluating all three hypotheses separates parameter curvature from shared signal and cap effects (Tables~\ref{tab:c1}, \ref{tab:c4}--\ref{tab:c5}). First, \emph{curvature flattening is ruled out}: letting $F_t$ denote the trajectory Fisher matrix at step~$t$, directional Fisher information $u_0^\top F_t u_0$ roughly doubles over steps 0--20, even as emerging local modes lower the normalized ratio $(u_0^\top F_t u_0)/\lambda_{1,t}$ (where $\lambda_{1,t}$ is the largest Ritz value at step~$t$). Second, the \emph{cross-problem shared reward gradient drops sharply toward noise}: from step~10 to step~20, reproducible energy $\widehat{s}_R$ falls from $11.1\%\text{--}22.0\%$ (where $45\%\text{--}82\%$ of the denoised gradient lies in $\Dsub$ along $+u_0$) to $1.5\%\text{--}3.6\%$, and $u_0^\top g_R$ loses statistical significance in every seed (with no lasting positive realignment thereafter; Table~\ref{tab:c4}). This drop is not merely denominator growth as outputs lengthen: the unnormalized cross-subset numerator $\sum_{i \ne j}\langle g_R^{(i)}, g_R^{(j)}\rangle$ itself plunges $7.3\text{--}7.6\times$ on Seeds~A and C and nearly halves on Seed~B (Table~\ref{tab:c1}). Different problems thus stop sharing a single low-dimensional update direction after step~20.

Third, \emph{truncation under the 4,096-token cap rises sharply in parallel}, climbing from $2.1\%\text{--}3.9\%$ at step~10 to $15.5\%\text{--}22.4\%$ at step~20 across seeds (Tables~\ref{tab:c1}, \ref{tab:c4}; Seed-A training-window truncation in Table~\ref{tab:b17b}). Thus, weakening shared gradients and rising 4k truncation coincide at step~20. Even when these 4k-trained step-20 checkpoints are evaluated under an 8,192-token cap, further $+u_0$ steering adds near-zero accuracy ($0$ pp on Seed~A; ${\sim}1$ pp 3-seed mean; Table~\ref{tab:b12}, Figure~\ref{fig:recovery}c), pointing to diminishing returns alongside the 4k cap (cleanly decoupling the two requires an 8k-cap training rerun; \S\ref{sec:limitations}).

\paragraph{Post-step-20 component roles over steps 20--100 (Figure~\ref{fig:dynamics}c, Table~\ref{tab:b10}).}
Once the shared lengthening gradient weakens near step~20, how do $\CD$ and $\Eperp$ contribute over steps 20--100? Anchoring at step~20 and splitting the subsequent BF16 displacement ($20 \to 100$) separates their roles cleanly. On Seed~A, adding $\Eperp^{20\to 100}$ alone cuts mean length by $25\%$ while lifting corrupted-prefix recovery $h^-$ from $25.7\%$ to $39.1\%$ and raising dev accuracy by $3.8$ pp, with standalone $h^-$ gains replicating across all three seeds (Table~\ref{tab:b10}). By contrast, adding residual $\CD^{20\to 100}$ alone stretches responses well past step~20 without improving recovery. Recombining both restores step-20 length while unlocking $7.6\text{--}17.8$ pp of super-additive recovery across Seeds~A and C (Table~\ref{tab:b10}). Over steps 20--100, $\Eperp^{20\to 100}$ therefore drives standalone accuracy and error-recovery gains alongside shorter outputs, whereas residual $\CD^{20\to 100}$ no longer helps alone yet counterbalances $\Eperp$'s shortening when combined.

\subsection{Mixed-Precision Geometry of Diffuse Updates: Sub-ULP FP32 Accumulation}
\label{sec:unified}

Unlike early shifts along $u_0$, which concentrate on ${\sim}10^{-5}$ parameters that cross BF16 grid steps immediately (\S\ref{sec:deadzone}), late-stage updates are diffuse and high-rank: in a per-matrix SVD cross-seed audit of the full step-20--60 displacement across the 196 core weight matrices, singular directions beyond rank $r = 128$ carry $60.52\%$ of the cross-seed inner-product alignment in BF16 (Table~\ref{tab:c7}). Spread across hundreds of millions of parameters, most per-step updates are sub-ULP. Across two seeds over steps 20--60, $99.93\%$ of FP32 master weights move, yet $\mathbf{79.3\%}$ keep identical BF16 values at steps 20, 40, and 60 (Figure~\ref{fig:dynamics}d). That discrepancy raises a direct question: do the hidden sub-ULP FP32 residuals on apparently static weights actually drive Phase~II learning?

Two synchronization ablations over steps 21--60 answer yes (Appendix~\ref{app:c2}, Table~\ref{tab:c7}). At each sync step, the intervention leaves read-back BF16 weights unchanged and clears only sub-ULP FP32 master residuals. Periodically snapping all FP32 master weights onto the BF16 grid every 20 or 10 steps (recalibrating $\Dsub$ before resuming) lowers 3-seed lockbox accuracy by $\mathbf{1.96}$ \textbf{pp} ($p=0.0266$) and $\mathbf{3.05}$ \textbf{pp} ($p=0.0202$), with matching $1.55\text{--}2.97$ pp drops on the pooled blind panel combining held-out and lockbox (Table~\ref{tab:c7}).

To isolate which coordinates drive this loss on Seed~A, we run coordinate-masked syncing every 10 steps, separating weights that have \emph{not yet crossed} a BF16 boundary since step~20 from already-crossed ones. Syncing only not-yet-crossed weights lowers pooled blind accuracy by $\mathbf{3.05}$ \textbf{pp} (nearly matching full sync), whereas syncing only already-crossed weights changes it by $\mathbf{+0.60}$ \textbf{pp}, yielding a paired gap of $\mathbf{+3.65}$ \textbf{pp} on pooled blind and $\mathbf{+3.96}$ \textbf{pp} on lockbox (Table~\ref{tab:c7}). Phase~II diffuse refinement therefore relies on sub-ULP FP32 residuals accumulating in static-BF16 weights until they cross grid boundaries in later steps (early/late synthesis in Table~\ref{tab:unified}).

\section{What Changes Beyond Mode Release? Error Recovery and Within-Support Answer Reallocation}
\label{sec:refinement}

Sections~\ref{sec:arbitrage}--\ref{sec:dynamics} established that Phase~I shifts along $\Dsub$ amplify pre-existing lengthening or formatting modes, whereas $\Eperp$ yields net accuracy gains outside $\Dsub$ and drives standalone accuracy and error-recovery gains after step~20. Aggregate accuracy alone, however, does not specify what behavioral changes emerge beyond mode release. We therefore open with the \textbf{two measurable behavioral changes} induced by orthogonal refinement: \textbf{(1)~stronger error-sensitive recovery from corrupted intermediate steps} (\S\ref{sec:recovery}), and \textbf{(2)~selective probability reallocation toward correct answers within the pre-RL solvable support}, which lifts small-$k$ hit rates and trims truncation without inflating wrong pre-RL modes (\S\ref{sec:coverage}--\S\ref{sec:sharpening}). Around these two findings, we separate genuine refinement from three residual confounds not settled by static single-sample splits: \textbf{(i)~off-axis lengthening, rewrite boilerplate, or restarting from scratch} (\S\ref{sec:recovery}), \textbf{(ii)~apparent large-$k$ coverage expansion under template suppression} (\S\ref{sec:coverage}), and \textbf{(iii)~uniform temperature scaling or answer-level power-law sharpening} (\S\ref{sec:sharpening}).

\subsection{Error-Sensitive Recovery and Frontier Shift Beyond Lengthening}
\label{sec:recovery}

\begin{figure}[t]
  \centering
  \includegraphics[width=\textwidth]{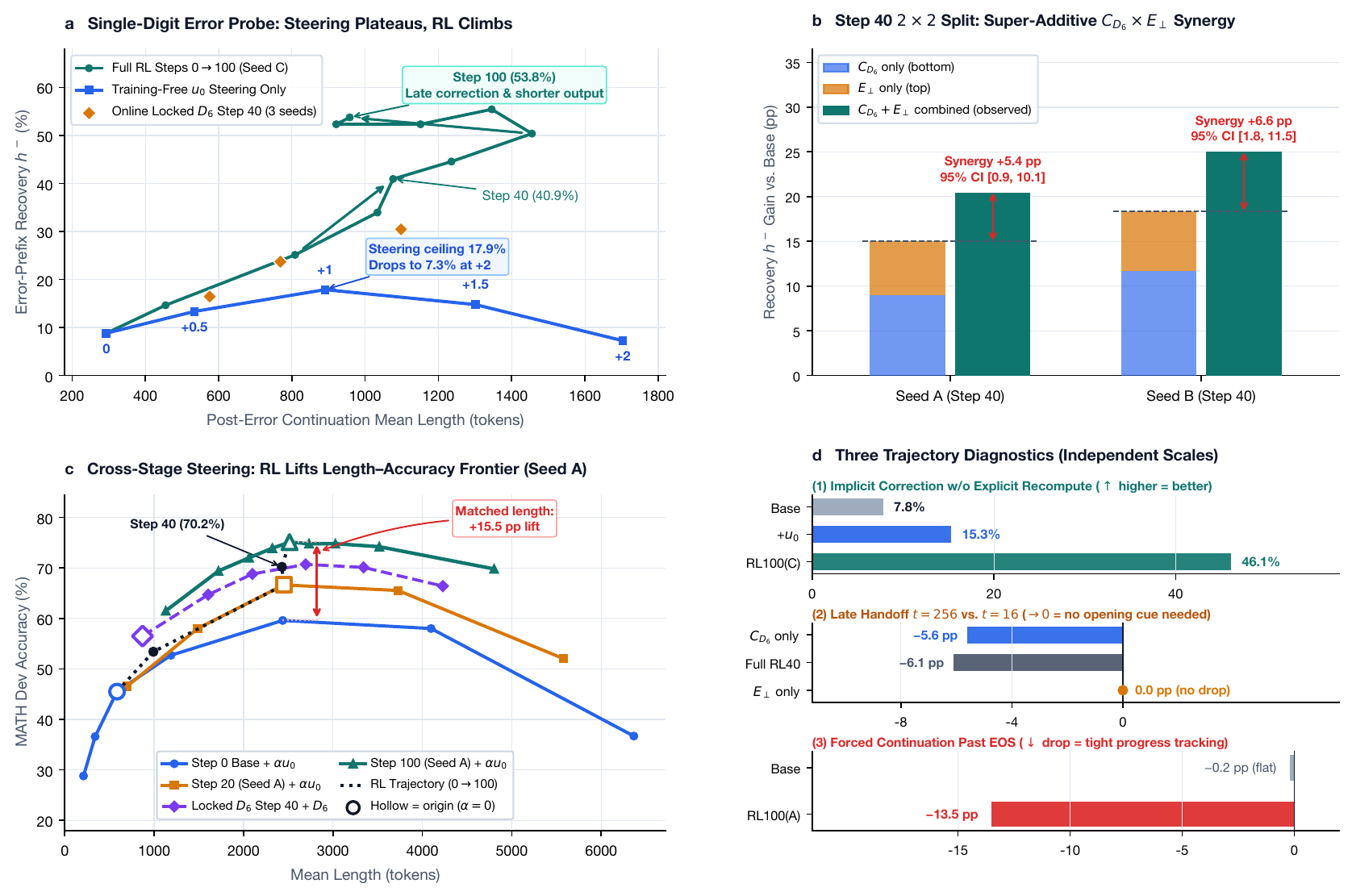}
  \caption{\textbf{Beyond Phase~I mode release, RL strengthens error-sensitive recovery and lifts the accuracy--length frontier.} \textbf{(a)} On 193 corrupted-prefix pairs (Table~\ref{tab:b9}), lengthening alone peaks at $17.9\%$ and collapses at $+2\,u_0$, whereas full RL reaches $42.9\%\text{--}52.6\%$ across 3 seeds ($53.8\%$ on paired Seed~C, shown) while shortening late continuations. \textbf{(b)} Step-40 $2\times 2$ splits (Table~\ref{tab:b10}) show super-additive recovery gains ($5.4$ and $6.6$ pp on Seeds~A/B) when pairing $\CD$ and $\Eperp$. \textbf{(c)} Cross-stage steering on MATH dev (Seed~A, 8k cap; Table~\ref{tab:b12}) rules out pure off-axis lengthening: full RL already sits ${\sim}7$ pp above the length-matched base steering peak by step~20 and climbs vertically to $+15.5$ pp at step~100. \textbf{(d)} Continuation diagnostics (Table~\ref{tab:b11}) rule out rewrite boilerplate and blind continuation: implicit recovery rises sharply without rewrite cues (Seed~C shown), isolated $\Eperp$ holds accuracy on mid-handoffs, and forcing generation past \texttt{EOS} hurts RL accuracy.}
  \label{fig:recovery}
\end{figure}

\paragraph{Single-digit corruption probe on 193 prefix pairs (Figure~\ref{fig:recovery}a--b, Tables~\ref{tab:b9}--\ref{tab:b10}).}
Our first measurable behavioral change is an increased ability to continue solving accurately after an intermediate error. We construct 193 prefix pairs truncated at the same intermediate equation \citep{lanham2023faithfulness,tyen2024llms}: each clean prefix is error-free, whereas its paired corrupted prefix alters one digit in the final intermediate step. Sampling continuations with matched seeds (Table~\ref{tab:b9}), we measure \textbf{clean accuracy $h^+$}, \textbf{corrupted recovery $h^-$}, and the \textbf{corruption penalty $h^+ - h^-$}. Pre-RL base recovery is $8.8\%$; $u_0$ steering peaks at $17.9\%$ at $+1\,u_0$ and collapses back to $7.3\%$ at $+2\,u_0$, showing that extra length alone yields only occasional retries before failing through drift and truncation. By contrast, online $\Dsub$-locked training raises recovery at matched length ($16.5\%\text{--}30.4\%$ across 3 seeds), and step-100 full RL reaches $\mathbf{42.9\%\text{--}52.6\%}$ across all three seeds in the unified trajectory audit ($\mathbf{48.7\%\text{--}53.8\%}$ on Seeds~A/C in the paired evaluation batch; Tables~\ref{tab:b9}, \ref{tab:b11b}, \ref{tab:c3}) while shrinking the clean-to-corrupt drop by $\mathbf{18.1\text{--}22.4}$ \textbf{pp} relative to the best steering dose on Seeds~A and C (Table~\ref{tab:b10}).

BF16 $2\times 2$ splits with problem-clustered bootstrap CIs isolate how $\CD$ and $\Eperp$ build this recovery (Figure~\ref{fig:recovery}b, Table~\ref{tab:b10}). By step~40, isolated $\Eperp$ lifts $h^-$ modestly while its clean-to-corrupt drop-reduction CIs still include zero; combining $\CD$ and $\Eperp$ lifts $h^-$ above additivity by $\mathbf{5.44}$ and $\mathbf{6.61}$ \textbf{pp} on Seeds~A and B (95\% CIs excluding zero). Direct narrowing of the clean-to-corrupt gap by isolated $\Eperp$ emerges later, in $\Eperp^{20\to 100}$ on Seeds~A and B (Table~\ref{tab:b10}). Thus, early recovery gains arise super-additively when $\Eperp$ is paired with $\CD$'s longer derivations, whereas late $\Eperp$ independently reduces the clean-to-corrupt accuracy drop.

\paragraph{Testing residual lengthening, boilerplate, and restart accounts (Figure~\ref{fig:recovery}c--d, Tables~\ref{tab:b11}--\ref{tab:b12}, Figure~\ref{fig:b1}).}
Because isolated $\Eperp^{0\to 40}$ still lengthens slightly, three residual hypotheses remain: off-axis lengthening along the base curve, memorized self-correction boilerplate, or ignoring the corrupted prefix to restart from scratch. First, across matched mean lengths (1k--2.5k tokens), step-100 RL beats the upper convex hull of all tested reward-free length controls (template switching, $u_0$ steering, and truncation) by $\mathbf{7\text{--}14}$ \textbf{pp}, and restoring $\CD$ after 40-step locked training still exceeds the pre-RL steering peak by $\mathbf{11.1}$ \textbf{pp} (Table~\ref{tab:b12}). Meanwhile, extra accuracy from $u_0$ steering falls from $14.0$ pp at step~0 to near zero by step~20 (${\sim}1$ pp 3-seed mean; $0$ pp on Seed~A) and $0$ pp at step~100, even along the recomputed top axis $B\text{-}u_0$ or a 1D reward-free length direction from the 24-dimensional cross-checkpoint length subspace $P_{24}$, showing that RL lifts the entire accuracy--length frontier rather than sliding further along a length axis.

Second, continuation audits rule out surface shortcuts: reciting rewrite boilerplate, copying an earlier true value, or discarding the prefix to restart from scratch (Figure~\ref{fig:recovery}d, Tables~\ref{tab:b11}--\ref{tab:b11b}). Even when restricted to continuations with no explicit rewrite phrase across all 193 pairs, step-100 RL lifts implicit recovery from $7.8\%$ (base) to $\mathbf{40.1\%\text{--}46.1\%}$ (Seeds~A/C); separately, on the 54-pair early-corruption subset where the prefix never restates the true value, it eliminates the clean-to-corrupt drop ($+11.1 \to 0.0$ pp on both seeds; Table~\ref{tab:b11}). Nor does RL discard the prefix to restart: explicit restart cues appear in only $0.9\%\text{--}3.3\%$ of recoveries across seeds, $44.3\%\text{--}55.0\%$ cite the altered digit within the first 500 characters, and clean continuations reuse the supplied prefix length rather than matching full \texttt{no\_prefix} generation (Table~\ref{tab:b11b}). Together, these checks show that RL reads and corrects the given prefix rather than ignoring it. Additional mid-solution handoff checks (continuing from a late 256-token base prefix, isolated $\Eperp$ holds accuracy at $0.0$ pp vs.\ $-5.6$ pp for $\CD$) and post-\texttt{EOS} checks (forcing generation past \texttt{EOS} drops RL accuracy by $13.5$ pp, confirming calibrated stopping; Figure~\ref{fig:recovery}d) appear in Appendix~\ref{app:b3} (Table~\ref{tab:b11}).

\subsection{Small-\texorpdfstring{$k$}{k} Concentration Within the Matched-Template Coverage Boundary}
\label{sec:coverage}

\begin{figure}[t]
  \centering
  \includegraphics[width=\textwidth]{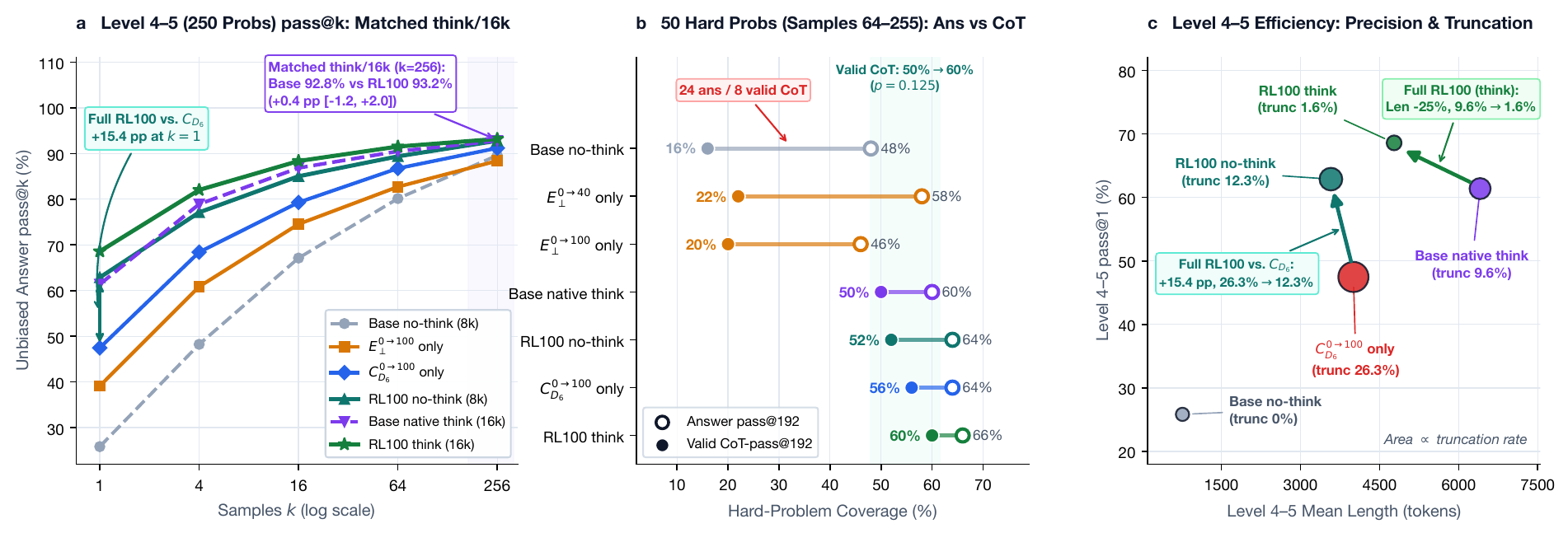}
  \caption{\textbf{Once template and budget are matched, 100-step RL lifts small-$k$ hit rates and trims truncation while aggregate large-$k$ coverage stays close to the pre-RL base (Seed~A; Table~\ref{tab:passk}).} \textbf{(a)} On \texttt{Level 4--5} ($250$ problems, $n=256$ samples), matched \think{}/16k evaluation shows RL raising \passk{1} while \passk{256} tracks the pre-RL base ($+0.4$ pp); under \nothink{}/8k, full RL leads $\CD$ alone by $15.4$ pp at $k=1$. \textbf{(b)} On Hard-50 (disjoint $n=192$; open: answer-only, filled: judge-valid), the apparent \cotpassk{192} jump over pre-RL \nothink{} mirrors Phase~I template release: both $\CD$ alone and unlocking pre-RL \think{}/16k recover most of the gap. \textbf{(c)} Length--\passk{1} efficiency (area $\propto$ truncation): full RL halves truncation vs.\ $\CD$ under \nothink{}/8k and cuts it from $9.6\%$ to $1.6\%$ under \think{}/16k.}
  \label{fig:passk}
\end{figure}

While \S\ref{sec:arbitrage}--\S\ref{sec:bf16} isolated Phase~I template release at single-sample accuracy, prior studies disagree on whether RL expands \emph{large-$k$} solvable coverage: \citet{yue2025does} find that pre-RL models catch up at large $k$ under answer-only \passk{k}, whereas \citet{wen2025rlvr} report lasting gains under judge-verified \cotpassk{K} and ProRL \citep{liu2025prorl} reports sustained \passk{k} gains from prolonged multi-domain RL. Extending our Phase~I template-release control to large~$k$ on Seed~A, we evaluate RL against the pre-RL base under \textbf{matched native \think{}/16k} alongside $\CD$, $\Eperp$, and \nothink{} RL under the same protocol (Figure~\ref{fig:passk}, Table~\ref{tab:passk}). We compute unbiased \passk{k} ($n=256$; \citealt{chen2021evaluating}) on \texttt{Level 4--5} problems, where large-$k$ coverage is not yet saturated. For \cotpassk{192}, we select the 50 hard problems missed by all first 64 pre-RL \nothink{} samples (\textbf{Hard-50}), then evaluate them on the \textbf{192 disjoint remaining samples} using \texttt{Qwen3.8-27B} as judge (full curves, paired CIs, and judge calibration in Tables~\ref{tab:b15a}--\ref{tab:b15b}).

\begin{table}[t]
  \centering
  \caption{Unbiased \texttt{Level 4--5} \passk{k} ($250$ problems, $n=256$) and hard-problem \cotpassk{192} ($50$ problems, disjoint samples \texttt{64--255}) on Seed~A (BF16). All lengths and truncation rates are measured over the same $250 \times 256$ \texttt{Level 4--5} samples. $^*$ marks 95\% CI excluding 0 vs.\ pre-RL \nothink{}. Relative to pre-RL \think{} under matched \think{}/16k, \cotpassk{192} differs by $+10\,[0, 20]$ pp ($p=0.125$) for transferred RL and $+2\,[-12, +14]$ pp ($p=1.00$) for native-\think{} RL (a separate 100-step RL run starting from the pre-RL checkpoint under \think{} with an 8k cap).}
  \label{tab:passk}
  \footnotesize
  \setlength{\tabcolsep}{4pt}
  \renewcommand{\arraystretch}{1.18}
  \resizebox{\textwidth}{!}{%
  \begin{tabular}{@{}llccccc@{}}
    \toprule
    Model / Configuration & Template / Cap & L4--5 @1 & L4--5 @64 & L4--5 @256 & Len / Trunc & Hard Ans / CoT@192 (Valid\%) \\
    \midrule
    Pre-RL base & \nothink{} / 8k & $25.84$ & $80.15$ & $89.6$ & 762 / $0.0\%$ & $48.0\%$ / $\mathbf{16.0\%}$ ($41.6\%$) \\
    $\Eperp^{0\to 40}$ only & \nothink{} / 8k & $35.89$ & $84.56$ & $91.2$ ($+1.6$) & 1,400 / $1.0\%$ & $58.0\%$ / $22.0\%$ ($57.2\%$) \\
    $\Eperp^{0\to 100}$ only & \nothink{} / 8k & $39.11$ & $82.72$ & $88.4$ ($-1.2$) & 1,306 / $0.1\%$ & $46.0\%$ / $20.0\%$ ($58.8\%$) \\
    \textbf{$\bm{\CD^{0\to 100}}$ only} & \nothink{} / 8k & $47.47$ & $86.76$ & $91.2$ ($+1.6$) & 4,004 / $\mathbf{26.3\%}$ & $64.0\%$ / $\mathbf{56.0\%}$ ($79.4\%$) \\
    Full RL step 100 & \nothink{} / 8k & $62.87$ & $89.39$ & $\mathbf{92.8}$ ($+3.2$) & 3,577 / $12.3\%$ & $64.0\%$ / $\mathbf{52.0\%}$ ($69.4\%$) \\
    \textbf{Pre-RL base \think{}} & \think{} / 16k & $61.38$ & $90.49$ & $\mathbf{92.8}$ ($+3.2^*$) & 6,407 / $9.6\%$ & $60.0\%$ / $\mathbf{50.0\%}$ ($63.6\%$) \\
    \textbf{RL step 100 \think{}} & \think{} / 16k & $\mathbf{68.58}$ & $91.53$ & $\mathbf{93.2}$ (vs.\ \think{} $+0.4$) & \textbf{4,776 / 1.6\%} & $66.0\%$ / $\mathbf{60.0\%}$ ($68.8\%$) \\
    \textbf{Native \think{} RL step 100} & \think{} / 16k & $63.93$ & $89.83$ & $\mathbf{92.4}$ (vs.\ \think{} $-0.4$) & \textbf{5,673 / 0.2\%} & $64.0\%$ / $\mathbf{52.0\%}$ ($61.2\%$) \\
    \bottomrule
  \end{tabular}}
\end{table}

\paragraph{Matched \think{}/16k finite-budget coverage.}
Comparing directly against pre-RL \think{} under matched \think{}/16k, both 100-step runs lift \passk{1} ($\mathbf{+7.2}$ \textbf{pp} transferred; $+2.6$ pp native-\think{}) and slash truncation from $9.6\%$ to $\le 1.6\%$ (Table~\ref{tab:passk}). At this sampling budget, however, aggregate large-$k$ coverage stays flat within uncertainty: \texttt{L4--5} \passk{256} remains within $\pm 0.4$ pp of pre-RL \think{}, and Hard-50 \cotpassk{192} shows wide CIs ($+2$ to $+10$ pp, $p \ge 0.125$) alongside problem-level turnover (full CIs, overlap counts, and 8k rescoring in Table~\ref{tab:b15b}). Over 100 steps, native-\think{} RL also shortens outputs relative to pre-RL \think{}, though its longer training rollouts under the 8k cap yield slower \passk{1} progress than transferred \nothink{} RL.

\paragraph{How Phase~I template release explains apparent \nothink{} coverage jumps.}
Comparing against the suppressed pre-RL \nothink{} baseline explains why apparent coverage gains look much larger without template matching. Under \nothink{}, the pre-RL model answers nearly half of the 50 hard problems correctly at least once ($48.0\%$) yet produces a judge-valid chain on only $16.0\%$ (\cotpassk{192}). Relative to this suppressed baseline, RL raises \cotpassk{192} by $36$ pp, matching the pattern reported by \citet{wen2025rlvr}. Yet $\CD$ alone recovers most of the \nothink{} \passk{64} gain (Table~\ref{tab:b15a}), reaches $\mathbf{56.0\%}$ on Hard-50 \cotpassk{192} (vs.\ $52.0\%$ for full RL), and accounts for half of the \texttt{L4--5} \passk{256} gap over pre-RL \nothink{}, whereas isolated $\Eperp$ stays close to base on both large-$k$ metrics (Table~\ref{tab:passk}). Evaluating the same pre-RL checkpoint under native \think{}/16k likewise brings its aggregate large-$k$ rates close to RL (Tables~\ref{tab:passk}, \ref{tab:b15b}). These results reconcile \citet{yue2025does} and \citet{wen2025rlvr} in our 100-step math setting: large \cotpassk{K} gains under a suppressed template are real within \nothink{}, yet mirror Phase~I template release rather than expanding aggregate coverage beyond native \think{}.

\paragraph{Small-$k$ accuracy and budget compression (Figure~\ref{fig:passk}c).}
Pairing \passk{1} with length and truncation clarifies the exact arms driving small-$k$ efficiency (Table~\ref{tab:passk}). Under \nothink{}/8k, full step-100 RL improves \texttt{L4--5} \passk{1} by $15.4$ pp and reduces truncation from $26.3\%$ to $12.3\%$ relative to the leading-component-only checkpoint ($\CD^{0\to 100}$). Under matched \think{}/16k, full step-100 RL improves \passk{1} by $7.2$ pp (native-\think{} RL: $+2.6$ pp), shortens outputs by $25\%$, and reduces truncation from $9.6\%$ to $1.6\%$ relative to the pre-RL base model. Thus, within the existing large-$k$ repertoire, full RL compresses verbose derivations inside the token budget and turns latent solvability into reliable single-sample hits.

\subsection{Is Small-\texorpdfstring{$k$}{k} Gain Just Sharpening? Within-Support Reallocation, Temperature, and Majority-Vote Controls}
\label{sec:sharpening}

\begin{figure}[t]
  \centering
  \includegraphics[width=\textwidth]{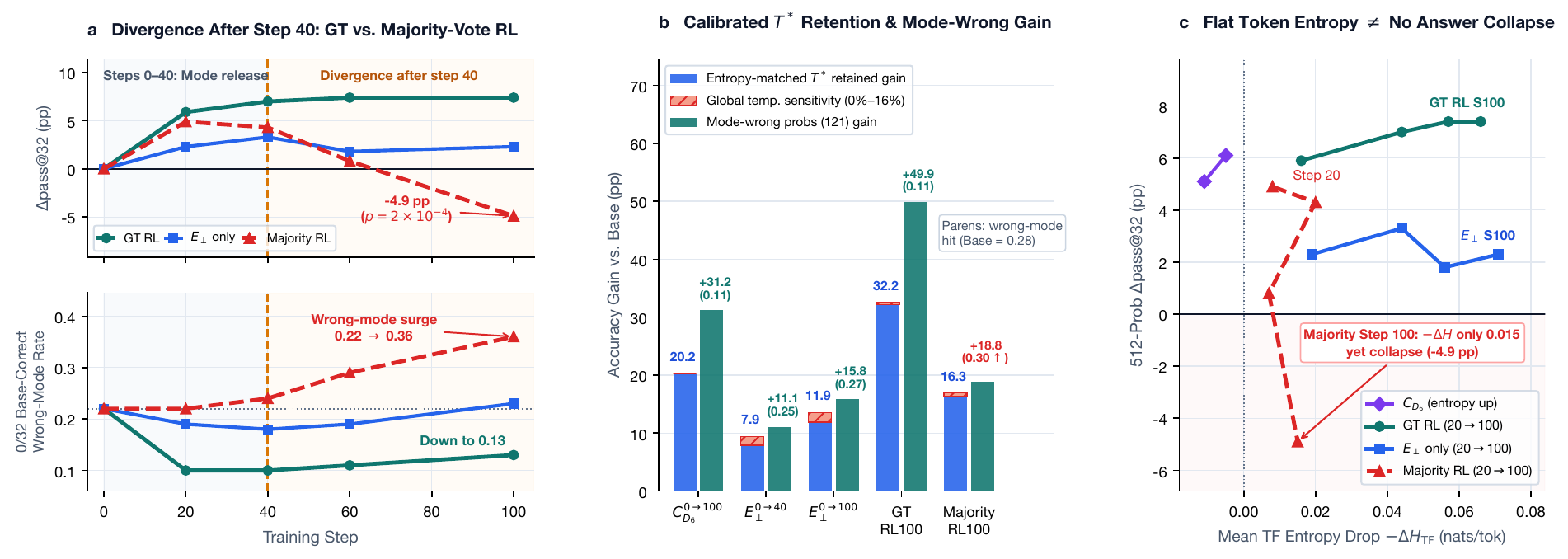}
  \caption{\textbf{Ground-truth RL and $\Eperp$ selectively reallocate mass toward correct answers, whereas majority-vote RL suffers answer-level collapse missed by mean token entropy (512 held-out, Seed~A; Table~\ref{tab:sharpening}).} \textbf{(a)} After step~40, majority-vote RL still leads base at $k=1$ ($+16.9$ pp) yet drops $4.9$ pp below base in \passk{32} ($p=2\times 10^{-4}$) as it concentrates onto fixed wrong pseudo-labels (\texttt{0/32 base-correct} wrong-mode hits reach $0.36$). \textbf{(b)} Entropy-matched temperature calibration ($T^*$) leaves most sampling gains intact (hatched: sensitivity), and on \texttt{mode\_wrong\_with\_support} prompts both $\Eperp$ and ground-truth RL raise accuracy without inflating wrong-mode hits (parentheses). \textbf{(c)} Token-averaged TF entropy barely drops ($-0.015$ nats/token) during majority-vote collapse because answer lock-in at a few branching steps is diluted over routine tokens.}
  \label{fig:sharpening}
\end{figure}

Because \S\ref{sec:coverage} shows that 100-step RL mainly lifts small-$k$ hit rates while aggregate large-$k$ coverage stays flat, our second behavioral question is \emph{how} probability mass shifts across answers: does $\Eperp$ selectively move mass toward correct answers without inflating wrong pre-RL modes, or is the gain merely uniform temperature scaling or answer-level power-law sharpening? Auditing our KL-free, clipping-free REINFORCE checkpoints on Seed-A held-out with fixed-pool teacher-forcing entropy $\HTF$ (Appendix~\ref{app:b4}), we run three diagnostics (Figure~\ref{fig:sharpening}, Table~\ref{tab:sharpening}; full steps in Tables~\ref{tab:b16a}--\ref{tab:b17b}): \textbf{(1)~base-polarity stratification} via 32 disjoint pre-RL samples into strata including \texttt{mode\_wrong\_with\_support} (121 prompts where the modal pre-RL answer is wrong yet the correct answer appears at least once) and \texttt{0/32 base-correct} (82 prompts where none of the 32 pre-RL samples is correct); \textbf{(2)~entropy-matched temperature calibration ($T^*$)} raising temperature until $\HTF$ matches or exceeds the pre-RL baseline; and \textbf{(3)~a ground-truth-free majority-vote GRPO control} rewarding each prompt's fixed pre-RL sample mode ($62.7\%$ ground-truth agreement across valid-mode prompts, so early steps carry partial correctness signal before late training concentrates on fixed wrong pseudo-labels).

\begin{table}[t]
  \centering
  \caption{Within-support reallocation and sharpening diagnostics on the 512-problem held-out split (Seed~A; full steps in Tables~\ref{tab:b16a}--\ref{tab:b17b}). The first 32 pre-RL samples define polarity strata (\texttt{mode\_wrong\_with\_support}, abbreviated \texttt{mode\_wrong}: 121 problems; \texttt{0/32 base-correct}: 82 problems) and the disjoint last 32 samples evaluate performance. $T^*$ is the sampling temperature restoring $\HTF$ to at least the pre-RL threshold. The last column reports the wrong-mode hit rate on the 82 \texttt{0/32 base-correct} prompts (lower means less wrong-mode lock-in) alongside overall $\Delta\passk{32}$. Asterisks ($^*$) mark the 95\% CI $[0.6, 8.8]$ for the $\Eperp^{0\to 100}$ greedy gain and the paired McNemar $p=2\times 10^{-4}$ for majority-vote step-100 overall $\Delta\passk{32}$.}
  \label{tab:sharpening}
  \footnotesize
  \setlength{\tabcolsep}{3.5pt}
  \renewcommand{\arraystretch}{1.18}
  \resizebox{\textwidth}{!}{%
  \begin{tabular}{@{}lccccc@{}}
    \toprule
    Model / Checkpoint & $T=1$ / Greedy $\Delta$ & $\Delta \HTF$ & Calibrated $T^*$ (Sens.\%) & Wrong-Mode $\Delta$ (Hit) & 0/32 Wrong-Mode / Overall $\Delta$@32 \\
    \midrule
    Pre-RL base & $42.78\%$ / $51.17\%$ & $0$ & --- & Base ($0.28$) & $0.22$ / Base \\
    $\CD^{0\to 100}$ only & $+20.2$ / $+8.6$ & $+0.011$ & $+20.2$ ($0\%$) & $+31.2$ ($0.11$) & $0.09$ / $+5.1$ \\
    $\Eperp^{0\to 40}$ only & $+9.4$ / $+2.5$ & $-0.044$ & $+7.9$ ($16\%$) & $+11.1$ ($0.25$) & $0.18$ / $+3.3$ \\
    \textbf{$\bm{\Eperp^{0\to 100}}$ only} & $\mathbf{+13.5}$ / $\mathbf{+4.7^*}$ & $-0.071$ & $\mathbf{+11.9}$ ($12\%$) & $\mathbf{+15.8}$ ($0.27$) & $0.23$ / $+2.3$ \\
    \textbf{Ground-truth RL step 100} & $\mathbf{+32.5}$ / $\mathbf{+22.1}$ & $-0.066$ & $\mathbf{+32.2}$ ($1\%$) & $\mathbf{+49.9}$ ($0.11$) & $\mathbf{0.13}$ / $+7.4$ \\
    Majority-vote RL step 40 & $+20.0$ / $+12.9$ & $-0.020$ & $+19.4$ ($3\%$) & $+25.5$ ($0.22$) & $0.24$ / $+4.3$ \\
    \textbf{Majority-vote RL step 100} & $\mathbf{+16.9}$ / $\mathbf{+8.2}$ (drop) & $-0.015$ & $+16.3$ ($3\%$) & $+18.8$ ($\mathbf{0.30}$) & $\mathbf{0.36}$ / $\mathbf{-4.9^*}$ \\
    \bottomrule
  \end{tabular}}
\end{table}

\paragraph{Selective answer reallocation versus answer-level power-law sharpening.}
To test whether gains come from blind mode amplification, we examine the 121 \texttt{mode\_wrong\_with\_support} prompts (Table~\ref{tab:sharpening}). Under answer-level power-law sharpening $P_0(a)^\alpha$ ($\alpha > 1$), concentrating mass on existing peaks would have to inflate the wrong pre-RL mode. Instead, both $\Eperp^{0\to 100}$ ($+15.8$ pp; wrong-mode hit $0.28 \to 0.27$) and ground-truth RL ($+49.9$ pp; $0.28 \to 0.11$) raise accuracy substantially without inflating wrong-mode hits, showing selective reallocation toward correct answers rather than uniform mode amplification.

\paragraph{Temperature calibration and mild late sharpening.}
A second test separates selective reallocation from uniform token-level temperature scaling (Table~\ref{tab:sharpening}). Because positive temperature scaling leaves greedy ($\arg\max$) decoding strictly invariant, cooling alone cannot produce greedy gains. Under entropy-matched temperature calibration ($T^*$), $\Eperp^{0\to 100}$ retains $\mathbf{11.9}$ \textbf{of its} $\mathbf{13.5}$ \textbf{pp} sampling gain alongside a significant $\mathbf{+4.7}$ \textbf{pp} greedy gain, and exceeds pre-RL greedy accuracy even at $T=1$ (Table~\ref{tab:sharpening}). Full RL likewise retains $97\%\text{--}100\%$ of its sampling gain under $T^*$ from step~20 onward ($92\%$ at step~10; Table~\ref{tab:b16a}), whereas $\CD$ increases mean TF entropy ($\Delta \HTF > 0$). Separately, isolated $\Eperp$ improves $T=1$ sampling before greedy decoding, while rising logit dispersion and non-zero temperature sensitivity indicate mild late sharpening alongside selective reallocation under KL-free training (full trajectories and counterfactual bounds in Appendix~\ref{app:b4}, Tables~\ref{tab:b16a}--\ref{tab:b16b}).

\paragraph{Majority-vote RL positive control: answer collapse missed by token entropy (Figure~\ref{fig:sharpening}a,c).}
Majority-vote RL provides a positive control showing how mode concentration onto fixed pre-RL pseudo-labels unfolds under identical GRPO hyperparameters. Through step~40, it gains $20.0$ pp at $k=1$ (Table~\ref{tab:sharpening}), consistent with partially accurate pseudo-labels and Fisher-aligned mode release. After step~40, ground-truth and majority-vote trajectories diverge sharply as majority-vote RL concentrates further onto fixed wrong pseudo-labels: by step~100, it still leads pre-RL base by $16.9$ pp at $k=1$, yet overall \passk{32} drops $\mathbf{4.9}$ \textbf{pp} below base ($p=2\times 10^{-4}$) and wrong-mode hits on \texttt{0/32 base-correct} prompts climb to $\mathbf{0.36}$. Despite this coverage collapse, mean TF entropy falls by only $0.015$ nats/token because answer-level lock-in at a few branching choices is diluted across thousands of routine tokens. Tracking \passk{k} and \texttt{0/32 base-correct} wrong-mode rates alongside token entropy is therefore essential to catch answer-level collapse.

\section{Related Work}
\label{sec:related}

\paragraph{Reasoning RL and long-CoT dynamics.}
Systems and algorithms such as DeepSeek-R1 \citep{deepseek2025r1}, Kimi~k1.5 \citep{kimi2025k15}, DAPO \citep{yu2025dapo}, Dr.\ GRPO \citep{liu2025understanding}, and ProRL \citep{liu2025prorl} train reasoning models using policy-gradient RL with verifiable rewards \citep{williams1992reinforce,schulman2017ppo,shao2024deepseekmath}. Prior studies attribute response lengthening either to emergent cognitive behaviors \citep{yeo2025demystifying,gandhi2025cognitive} or to template and aggregation biases \citep{liu2025understanding}. We analyze the Fisher spectrum at step~0 to predict early response-length directions and prompt-dependent reversals before RL begins, and we separate low-dimensional mode arbitrage from orthogonal refinement.

\paragraph{Test-time compute control and steering.}
Prior work controls reasoning length externally through decoding budget forcing \citep{muennighoff2025s1}, length-controlled RL \citep{aggarwal2025l1}, overthinking mitigation \citep{chen2024overthinking,wu2025moreisless,fatemi2025concise,wang2025l2t}, and no-thinking inference \citep{ma2025without}. Separately, activation-steering methods extract reasoning vectors from contrastive rollouts \citep{turner2023steering,zou2023representation,arditi2024refusal,chen2025seal,liu2025fractional}. By contrast, we estimate the leading step-0 Fisher axis $u_0$ from unlabeled rollouts and shift weights directly, adjusting pre-box derivation or format tightening without contrastive trajectories or runtime hooks.

\paragraph{Habit elicitation, sharpening, and capability boundaries.}
One line of work shows that one-shot RL \citep{wang2025oneshot}, entropy minimization \citep{agarwal2025unreasonable}, and self-certainty rewards \citep{zhao2025without} can amplify pretraining priors \citep{zhao2025echo}. Consistent with this view, base models often approach RL performance at large $k$ \citep{yue2025does}. Another line reports persistent coverage gains under judge-verified \cotpassk{K} \citep{wen2025rlvr} and prolonged multi-domain \passk{k} evaluation \citep{liu2025prorl}, alongside evidence of broader generalization and retention \citep{chu2025sft,shenfeld2025rlrazor}, higher semantic rank \citep{huang2025semantic}, and improved self-correction \citep{huang2024selfcorrect,kumar2025score}. We combine three distinct controls to separate these effects: matched native-\think{} baselines test pre-existing coverage, corrupted-prefix probes \citep{lanham2023faithfulness,tyen2024llms} and scratchpad-free arithmetic \citep{deng2023implicit,pfau2024dot} test behavioral recovery and non-CoT gains, and majority-vote RL with temperature calibration tests distribution sharpening.

\paragraph{Optimization geometry and evolutionary parallels.}
Natural-gradient methods establish the Fisher information matrix as a local metric for neural optimization \citep{amari1998natural,kakade2001natural,schulman2015trpo,martens2015kfac,martens2020new,kunstner2019empirical}, while spectral and subspace studies show that curvature and effective updates often concentrate in a small top eigenspace or low-dimensional subspace \citep{sagun2017hessian,gurari2018tiny,ghorbani2019hessian,li2018intrinsic,aghajanyan2021intrinsic}. Recent RLVR studies examine parameter sparsity and off-principal updates \citep{adewuyi2026multiple,zhu2025path,wu2026sparsity,mukherjee2025small,zhang2026fisher}; other work develops curvature-based stabilization and training curricula \citep{kim2026flatrlvr,ge2026grpo,melo2026capo,okonkwo2026prompts}. Drawing on selection--response theory from quantitative genetics \citep{fisher1930genetical,robertson1966culling,price1970selection,lande1979quantitative,lande1983measurement,schluter1996adaptive,grimmer2026darwin}, we link the step-0 Fisher information matrix to early behavioral shifts and identify the step-20 transition window directly from cross-subset gradient reproducibility.

\paragraph{Mixed-precision numerics.}
Work on numerical precision has examined mixed-precision underflow, low-bit quantization, and rollout staleness \citep{gupta2015limited,micikevicius2018mixed,micikevicius2022fp8,kalamkar2019bfloat16,zamirai2020revisiting,dettmers2022int8,frantar2023gptq,qi2025fp16,gao2026bf16,weng2026lost}. Here, we characterize the asymmetric half-ULP dead-zone bias that arises when removing concentrated directions from diffuse updates in FP32 before BF16 rounding. Through periodic synchronization across three seeds and coordinate masking on Seed~A, we further link late-stage accuracy loss under BF16 master syncing to sub-ULP FP32 residuals on static-BF16 (not-yet-crossed) weights.

Prior work largely treats early lengthening, capability sharpening, and mixed-precision numerics separately, leaving open how much of the observed RL gain follows from the leading pre-RL Fisher geometry once the trained displacement is evaluated in BF16.

\section{Discussion, Limitations, and Conclusion}
\label{sec:discussion}

\subsection{Implications for Reasoning-RL Analysis and Mixed-Precision Training}
\label{sec:accounting}

Aggregate accuracy under a single prompt template conflates mode release, orthogonal refinement, and sharpening. In our primary setup, low-dimensional mode arbitrage accounts for roughly $60\%$ of the overall accuracy gain and reproduces the apparent Hard-50 \cotpassk{192} jump over \nothink{}, while majority-vote RL can raise single-sample accuracy yet still lose \passk{32} coverage with almost no change in mean token entropy (Figure~\ref{fig:sharpening}). Table~\ref{tab:accounting} summarizes the diagnostic controls used in this work to disentangle those mechanisms. Our mixed-precision findings carry two direct implications for analyzing and training reasoning RL:
\begin{enumerate}
  \item \textbf{Weight attribution should be verified in forward-pass precision.} Subspace projections, ablations, and splits should be performed directly on read-back BF16 weights (e.g., via Equation~\eqref{eq:placer}) rather than FP32 master weights; otherwise, half-ULP dead-zone offsets from FP32-only locking can mask orthogonal accuracy gains.
  \item \textbf{Late-stage RL relies on retaining FP32 master residuals.} Periodic BF16 master-weight synchronization lowers Phase~II accuracy across three seeds on both the pooled blind panel and the 1,024-problem lockbox, and Seed-A coordinate masking traces this loss specifically to clearing sub-ULP FP32 residuals on not-yet-crossed weights (\S\ref{sec:unified}, Appendix~\ref{app:c2}, Tables~\ref{tab:c7}--\ref{tab:unified}). Even when sampled BF16 weights appear static over many steps, discarding FP32 master state interrupts the multi-step accumulation needed for diffuse updates to cross BF16 grid boundaries.
\end{enumerate}

\begin{table}[t]
  \centering
  \caption{Summary of diagnostic controls separating mode release from orthogonal refinement.}
  \label{tab:accounting}
  \footnotesize
  \begin{tabularx}{\textwidth}{@{}>{\raggedright\arraybackslash}p{2.5cm}>{\raggedright\arraybackslash}X>{\raggedright\arraybackslash}X>{\raggedright\arraybackslash}p{3.0cm}@{}}
    \toprule
    Diagnostic Axis & Core Operations \& Controls & Key Metrics & Mechanistic Question \\
    \midrule
    \textbf{Mode Release \& Native Template} & (i)~Prefill the target format and apply unsupervised BF16 steering along step-0 $u_0$ on the target prompt; (ii)~compare pre-RL base and RL under the same unsuppressed native template (e.g., \think{}). & Training-free gain $\Delta_{\mathrm{arb}}$ and truncation rate; unbiased \passk{K} and \cotpassk{K} under the matched native template. & How much existing behavior and coverage did the prompt suppress? How much apparent coverage growth disappears under the native template? \\
    \addlinespace[4pt]
    \textbf{BF16 Split \& Reallocation} & (i)~Decompose the BF16 update into $\CD$, $\Eperp$, and $\CD + \Eperp$ and compare against the shuffled-reward control $\Eperp^{\mathrm{shuf}}$; (ii)~run recovery and sharpening diagnostics. & Recovery from corrupted prefixes ($h^-$) and \texttt{no\_prefix} control; base-entropy calibrated $T^*$ retention; \texttt{mode\_wrong} and \texttt{0/32 base-correct} wrong-mode hit rates; matched majority-vote RL $\Delta\passk{K}$. & Do accuracy and recovery gains persist after removing $\CD$? How much gain survives $T^*$ calibration, and how do wrong-mode hit rates change? \\
    \bottomrule
  \end{tabularx}
\end{table}

\subsection{Limitations}
\label{sec:limitations}

\begin{itemize}
  \item \textbf{Single-seed diagnostics, training horizon, and cap decoupling:} Our coverage and sharpening diagnostics come from math benchmarks and a single \texttt{Qwen3-0.6B} run (Seed~A) evaluated under finite sampling budgets (\passk{256} and \cotpassk{192}). In addition, our main trajectories, native-\think{} RL, and majority-vote RL run for 100 steps, while coordinate-masked syncing covers steps 21--60; whether these patterns hold across additional seeds, multi-thousand-step horizons, or an 8k-cap training rerun (to cleanly decouple cap effects at step~20) remains to be tested.
  \item \textbf{Scale, architecture, and domain breadth:} Our Llama comparison (\S\ref{sec:puzzles}) pairs a 1B native model with an 8B distilled model. Matched-scale native 8B controls, models above 10B, and non-math RLVR domains such as code and tool use remain to be tested. It also remains open whether $\Eperp$ itself contains task-specific low-dimensional subspaces.
  \item \textbf{Training-time stochastic rounding and benchmark exposure:} Future work should directly test how training-time stochastic rounding of rollout weights \citep{gao2026bf16} affects Phase~II sub-ULP accumulation. Although our training and evaluation splits are hash-disjoint, the open pre-RL checkpoints may still have encountered public math benchmarks during pretraining \citep{wu2025contamination}.
\end{itemize}

\subsection{Conclusion}
\label{sec:conclusion}

Pre-RL Fisher geometry predicts the sign of early reasoning-RL length changes across diverse prompt and task settings on \texttt{Qwen3-0.6B}, while single-axis steering along the leading step-0 direction elicits analogous lengthening or format-tightening regimes across model families. The six-dimensional Fisher subspace $\Dsub$ captures a large share of the reproducible early reward gradient, and small concentrated shifts along $\Dsub$ strongly shift response length and accuracy, amplifying pre-existing thinking or formatting habits and reproducing the apparent Hard-50 valid-CoT jump over \nothink{}.

Separating mode release from orthogonal refinement requires operating directly in BF16 execution precision, avoiding the FP32 half-ULP dead-zone bias that otherwise masks orthogonal gains. Near step~20, the cross-problem shared reward gradient driving early lengthening drops sharply across seeds alongside rising 4k truncation; afterward, accuracy and corrupted-prefix recovery keep climbing even as response length stabilizes. Beyond Phase~I mode release, RL delivers two measurable behavioral improvements: stronger error-sensitive recovery from corrupted intermediate steps, and selective within-support probability reallocation that raises small-$k$ accuracy and cuts truncation while leaving matched-template aggregate 100-step large-$k$ coverage flat.

Finally, mixed-precision state governs late-stage optimization even when sampled BF16 weights appear static: periodic BF16 master-weight syncing degrades Phase~II accuracy across three seeds on both pooled blind and lockbox evaluations, driven by clearing sub-ULP FP32 residuals on not-yet-crossed weights. Together, these controls distinguish accuracy gains that release pre-existing habits from gains that persist after controlling the prompt, response length, and leading Fisher component.

\bibliographystyle{plainnat}
\bibliography{references}

\clearpage
\appendix
\numberwithin{table}{section}
\numberwithin{figure}{section}
\numberwithin{equation}{section}
\section{Theoretical Derivations and BF16 Half-ULP Dead Zone}
\label{app:a}

\subsection{First-Order Derivation of the Selection--Response Relation (Proposition~\ref{prop:sr})}
\label{app:a1}

\paragraph{Step 1: Fisher eigendecomposition and normalized scores.}
Let $G(x,y) = \nabla_\theta \log \pi_{\theta_0}(y \mid x)$ be the trajectory score of the pre-RL base policy $\pi_{\theta_0}$. It satisfies $\E[G \mid x] = 0$ and $\E[G] = 0$, and has trajectory Fisher information matrix $F_0 = \E[GG^\top]$. Since $F_0$ is symmetric positive semidefinite, its eigendecomposition reads
\begin{equation}
  F_0 v_k = \mu_k v_k, \qquad v_j^\top v_k = \delta_{jk}, \qquad \mu_0 \ge \mu_1 \ge \dots \ge 0.
\end{equation}
For each $\mu_k > 0$, define the normalized score random variable $\phi_k(x,y) = \frac{v_k^\top G(x,y)}{\sqrt{\mu_k}}$. Each $\phi_k$ has zero mean, and the collection $\{\phi_k\}_{\mu_k > 0}$ is orthonormal under expectation:
\begin{equation}
  \E[\phi_k] = 0, \qquad \E[\phi_j \phi_k] = \frac{1}{\sqrt{\mu_j \mu_k}} v_j^\top F_0 v_k = \delta_{jk}.
\end{equation}
Directions with $\mu_k = 0$ satisfy $\E[(v_k^\top G)^2] = 0$, hence $v_k^\top G = 0$ almost surely and $G = \sum_{\mu_k > 0} \sqrt{\mu_k}\,\phi_k\,v_k$.

\paragraph{Step 2: Orthogonal projection of behavior and reward.}
Let $z(x,y)$ be any behavioral statistic independent of $\theta$ (e.g., $\log L$) and $R(x,y)$ be the reward. Within-prompt centering gives $\tilde{z} = z - \E[z \mid x]$ and $\tilde{R} = R - \E[R \mid x]$. Because $\E[G \mid x] = 0$, centering leaves score projections and policy gradients unchanged and isolates within-prompt covariance. Under the inner product $\langle A, B \rangle = \E[AB]$, projecting onto $\spanop\{\phi_k\}_{\mu_k > 0}$ and its complement gives
\begin{equation}
  \tilde{z} = \sum_{\mu_k > 0} z_k \phi_k + z_{\mathrm{res}}, \qquad
  \tilde{R} = \sum_{\mu_k > 0} r_k \phi_k + R_{\mathrm{res}},
\end{equation}
where $z_k = \E[\tilde{z}\phi_k] = \mu_k^{-1/2} v_k^\top \E[\tilde{z}G]$, $r_k = \E[\tilde{R}\phi_k] = \mu_k^{-1/2} v_k^\top \E[\tilde{R}G]$, and $\E[z_{\mathrm{res}} G] = \E[R_{\mathrm{res}} G] = 0$. Expanding the within-prompt covariance yields
\begin{equation}
  \E_x[\Cov(z, R \mid x)] = \E[\tilde{z}\tilde{R}] = \sum_{\mu_k > 0} z_k r_k + \langle z_{\mathrm{res}}, R_{\mathrm{res}} \rangle.
\end{equation}

\paragraph{Step 3: First-order response under policy-gradient flow.}
By the log-derivative identity and $\E[G \mid x] = 0$,
\begin{align}
  g_z &= \left.\nabla_\theta \E[z]\right|_{\theta_0} = \E[\tilde{z}G] = \sum_{\mu_k > 0} \sqrt{\mu_k}\,z_k\,v_k, \nonumber \\
  g_R &= \left.\nabla_\theta \E[R]\right|_{\theta_0} = \E[\tilde{R}G] = \sum_{\mu_k > 0} \sqrt{\mu_k}\,r_k\,v_k.
\end{align}
Residuals orthogonal to $G$ vanish from both gradients. Under ideal Euclidean flow $\dot{\theta} = g_R$,
\begin{equation}
  \left.\frac{\mathrm{d}\E[z]}{\mathrm{d}\tau}\right|_{\theta_0} = g_z^\top g_R = \sum_{\mu_k > 0} \mu_k z_k r_k.
\end{equation}
Now replace $g_R$ with the damped natural gradient $v_\lambda = (F_0 + \lambda I)^{-1} g_R = \sum_{\mu_k > 0} \frac{\sqrt{\mu_k}}{\mu_k + \lambda} r_k v_k$ ($\lambda = 0.001\mu_0$). This reweights the per-mode Fisher factors sent into the update to $\frac{\mu_k}{\mu_k + \lambda} \in (0,1)$, removing top-eigenvalue amplification; Table~\ref{tab:b3} compares $g_R$ and $v_\lambda$ at matched Fisher norm $\sqrt{v^\top F_0 v}$.

\paragraph{Step 4: Forward sensitivity under AdamW.}
Under coordinate-wise normalized AdamW updates $d_{\mathrm{AdamW}} = \hat{m}/(\sqrt{\hat{v}} + \epsilon)$, the optimizer rescales components along the update direction, yielding the first-order behavioral change
\begin{equation}
  g_z^\top d_{\mathrm{AdamW}} = \sum_{\mu_k > 0} \sqrt{\mu_k}\,z_k\,\bigl(v_k^\top d_{\mathrm{AdamW}}\bigr).
\end{equation}
Although second-moment scaling lowers parameter-space alignment with $u_0$ by ${\sim}5\times$ (Table~\ref{tab:c6}, where $-m$ and $-d$ correspond to the reward-ascent direction $d_{\mathrm{AdamW}}$ under loss-descent sign convention), the forward-sensitivity factor $\sqrt{\mu_k}$ inside $g_z$ is unaffected by optimizer preconditioning, explaining why leading Fisher directions retain measurable length leverage even when single-step parameter alignment is small.

\subsection{BF16 Half-ULP Dead Zone and Closed-Loop Validation (Proposition~\ref{prop:deadzone})}
\label{app:a2}

Let $Q(\cdot)$ denote element-wise IEEE~754 round-to-nearest-even BF16 rounding, let $\theta_a$ lie on the BF16 grid, and let $u$ be a unit sensitive direction. For each grid anchor $\theta_{a,i}$ with adjacent representable BF16 values $\theta_{a,i}^- < \theta_{a,i} < \theta_{a,i}^+$, any perturbed value strictly inside the midpoint interval $\bigl(\tfrac{1}{2}(\theta_{a,i}^- + \theta_{a,i}),\, \tfrac{1}{2}(\theta_{a,i} + \theta_{a,i}^+)\bigr)$ rounds back to $\theta_{a,i}$, with ties at either midpoint broken to the adjacent BF16 value having an even least-significant significand bit. Away from powers of two, $\theta_{a,i} - \theta_{a,i}^- = \theta_{a,i}^+ - \theta_{a,i} = \mathrm{ULP}(\theta_{a,i})$, so this rounding cell is $\pm\tfrac{1}{2}\mathrm{ULP}$; at a power-of-two boundary (e.g., $\theta_{a,i} = 1$), the lower spacing ($2^{-8}$) is half the upper spacing ($2^{-7}$), so the cell is asymmetric. Because Proposition~\ref{prop:deadzone} defines the active set $\mathcal{L} = \{i : Q(\theta_{a,i} + \delta'_i) \ne \theta_{a,i}\}$ directly through the rounding map $Q$, Equation~\eqref{eq:deadzone} holds exactly everywhere. For an FP32 natural displacement $\Delta$, define its FP32 and BF16 execution projections along $u$ as $c = u^\top \Delta \ne 0$ and $\cexec = u^\top [Q(\theta_a + \Delta) - \theta_a]$. Subtracting $cu$ in FP32 gives $\delta'_i = \Delta_i - c u_i$. After BF16 rounding, execution weights change only on $\mathcal{L}$, while coordinates inside the rounding cell map back to $\theta_{a,i}$. Define the signed projection ratio $f := \frac{1}{c}\sum_{i \in \mathcal{L}} u_i \Delta_i \in \mathbb{R}$ (which need not lie in $[0,1]$), the active energy $\mL := \sum_{i \in \mathcal{L}} u_i^2 \in [0, 1]$, and the projected rounding remainder $\varepsilon_Q := \sum_{i \in \mathcal{L}} u_i\bigl[Q(\theta_{a,i} + \delta'_i) - (\theta_{a,i} + \delta'_i)\bigr]$. Expanding $\bexec$ over $\mathcal{L}$ then yields the exact identity
\begin{equation}
  \bexec = \sum_{i \in \mathcal{L}} u_i\bigl(Q(\theta_{a,i} + \delta'_i) - \theta_{a,i}\bigr)
  = \sum_{i \in \mathcal{L}} u_i(\Delta_i - c u_i) + \varepsilon_Q
  = (f - \mL)\,c + \varepsilon_Q.
\end{equation}
Whenever $f < \mL$ and $|\varepsilon_Q| < (\mL - f)|c|$, $\bexec$ reverses sign relative to $c$. We report the offset magnitude relative to two complementary denominators:
\begin{itemize}
  \item \textbf{Relative to FP32 projection $c$ (size of the subtraction bias):} $\bexec / c = (f - \mL) + \varepsilon_Q/c$. On \texttt{Qwen3-0.6B} at step~40 ($f \approx 0.64\text{--}0.65$, $\mL \to 1$), the measured bias is $-30\%$ to $-44\%$. On \texttt{Qwen3-8B} at step~60 ($f = 0.152\text{--}0.162$, $\mL = 0.731\text{--}0.782$), it is $-57\%$ to $-63\%$.
  \item \textbf{Relative to BF16 execution projection $\cexec$ (why the bias overshoots the executed target):} On \texttt{Qwen3-8B}, only $6.72\%$ of weights cross a BF16 step, shrinking $\cexec$ to ${\sim}20\%$ of $c$. Relative to this smaller denominator, $\bexec / \cexec$ reaches a $\mathbf{-2.71\times}$ reversed overshoot (Table~\ref{tab:d1}).
\end{itemize}
Table~\ref{tab:a1} reports the offset-injection and misaligned-grid re-quantization tests shown in Figure~\ref{fig:deadzone}b.

\begin{table}[htbp]
  \centering
  \caption{Closed-loop verification of the BF16 half-ULP dead zone (MATH dev, 1,024 problems, 8k cap, 3 seeds; \S\ref{sec:deadzone}, Figure~\ref{fig:deadzone}b).}
  \label{tab:a1}
  \footnotesize
  \resizebox{\textwidth}{!}{%
  \begin{tabular}{@{}lccl@{}}
    \toprule
    Intervention & Accuracy (Seeds A / B / C) & Mean & Mechanism \\
    \midrule
    Base $\theta_0$ on misaligned grids (\texttt{int8c}/\texttt{g128}/\texttt{fp8t}) & $44.34\% / 45.56\% / 44.24\%$ & --- & Neg.\ control (BF16 base $45.53\%$) \\
    BF16 zeroing within tolerance ($d=0.0$) & $53.8\% / 53.2\% / 54.4\%$ & $\mathbf{53.8\%}$ & Unbiased zeroing baseline \\
    Inject $0.5\times$ reversed offset ($d=+0.5$) & $50.3\% / 48.4\% / 48.6\%$ & $49.1\%$ & Accuracy decreases over tested doses \\
    Inject $1.0\times$ reversed offset ($d=+1.0$) & $45.7\% / 44.0\% / 44.5\%$ & $44.7\%$ & Reproduces FP32-only collapse \\
    FP32-only locking exported to BF16 & $46.6\% / 45.4\% / 45.1\%$ & $\mathbf{45.7\%}$ & Carries $-30\%$ to $-44\%$ offset \\
    Same FP32 weights $\to$ misaligned INT8 (per-ch.) & $55.1\% / 52.8\% / 55.3\%$ & $\mathbf{54.4\%}$ & Recovers $7.4\text{--}10.2$ pp vs.\ FP32-only \\
    Same FP32 weights $\to$ misaligned INT8 (group-128) & $54.8\% / 53.1\% / 54.9\%$ & $\mathbf{54.3\%}$ & Matches BF16 zeroing baseline \\
    Same FP32 weights $\to$ misaligned FP8-E4M3 & $52.5\% / 52.3\% / 52.6\%$ & $\mathbf{52.5\%}$ & Recovers $5.9\text{--}7.5$ pp \\
    \bottomrule
  \end{tabular}}
\end{table}

\clearpage
\section{Experimental Protocol, Notation, and Supporting Data for \S\ref{sec:geometry}--\S\ref{sec:refinement}}
\label{app:b}

\subsection{Notation, Orthogonalization, BF16 Placement Audit, and Dataset Splits}
\label{app:b1}

\begin{table}[htbp]
  \centering
  \caption{Notation and symbol definitions.}
  \label{tab:b1}
  \footnotesize
  \begin{tabularx}{\textwidth}{@{}>{\raggedright\arraybackslash}p{3.0cm}X@{}}
    \toprule
    Symbol & Definition \\
    \midrule
    $\theta_0$, $Q(\theta)$ & Pre-RL base weights; element-wise round-to-nearest BF16 execution weights of FP32 master $\theta$. \\
    $\mathrm{ULP}$, half-ULP, sub-ULP & Unit in the last place (local spacing between adjacent representable BF16 values); half-ULP dead zone is the rounding cell bounded by midpoints to the lower and upper adjacent BF16 values ($\pm\tfrac{1}{2}\mathrm{ULP}$ away from powers of two, with tie-to-even; \S\ref{sec:deadzone}, App.~\ref{app:a2}); sub-ULP updates are smaller than one local BF16 grid step (\S\ref{sec:unified}). \\
    $G$, $F_0$, $F_t$, $\mu_k$, $u_k$ & Trajectory score $\nabla_\theta \log \pi_{\theta_0}(y \mid x)$, step-0 / step-$t$ trajectory Fisher information matrices, Fisher eigenvalues, and orthonormal basis ($u_0$ leading). \\
    $g_R$, $g_z$, $g_L$, $\widehat{s}_R$, $\eta_S(g)$ & Policy gradients of expected reward $R$, statistic $z$, and log-length; cross-subset reproducible reward-gradient energy $\widehat{s}_R$ (\S\ref{sec:selection-response}, \S\ref{sec:step20}); energy fraction $\eta_S(g) = \|P_S g\|_2^2 / \|g\|_2^2$ of gradient $g$ in subspace or direction $S$ (Appendix~\ref{app:c1}). \\
    $\Dsub$, $U$ & Unsupervised 6D step-0 sensitive subspace spanned by orthonormal basis $U = [u_0,\dots,u_5]$ ($U^\top U = I_6$). \\
    $\Dexec_{a\to b}$, $\CD$, $\Eperp$ & BF16 execution displacement $Q(\theta_b) - Q(\theta_a)$, algebraic $\Dsub$ projection $UU^\top\Dexec$, and orthogonal complement $(I - UU^\top)\Dexec$ (distinguished from their BF16 realizations in \S\ref{sec:deadzone}). \\
    $\bexec$, $\cexec$, $f$, $\mL$ & BF16 executed projection after FP32 subtraction ($u^\top[Q(\theta_a+\delta')-Q(\theta_a)]$), unprojected BF16 coordinate ($u^\top\Dexec$), active-set signed projection ratio, and active-set $u_i^2$-energy (\S\ref{sec:deadzone}). \\
    Steering \& dose $\alpha$ & Training-free BF16 displacement of dose $\alpha$ along a chosen axis via Equation~\eqref{eq:placer}. Unit dose $\alpha=1$ is calibrated to $s = 0.0248$ (reference step-40 coordinate in \texttt{0.6B}); positive doses lengthen responses along $u_0$ and format axes. \\
    $h^+$, $h^-$ & Clean-prefix accuracy and corrupted-prefix recovery rate in the 193-pair probe (\S\ref{sec:step20}, \S\ref{sec:recovery}); $h^+ - h^-$ is the accuracy drop caused by corrupting the prefix. \\
    \passk{k}, \cotpassk{K}, $\HTF$, $T^*$ & Unbiased \passk{k} (probability of $\ge 1$ correct answer in $k$ samples) and judge-valid \cotpassk{K} (\S\ref{sec:coverage}); mean full-vocabulary teacher-forcing entropy $\HTF$ and calibrated temperature $T^*$ restoring $\HTF$ to at least the pre-RL threshold (\S\ref{sec:sharpening}). \\
    \bottomrule
  \end{tabularx}
\end{table}

\paragraph{Subspace orthogonalization.}
Let $a_0, a_1, a_2$ and $b_0, b_1, b_2$ be the top-3 Lanczos eigenvectors from disjoint dev halves A and B. We retain the three half-A vectors directly as $u_k = a_k$ ($k=0,1,2$). For $j=0,1,2$, we then orthogonalize $b_j$ sequentially via $\tilde{b}_j = b_j - \sum_{k=0}^{2+j}(u_k^\top b_j)u_k$ and normalize $u_{3+j} = \tilde{b}_j / \|\tilde{b}_j\|_2$, yielding an orthonormal basis $U^\top U = I_6$.

\paragraph{Dose calibration, per-arm BF16 spillover audit, and online locking.}
To steer along $u_k$ at dose $\alpha$, we initialize Equation~\eqref{eq:placer} at $d^{(0)} = 0$ with target $b^\star = \alpha s e_k$. On \texttt{Qwen3-0.6B}, the unit scale is $s = 0.0248$, matching the reference seed's step-40 BF16 coordinate ($0.02480$); the other two seeds reach $0.0196$ and $0.0166$. We run up to 12 iterations with a steering tolerance of $10^{-5}$.
\begin{itemize}
  \item \textbf{Coordinate residuals and recombination fidelity:} Across all arms, the maximum $\Dsub$ read-back residual is $< 1.84\times 10^{-4}$ ($< 8\times 10^{-5}$ for $\Eperp$-only arms; $5.5\times 10^{-6}$ on Seed~C). Small $\Dsub$ residuals can coexist with nonzero full-parameter discrepancy: on $0\to 40$, the full-parameter $\ell_2$ discrepancy between recombined $\CD + \Eperp$ and the original checkpoint is $0.00093$ on Seed~C (earlier builds: $0.003\text{--}0.004$, ${\sim}0.005\%\text{--}0.006\%$ differing weights), and $1.2\text{--}1.5\times 10^{-5}$ on $20\to 100$.
  \item \textbf{Per-arm discrete grid spillover and output impact:} To verify that BF16 grid spillover does not distort our decomposition, we audit per-arm norm errors and conditional KL shifts. Quantizing a continuous placement onto BF16 leaves an off-target Euclidean component whose norm varies by arm: on Seed~C ($0\to 40$, ideal $\|\CD\|_2 = 0.0221$, $\|\Eperp\|_2 = 0.4867$), the realized $\CD$-only arm has in-subspace norm $0.0221$ and off-subspace spillover norm $0.0130$ (Seeds~A/B: $0.0150\text{--}0.0168$ off-subspace vs.\ $0.0262\text{--}0.0301$ in-subspace); the realized $\Eperp$-only arm has $\Dsub$ residual $5.5\times 10^{-6}$ and target discrepancy $0.0130$ on Seed~C ($0.0643\text{--}0.0698$, or $13\%\text{--}14\%$ of target norm, on Seeds~A/B; $0.008\text{--}0.011$ on $20\to 100$); the recombined arm has target error $0.00093$. Crucially, because this spillover spreads across $5.96\times 10^8$ low-curvature parameters, it induces only $0.7\%$ of the total fixed-prefix conditional KL shift, and unbiased stochastic rounding at $+1.0\,u_0$ closely matches round-to-nearest placement ($59.8\%$ accuracy and $2{,}374$ tokens vs.\ $59.6\%$ and $2{,}438$ tokens), confirming that grid spillover does not drive our split results.
  \item \textbf{Online BF16 $\Dsub$-locking:} After each standard AdamW step, we leave the first and second moments $m, v$ unchanged and apply up to 8 iterations of Equation~\eqref{eq:placer} directly to the FP32 master weights along $U$, stopping once $\max_j |u_j^\top [Q(\theta) - Q(\theta_0)]| < 10^{-4}$.
  \item \textbf{Learning rates and base-model accuracy:} GRPO uses $\mathrm{lr}=4\times 10^{-6}$ for \texttt{Qwen3-0.6B} and $2\times 10^{-6}$ for \texttt{Llama-3.2-1B-Instruct}. Base \texttt{0.6B} \nothink{} accuracy is $45.5\%$ (dev 8k), $45.8\%$ (dev 4k), and $43.7\%$ (Seed-A held-out 8k). Dataset splits appear in Table~\ref{tab:b2}.
\end{itemize}

\begin{table}[htbp]
  \centering
  \caption{Dataset splits, sampling budgets, token caps, and experimental roles.}
  \label{tab:b2}
  \scriptsize
  \setlength{\tabcolsep}{3.5pt}
  \begin{tabularx}{\textwidth}{@{}l>{\raggedright\arraybackslash}p{3.0cm}c>{\raggedright\arraybackslash}X@{}}
    \toprule
    Task / Split & Problems $\times$ Samples & Cap (tok) & Role \\
    \midrule
    MATH Train & 6,400 & 4,096 & Main GRPO \& majority-vote RL (5,734 valid-mode prompts); hash-disjoint from all eval sets. \\
    MATH Dev (\nothink{}) & $1{,}024 \times 4$ (eval); $1{,}024 \times 16$ ($\Dsub$); $512 \times 16$ (grad./norm.) & 8,192 / 4,096 & Eval/steering ($1{,}024 \times 4$, 8k); $\Dsub$ extraction \& damped nat.\ grad.\ ($512 \times 16$ per parity half, 8k); gradient trajectories \& B.2 norm.\ audit (first $512 \times 16$, 4k). \\
    MATH Native \think{} (\texttt{0.6B}) & $1{,}024$ dev; $250$ L4--5 $\times 256$ & 16,384 & Control~1 (Table~\ref{tab:controls}) \& matched large-$k$ \passk{256}/\cotpassk{192} (Table~\ref{tab:passk}). \\
    MATH Held-out & $512 \times 4$ ($\times 64$; L4--5 $\times 256$) & 8,192 & Locked training, $2\times 2$ splits, shuffled reward, 8B/1.5B eval, large-$k$, sharpening (disjoint from train/dev/lockbox). \\
    MATH Pooled Blind (Held-out $+$ Lockbox) & $1{,}536$ $(512 + 1{,}024) \times 4$ & 8,192 & Phase~II BF16 sync ablation pooling held-out ($512$) and disjoint lockbox ($1{,}024$), with lockbox-only metrics in Table~\ref{tab:c7}; first 128 lockbox problems used only to estimate the R1 axis. \\
    MATH Direct (\texttt{0.6B}) & 6,400 train / $1{,}024 \times 4$ dev & 32 & Prompt switch (Table~\ref{tab:pred}, Figure~\ref{fig:causal}b), format steering (Table~\ref{tab:quadrants}), format control (Table~\ref{tab:b13}). \\
    Fixed TF Pool (\texttt{0.6B}) & 8,192 trajectories & Full vocab & TF entropy, calibrated $T^*$, regressions, \texttt{Hard-Flip} (Tables~\ref{tab:sharpening}, \ref{tab:b16a}--\ref{tab:b16b}). \\
    \texttt{Qwen3-8B} MATH Steering & 682 extract / $342 \times 4$ eval & 8,192 & Bidirectional $u_0$ steering (Table~\ref{tab:quadrants}, Figure~\ref{fig:b2}, Table~\ref{tab:b7}); 60-step RL evaluated on full dev and held-out. \\
    \texttt{Llama-3.2-1B-Instruct} MATH & $1{,}024 \times 4$ dev & 8,192 & Format steering, saturation test, equal-LR control (Figure~\ref{fig:quadrants}, Table~\ref{tab:quadrants}, Table~\ref{tab:d2}). \\
    \texttt{R1-Distill-Llama-8B} MATH & 128 extract / $512 \times 4$ held-out & 16,384 & Lengthening steering under \nothink{} \& random-direction control (Figure~\ref{fig:quadrants}, Table~\ref{tab:quadrants}). \\
    GSM8K (\texttt{8B} mental / \texttt{0.6B}) & 7,473 / $1{,}319 \times 4$ & 64 / 4,096 & \texttt{8B} \texttt{"\#\#\#\# "} mental-math control (Table~\ref{tab:b14}) \& \texttt{0.6B} transfer (Table~\ref{tab:b8}). \\
    Knights \& Knaves / MMLU / \texttt{wiki2} & 1,024 / $1{,}000\times 4$ / test & 4,096 / 2,048 & Logic length forecast (Table~\ref{tab:pred}) / cross-task transfer \& perplexity (Table~\ref{tab:b8}). \\
    \bottomrule
  \end{tabularx}
\end{table}

\subsection{Supporting Data for \S\ref{sec:geometry}--\S\ref{sec:arbitrage}: Damped Natural Gradient, Axis Fingerprints, and Cross-Model Sweeps}
\label{app:b2}

\paragraph{Cross-half stability and length-normalized Lanczos (8B scale and 0.6B invariance check).}
On \texttt{Qwen3-0.6B}, the original $\Dsub$ basis is extracted from the two disjoint parity halves of the 1,024 dev prompts ($512 \times 16 = 8{,}192$ pre-RL rollouts per half, 8k cap; Table~\ref{tab:b2}). In $5.96 \times 10^8$ parameter dimensions, random unit vectors have inner-product standard deviation ${\sim}4\times 10^{-5}$; by contrast, the top-3 Lanczos Ritz vectors estimated independently on the two disjoint halves (Ritz values $1.7\text{--}2.3\times 10^5$) align with parameter-space inner products of $0.57\text{--}0.73$ (dropping to noise from the fourth vector onward) and induce nearly identical token log-probability shifts (cosine similarities $0.98\text{--}0.99$). At the 8B scale, a single extreme rollout can dominate $F_0$ (for example, one ${\sim}15$k-token sample on \texttt{R1-Distill-Llama-8B} captures $81\%$ of the Rayleigh quotient). To prevent outlier trajectories from dictating the leading direction, we normalize per-sample directional derivatives by $\sqrt{\text{tokens}}$, capping any single sample's weight below $1\%$. Under this normalization, \texttt{Qwen3-8B} uses 6 Lanczos steps on 682 prompts (Ritz value $117.9$, residual $1.5\%$), and \texttt{R1-Distill-Llama-8B} uses 4 steps on 128 prompts (Ritz value $533.8$, residual $1.2\%$). To verify that the unnormalized step-0 trajectory Fisher on \texttt{Qwen3-0.6B} (which matches the sequence-level KL metric of Proposition~\ref{prop:sr}) is not driven by long-sequence outliers or length weighting, we audit the 8,192 pre-RL rollouts on the first 512 dev prompts ($512 \times 16$, 4k cap, shared with the step-0 gradient baseline in Appendix~\ref{app:c1}): on raw $F_0$, the single largest sample accounts for only $0.270\%$ of the Rayleigh quotient along $u_0$ (top 10 samples combined: $2.23\%$). Running independent random-start block Lanczos ($B=2$, 6 steps) on \texttt{Qwen3-0.6B} with $1/\sqrt{L}$ normalization (top-3 Ritz values $542.1, 485.0, 338.1$; residuals $0.38\%, 0.23\%, 2.65\%$; max single-sample share $0.364\%$) yields a top-3 subspace with principal cosines $[0.969, 0.961, 0.929]$ against $\Dsub$, capturing $89.5\%$ of original $u_0$'s squared energy (subspace cosine $0.946$; individual cosines $|\cos(u_0, u_0^{\mathrm{sqrt}})| = 0.573$ and $|\cos(u_0, u_1^{\mathrm{sqrt}})| = 0.721$ reflect rotation within the closely spaced top-2 plane, while $89.7\%$, $93.7\%$, and $89.2\%$ of the top-3 normalized Ritz vectors' energies lie in $\Dsub$). Even stronger $1/L$ per-token-mean normalization (top-3 Ritz values $2.40, 1.78, 1.12$; residuals $0.13\%, 0.09\%, 1.89\%$; max share $0.696\%$) retains principal cosines $[0.940, 0.914, 0.844]$ against $\Dsub$ and captures $81.2\%$ of $u_0$'s energy (subspace cosine $0.901$; $|\cos(u_0, u_0^{\mathrm{tok}})| = 0.438$, $|\cos(u_0, u_1^{\mathrm{tok}})| = 0.743$). Thus, $\Dsub$ is a genuine pre-RL geometric subspace rather than an artifact of sequence-length weighting.

\paragraph{Sign convention and unit doses.}
Main axes $u_0$ and format axes are signed so that positive doses lengthen outputs. Unit dose $\alpha=1$ equals $0.0248$ on \texttt{0.6B} and $0.0154$ on \texttt{Llama-3.2-1B}. On \texttt{0.6B}, dose $-1$ loses $16.7$ pp, whereas over-steering to dose $+2$ lowers accuracy to $36.7\%$ at $60\%$ truncation.

\paragraph{Native versus distilled Llama-family sweeps.}
To separate architecture family from pre-existing reasoning habits, we compare bidirectional sweeps on native \texttt{Llama-3.2-1B} and distilled \texttt{R1-Distill-Llama-8B}, confirming opposite regimes within the same architecture family. Native \texttt{Llama-3.2-1B} exhibits a step-0 top Fisher eigenvalue of $\mu_0 = 2.7\times 10^5$, on the same scale as \texttt{Qwen3-0.6B} ($1.7\text{--}2.3\times 10^5$), yet lacking latent long CoT, it gains only by tightening format: dose $-1$ shortens responses ($752 \to 573$ tokens), raises box rate ($52.1\% \to 59.8\%$), and lifts accuracy ($11.3\% \to 14.8\%$), whereas positive dose $+1$ stretches outputs to $3{,}043$ tokens while breaking \boxedtok{} compliance ($27.0\%$ box rate, $3.5\%$ accuracy); by comparison, the shuffled-direction control stays flat ($11.6\%$) while GRPO saturates format ($508$ tokens, $99.8\%$ box rate). By contrast, distilled \texttt{R1-Distill-Llama-8B} carries latent long-CoT habits over $[-0.08, +0.08]$: moderate positive dose $+0.04$ lengthens outputs to $1{,}849$ tokens, boosts reflection words ($3.63 \to 8.97$), re-opens thinking blocks ($3.0\% \to 23.3\%$), and lifts accuracy to $81.30\%$ (paired 95\% CI $+8.35\,[+6.19, +10.50]$ pp), whereas negative doses $-0.04$ and $-0.08$ suppress derivation and collapse accuracy to $46.73\%$ and $10.55\%$. Matched $\pm 0.08$ random controls shift accuracy by only $0.83\text{--}1.12$ pp, ruling out generic perturbations and confirming that steering unlocks whichever habit the base model already holds.

\begin{table}[htbp]
  \centering
  \caption{First-order step-0 behavioral responses before and after damped natural-gradient preconditioning (disjoint dev halves A/B, KL-normalized; \S\ref{sec:selection-response}, Figure~\ref{fig:causal}a).}
  \label{tab:b3}
  \footnotesize
  \setlength{\tabcolsep}{4.5pt}
  \begin{tabular}{@{}lccl@{}}
    \toprule
    Statistic & Raw Policy Gradient (A / B) & Damped Natural Gradient (A / B) & Relative Change \\
    \midrule
    Log response length & $+0.075$ / $+0.093$ & $+0.010$ / $+0.014$ & Reduced $\mathbf{85\%\text{--}87\%}$ \\
    Reflection tokens (\texttt{wait}/\texttt{but}) & $+0.046$ / $+0.035$ & $-0.003$ / $-0.007$ & Growth eliminated \\
    Expected reward (accuracy) & $+0.030$ / $+0.026$ & $+0.026$ / $+0.025$ & Retained $\mathbf{87\%\text{--}96\%}$ \\
    \bottomrule
  \end{tabular}
\end{table}

\begin{table}[htbp]
  \centering
  \caption{Exact step-1 GRPO update replay on \texttt{Qwen3-0.6B} MATH \nothink{} (4 disjoint $64\times 8$ batches, 2,048 rollouts; \S\ref{sec:selection-response}). The pooled standardized length gradient has norm $\|g_L\|_2 = 0.3563$, with $5.52\%$ of its energy along $u_0$ and $17.52\%$ in $\Dsub$. On step~1, $4.74\%$ of parameters cross a BF16 boundary, and AdamW sign-normalization plus BF16 rounding reduce the single-step $\Dsub$ squared-norm share $\|P_{\Dsub}v\|_2^2 / \|v\|_2^2$ from $2.926\%$ to $0.082\%$ and then to $0.0115\%$. At all three stages on step~1, the negative $\Eperp$ length term exceeds the positive $\Dsub$ term in magnitude, leaving a small negative net first-order response. Thus, the two subspaces already act with opposite signs on step~1 ($+1.95\times 10^{-5}$ from $\Dsub$ carrying $0.0115\%$ of squared execution norm vs.\ $-3.69\times 10^{-5}$ from $\Eperp$), while coherent $\Dsub$ updates accumulate over steps 1--20 (Table~\ref{tab:c6}).}
  \label{tab:b3b}
  \scriptsize
  \setlength{\tabcolsep}{2.2pt}
  \begin{tabular}{@{}lcccccc@{}}
    \toprule
    Step-1 Update $v$ & $\|v\|_2$ & $\|P_{\Dsub}v\|_2^2 / \|v\|_2^2$ & $g_L^\top P_{u_0} v$ & $g_L^\top P_{\Dsub} v$ & $g_L^\top P_{\Eperp} v$ & $g_L^\top v$ \\
    \midrule
    Raw GRPO $g_{\mathrm{GRPO}}$ & $2.28\mathrm{e}{-1}$ & $2.926\%$ & $+9.74\mathrm{e}{-4}$ & $+2.48\mathrm{e}{-3}$ & $-2.68\mathrm{e}{-3}$ & $-2.00\mathrm{e}{-4}$ \\
    AdamW FP32 $\Delta^{\mathrm{fp32}}$ & $8.31\mathrm{e}{-2}$ & $0.082\%$ & $+6.18\mathrm{e}{-5}$ & $+1.28\mathrm{e}{-4}$ & $-2.93\mathrm{e}{-4}$ & $-1.65\mathrm{e}{-4}$ \\
    \textbf{Exec BF16 $\bm{\Delta^{\mathrm{exec}}}$} & $\mathbf{3.17\mathrm{e}{-2}}$ & $\mathbf{0.0115\%}$ & $\mathbf{+9.04\mathrm{e}{-6}}$ & $\mathbf{+1.95\mathrm{e}{-5}}$ & $\mathbf{-3.69\mathrm{e}{-5}}$ & $-1.74\mathrm{e}{-5}$ \\
    \bottomrule
  \end{tabular}
\end{table}

\begin{table}[htbp]
  \centering
  \caption{Token counts before and after the first \boxedtok{} under weight steering and decoding \texttt{EOS} suppression (\texttt{0.6B} dev, 4,096-token cap; \S\ref{sec:anatomy}, Figure~\ref{fig:anatomy}c). Weight steering ($u_0$, $\CD$) adds tokens before the first \boxedtok{} and raises accuracy, whereas decoding \texttt{EOS} suppression adds repetitive text afterward with no accuracy gain. Mean total length includes unboxed truncated samples (${\sim}14\%\text{--}16\%$ in rows 3--4).}
  \label{tab:b4}
  \footnotesize
  \begin{tabular}{@{}lccccc@{}}
    \toprule
    Base Intervention & Accuracy & Mean Total & Pre-\boxedtok{} & Post-\boxedtok{} & Boxes / Resp. \\
    \midrule
    Pre-RL base & $45.8\%$ & 589 & 563 & 26 & 1.20 \\
    Random / shuffled direction & $45.7\%$ & 585 & 560 & 25 & 1.20 \\
    Steer along $u_0$ ($+1$ dose) & $56.8\%$ (8k: $59.6\%$) & 1,812 & $\mathbf{1{,}211}$ & $\mathbf{21}$ & 0.86 \\
    Steer along $\CD$ (unit dose) & $57.8\%$ (8k: $61.2\%$) & 1,760 & $\mathbf{1{,}162}$ & $\mathbf{20}$ & 0.84 \\
    Decode \texttt{EOS} logit $-8$ & $44.8\%$ & 1,347 & 621 & $\mathbf{725}$ & 4.44 \\
    Decode \texttt{EOS} logit $-12$ & $44.9\%$ & 3,680 & 809 & $\mathbf{2{,}871}$ & 11.69 \\
    \bottomrule
  \end{tabular}
\end{table}

\begin{table}[htbp]
  \centering
  \caption{Effects of steering at doses $-1$ and $+1$ along each of the six $\Dsub$ axes on \texttt{Qwen3-0.6B} (dev 8k cap; Figure~\ref{fig:anatomy}a--b), showing functional specialization: only the three think-more axes ($u_0, u_3, u_4$) improve accuracy, whereas $u_1$ (formatting) and $u_2$ (early stopping) lower it. At two decimals, $+1\,u_0$ raises accuracy from $45.53\%$ to $59.57\%$, a gain of $14.04$ pp (reported as $14.0$ pp at one decimal throughout). $u_2$ retains its orthonormal basis orientation.}
  \label{tab:b5}
  \footnotesize
  \begin{tabular}{@{}lccl@{}}
    \toprule
    Axis & $+1$ Dose Acc / Len & $-1$ Dose Acc / Len & Primary Functional Role \\
    \midrule
    $u_0$ & $59.6\%$ / 2,438 & $28.8\%$ / 210 & Pre-box derivation; positive displacement in all RL seeds \\
    $u_3$ & $49.4\%$ / 1,045 & $33.5\%$ / 419 & Extended thinking; uses more reflection words \\
    $u_4$ & $46.1\%$ / 1,146 & $31.2\%$ / 416 & Extended thinking \\
    $u_5$ & $44.7\%$ / 662 & $39.7\%$ / 522 & Local paragraphing adjustment \\
    $u_1$ & $40.8\%$ / 743 & $41.2\%$ / 443 & More headings and Markdown formatting; lowers accuracy \\
    $u_2$ & $36.5\%$ / 305 & $43.9\%$ / 1,011 & Post-\boxedtok{} early stopping (shortens at $+1$) \\
    \bottomrule
  \end{tabular}
\end{table}

\begin{table}[htbp]
  \centering
  \caption{Directional Fisher information ($v^\top F_{0,s} v$, with $s$ indexing the prompt-and-budget setting; top) and cross-prompt steering (bottom; \S\ref{sec:selection-response}, Figure~\ref{fig:causal}b, Figure~\ref{fig:quadrants}). Switching from 8k CoT to 32-token direct answering shifts dominant directional Fisher information from the CoT axis ($8\times$) to the format axis ($60\times$), matching the accompanying reversal in steering and GRPO length trends.}
  \label{tab:b6}
  \footnotesize
  \resizebox{\textwidth}{!}{%
  \begin{tabular}{@{}llll@{}}
    \toprule
    Experiment & Setting $s$ & Measurement & Takeaway \\
    \midrule
    \textbf{Fisher ($v^\top F_{0,s} v$)} & MATH CoT prompt (8k) & CoT $1.87\times 10^5$ vs.\ Format $2.41\times 10^4$ & CoT axis $\mathbf{8\times}$ larger \\
    \textbf{Fisher ($v^\top F_{0,s} v$)} & MATH Direct prompt (32 tok) & Format $8.62\times 10^5$ vs.\ CoT $1.39\times 10^4$ & Format axis $\mathbf{60\times}$ larger \\
    \textbf{Cross-steer} & CoT $u_0$ on Direct prompt & Len: neg $17.3$ / base $27.7$ / pos $30.9$ & Neg dose: box $24\%\to 69\%$, acc $3.6\%\to 6.6\%$ \\
    \textbf{Cross-steer} & Format axis on CoT prompt & Len: neg $490$ / base $580$ / pos $647$ & Box $\ge 99.8\%$; length increases monotonically with dose \\
    \bottomrule
  \end{tabular}}
\end{table}

\begin{table}[htbp]
  \centering
  \caption{Bidirectional BF16 steering along step-0 $u_0$ on \texttt{Qwen3-8B} ($342$ held-out dev problems $\times 4$, 8k cap; Table~\ref{tab:b2}, Figure~\ref{fig:b2}), confirming monotonic lengthening and accuracy gains inside the positive band before breakdown at extreme negative doses.}
  \label{tab:b7}
  \scriptsize
  \setlength{\tabcolsep}{3.5pt}
  \begin{tabular}{@{}lccccccccc@{}}
    \toprule
    BF16 Dose & $-0.10$ & $-0.06$ & $-0.05$ & $-0.03$ & $0$ (Base) & $+0.03$ & $+0.05$ & $+0.06$ & $+0.10$ \\
    \midrule
    Mean Len (tok) & 1,437 (deg.) & 600 & 613 & 686 & 820 & 990 & 1,093 & 1,211 & 1,436 \\
    Acc (Trunc) & $57.3\%$ ($12.3\%$) & $70.5\%$ & $74.1\%$ & $77.1\%$ & $81.7\%$ & $85.9\%$ & $86.0\%$ & $86.1\%$ & $\mathbf{87.4\%}$ ($1.6\%$) \\
    \bottomrule
  \end{tabular}
\end{table}

\begin{figure}[htbp]
  \centering
  \includegraphics[width=\textwidth]{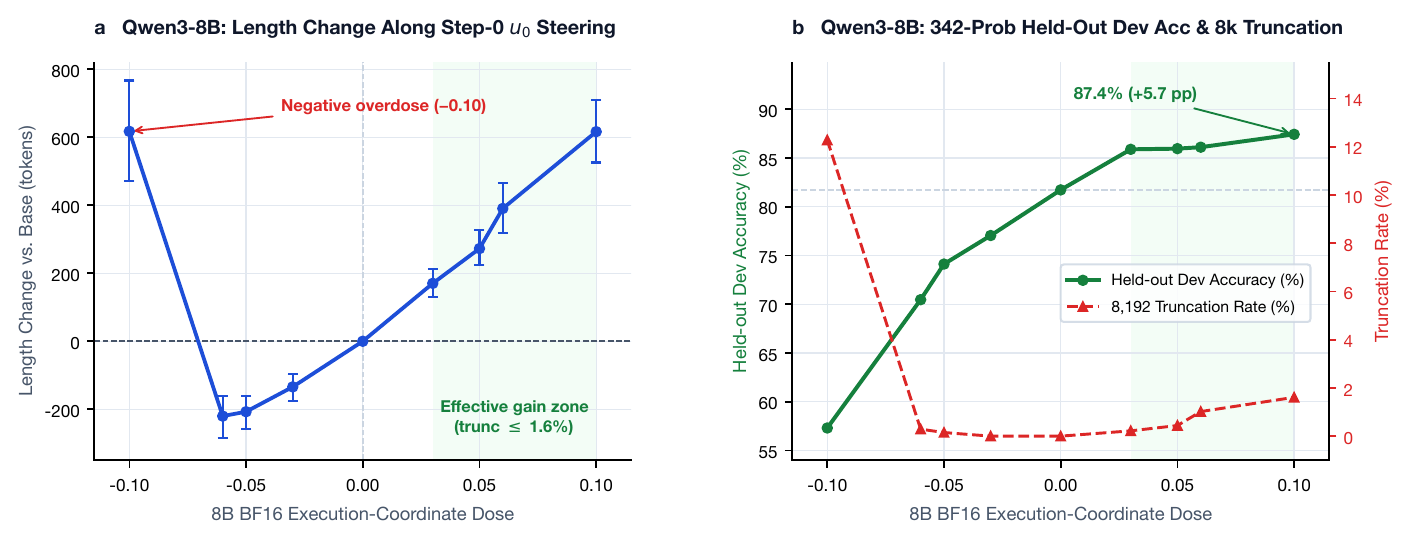}
  \caption{\textbf{Unsupervised $u_0$ steering scales to \texttt{Qwen3-8B} within a well-behaved positive band before breaking down at extreme negative doses (342 held-out dev problems, Table~\ref{tab:b7}).} Across the shaded positive band, shifting along step-0 $u_0$ smoothly lengthens reasoning and adds up to $5.7$ pp with minimal truncation ($\le 1.6\%$), confirming that pre-RL mode release is not confined to small models. Outside the useful band at $-0.10$, over-steering disrupts generation and stopping behavior, causing degenerate lengthening, $12.3\%$ truncation, and a sharp drop in accuracy.}
  \label{fig:b2}
\end{figure}

\clearpage
\subsection{Supporting Data for \S\ref{sec:net-gain} and \S\ref{sec:recovery}: Transfer, Corruption Probe, Frontier Shift, and Controls}
\label{app:b3}

\begin{table}[htbp]
  \centering
  \caption{Out-of-domain transfer of MATH step-0 $u_0$ steering and Seed-A BF16 $2 \times 2$ splits on \texttt{Qwen3-0.6B} (1 seed; \S\ref{sec:net-gain}), showing that $\CD$ breaks strict GSM8K formatting by lengthening responses whereas $\Eperp$ lifts MMLU and strict GSM8K at matched length and stable \texttt{wikitext-2} NLL.}
  \label{tab:b8}
  \footnotesize
  \renewcommand{\arraystretch}{1.18}
  \begin{tabular}{@{}lccc@{}}
    \toprule
    Configuration & MMLU Gen.\ Acc (Len) & GSM8K Strict Acc (\texttt{\#\#\#\#} / Len) & \texttt{wikitext-2} NLL \\
    \midrule
    Pre-RL base & $43.8\%$ (265 tok) & $13.9\%$ ($26.3\%$ / 179; lenient $50.5\%$) & $3.152$ \\
    Base $+1.0\,u_0$ (training-free) & $47.6\%$ (842, $+3.8$) & $7.0\%$ ($11.6\%$ / 557; lenient $63.3\%$) & $3.192$ \\
    $\CD^{0\to 40}$ only & $48.8\%$ (803, $+5.0$) & $8.2\%$ ($14.2\%$ / 524) & $3.186$ \\
    \textbf{$\bm{\Eperp^{0\to 40}}$ only} & $\mathbf{47.5\%}$ (292, $+3.7$) & $\mathbf{20.8\%}$ ($38.4\%$ / 178) & $\mathbf{3.164}$ \\
    \textbf{$\bm{\CD^{0\to 40} + \Eperp^{0\to 40}}$} & $\mathbf{50.9\%}$ (812, $+7.1$) & $\mathbf{27.1\%}$ ($37.1\%$ / 612) & $\mathbf{3.162}$ \\
    Full MATH GRPO (step 40) & $51.7\%$ (796, $+7.9$) & $26.4\%$ ($36.8\%$ / 611) & $3.162$ \\
    \bottomrule
  \end{tabular}
\end{table}

\begin{table}[htbp]
  \centering
  \caption{Clean accuracy ($h^+$) and corrupted-prefix recovery ($h^-$) after changing one digit in 193 prefix pairs across 156 problems (2,048-token continuation cap; \S\ref{sec:recovery}, Figure~\ref{fig:recovery}a--b). Whereas $h^-$ measures absolute recovery on corrupted prefixes, $h^+ - h^-$ measures the accuracy drop relative to clean prefixes. Paired bootstrap interaction CIs appear in Table~\ref{tab:b10}. Step-100 entries for Seeds~A and C use the paired main evaluation batch ($48.7\% / 53.8\%$); the unified 3-seed trajectory audit gives $46.6\% / 42.9\% / 52.6\%$ for Seeds~A/B/C (Tables~\ref{tab:b11b}, \ref{tab:c3}).}
  \label{tab:b9}
  \footnotesize
  \resizebox{\textwidth}{!}{%
  \begin{tabular}{@{}lccc@{}}
    \toprule
    Model / Configuration & Clean $h^+$ & Corrupted $h^-$ & Corrupted Cont.\ Len (tok) \\
    \midrule
    Pre-RL base & $39.4\%$ & $8.8\%$ & 294 \\
    Steer along $u_0$: $+1$ / $+2$ dose & $44.8\% \;/\; 21.4\%$ & $17.9\% \;/\; 7.3\%$ & $891 \;/\; 1{,}703$ \\
    Online $\Dsub$-locked GRPO step 40 (3 seeds) & $46.2\%$--$49.6\%$ & $16.5\%$--$30.4\%$ & 576--1,098 \\
    Seed A, step 40: $\CD$ / $\Eperp$ / $\CD+\Eperp$ & $47.0\% / 43.9\% / 51.3\%$ & $17.7\% / 14.9\% / \mathbf{29.3\%}$ & $885 / 531 / 1{,}049$ \\
    Seed B, step 40: $\CD$ / $\Eperp$ / $\CD+\Eperp$ & $42.6\% / 45.6\% / 51.3\%$ & $20.5\% / 15.5\% / \mathbf{33.8\%}$ & $1{,}158 / 508 / 1{,}223$ \\
    Full GRPO step 100 (Seed A / Seed C) & $57.5\% \;/\; 58.3\%$ & $\mathbf{48.7\% \;/\; 53.8\%}$ & $1{,}290 \;/\; 958$ \\
    \bottomrule
  \end{tabular}}
\end{table}

\begin{table}[htbp]
  \centering
  \caption{Effects and 95\% confidence intervals on the 193-pair corruption probe from a paired bootstrap clustered by problem (20,000 resamples across 156 problems, in pp; \S\ref{sec:step20}, \S\ref{sec:recovery}). A positive interaction on recovery $\Delta h^-$ does not by itself imply a positive interaction on the clean-to-corrupt drop reduction $\Delta h^- - \Delta h^+$, whose CIs still include zero in several interaction rows.}
  \label{tab:b10}
  \scriptsize
  \setlength{\tabcolsep}{4pt}
  \resizebox{\textwidth}{!}{%
  \begin{tabular}{@{}lcccc@{}}
    \toprule
    Contrast & Seed & $\Delta h^+$ & $\Delta h^-$ & Drop Reduction $\Delta h^- - \Delta h^+$ \\
    \midrule
    $0\to 40$ $\Eperp$ only & A & $+4.53\,[+1.62, +7.56]$ & $+6.09\,[+3.11, +9.06]$ & $+1.55\,[-2.16, +5.29]$ \\
    $0\to 40$ $\Eperp$ only & B & $+6.22\,[+3.01, +9.67]$ & $+6.74\,[+3.50, +10.13]$ & $+0.52\,[-3.43, +4.28]$ \\
    $0\to 40$ $\CD \times \Eperp$ interaction & A & $-0.26\,[-4.40, +3.82]$ & $\mathbf{+5.44\,[+0.91, +10.05]}$ & $\mathbf{+5.70\,[+0.37, +11.22]}$ \\
    $0\to 40$ $\CD \times \Eperp$ interaction & B & $+2.46\,[-1.70, +6.76]$ & $\mathbf{+6.61\,[+1.77, +11.55]}$ & $+4.15\,[-1.64, +10.06]$ \\
    Best steering ($+1\,u_0$) $\to$ RL step 100 & A & $+12.69\,[+8.81, +16.80]$ & $\mathbf{+30.83\,[+25.00, +36.70]}$ & $\mathbf{+18.13\,[+11.76, +24.62]}$ \\
    Best steering ($+1\,u_0$) $\to$ RL step 100 & C & $+13.47\,[+9.02, +18.19]$ & $\mathbf{+35.88\,[+30.08, +41.71]}$ & $\mathbf{+22.41\,[+15.76, +29.10]}$ \\
    $20\to 100$ add $\Eperp$ only & A & $+6.61\,[+3.12, +10.16]$ & $+13.47\,[+8.89, +18.16]$ & $\mathbf{+6.87\,[+1.55, +12.18]}$ \\
    $20\to 100$ add $\Eperp$ only & B & $+0.78\,[-2.42, +3.93]$ & $+8.29\,[+3.53, +13.23]$ & $\mathbf{+7.51\,[+1.97, +13.01]}$ \\
    $20\to 100$ add $\Eperp$ only & C & $-0.13\,[-3.01, +3.01]$ & $+4.02\,[+0.13, +7.96]$ & $+4.15\,[-0.50, +8.59]$ \\
    $20\to 100$ $\CD \times \Eperp$ interaction & A & $+7.25\,[+2.96, +11.78]$ & $\mathbf{+7.64\,[+2.14, +13.20]}$ & $+0.39\,[-5.88, +6.58]$ \\
    $20\to 100$ $\CD \times \Eperp$ interaction & C & $+13.99\,[+8.95, +19.12]$ & $\mathbf{+17.75\,[+11.98, +23.53]}$ & $+3.76\,[-2.49, +10.08]$ \\
    \bottomrule
  \end{tabular}}
\end{table}

\begin{table}[htbp]
  \centering
  \caption{Continuation trajectory diagnostics on the 193-pair corruption probe (\S\ref{sec:recovery}, Figure~\ref{fig:recovery}d). ``Implicit'' recovery denotes textually correct continuations with no explicit rewrite cue. A pair belongs to the ``early-corruption subset'' (54 pairs) if its original text does not restate the true value within 172 characters; on these 54 pairs, the standard last-step corruption drop is $+14.8$ pp for the pre-RL base versus $+1.9$ pp for RL100, while the early-corruption drop (row~3 below) falls from $+11.1$ pp to $0.0$ pp. For mid-solution handoffs ($t=16\to 256$ base tokens), isolated $\Eperp$ holds accuracy ($0.0$ pp vs.\ $-5.6$ pp for $\CD$), showing that $\CD$ depends on generating its own opening whereas $\Eperp$ continues reliably from an external intermediate state; suppressing \texttt{EOS} logits ($-12$) after natural stopping degrades RL100 accuracy by $13.5$ pp while leaving the pre-RL base flat ($-0.2$ pp), indicating calibrated stopping. In the last column, entries give Seed~C / Seed~A.}
  \label{tab:b11}
  \scriptsize
  \setlength{\tabcolsep}{2.8pt}
  \resizebox{\textwidth}{!}{%
  \begin{tabular}{@{}lccccc@{}}
    \toprule
    Diagnostic & Base & $\CD$ / $+u_0$ & $\Eperp$ only & RL 40 (A) & RL 100 (C / A) \\
    \midrule
    Explicit rewrite (clean$\to$corr.) & $4.1\!\to\!6.5\%$ & $7.0\!\to\!12.7\%$ & $6.5\!\to\!11.5\%$ & $16.3\!\to\!26.9\%$ & $20.5\!\to\!34.6\%$ / $22.4\!\to\!36.7\%$ \\
    Recovery $h^-$ (expl.\ / impl.) & $24.0 / 7.8\%$ & $35.7 / 15.3\%$ & $34.8 / 12.3\%$ & $58.7 / 32.6\%$ & $\mathbf{68.2 / 46.1\%}$ / $63.6 / 40.1\%$ \\
    Early-corr.\ drop $h^+\!-\!h^-$ (54) & $+11.1$ pp & $+7.4$ pp & $+6.0$ pp & $+1.9$ pp & $\mathbf{0.0}$ / $0.0$ pp \\
    Mid-handoff $t{=}16\!\to\!256$ ($\Delta$) & $35.6\%$ (self) & $55.9\!\to\!50.3$ ($-5.6$) & $43.9\!\to\!43.9$ ($\mathbf{0.0}$) & $64.9\!\to\!58.8$ ($-6.1$) & --- \\
    Post-\texttt{EOS} (\texttt{-12}) $\Delta$ acc (harm) & $-0.2$ ($1.59\%$) & --- & --- & --- & --- / $\mathbf{-13.5}$ ($18.72\%$) \\
    \bottomrule
  \end{tabular}}
\end{table}

\begin{table}[htbp]
  \centering
  \caption{Prefix continuation versus solving from scratch on the 193-pair (156-problem) corruption probe under a matched 2,048-token continuation cap (\S\ref{sec:recovery}). Lengths report newly generated continuation tokens only (excluding the $403.9$-token mean supplied prefix). Top: audit of $n=772$ corrupted continuations per seed across all three RL100 seeds, where \texttt{Err-Catch} / \texttt{Wrong} denote error-detection cues and citing the altered digit within the first 500 characters of recovered continuations. Bottom: solving from scratch (\texttt{no\_prefix}, 193-pair weighted) compared with clean ($h^+$) and corrupted ($h^-$) prefix continuations on Seed~A.}
  \label{tab:b11b}
  \scriptsize
  \setlength{\tabcolsep}{2.6pt}
  \begin{tabular}{@{}lcccc@{}}
    \toprule
    Model / Condition & Acc (193-wt.) & Len (All / Ok) & Restart (All / Rec.) & Err-Catch / Wrong \\
    \midrule
    RL 100 (Seed A, $h^-$) & $46.6\%$ & 1,265 / 1,028 & $2.5\%$ / $\mathbf{3.3\%}$ & $\mathbf{76.7\%}$ / $\mathbf{55.0\%}$ \\
    RL 100 (Seed B, $h^-$) & $42.9\%$ & 1,039 / 842 & $0.8\%$ / $\mathbf{0.9\%}$ & $\mathbf{61.0\%}$ / $\mathbf{53.2\%}$ \\
    RL 100 (Seed C, $h^-$) & $52.6\%$ & 967 / 808 & $1.0\%$ / $\mathbf{1.2\%}$ & $\mathbf{73.9\%}$ / $\mathbf{44.3\%}$ \\
    \midrule
    Base (\texttt{no\_prefix}) & $33.2\%$ ($1.9\%$ tr.) & 561 / 381 & --- & --- \\
    RL 100 A (\texttt{no\_prefix}) & $57.6\%$ ($\mathbf{43.1\%}$ tr.) & 1,502 / $\mathbf{1{,}225}$ & --- & --- \\
    RL 100 A ($h^+$, 404 pref.) & $57.5\%$ & 1,117 / $\mathbf{768}$ & --- & $404{+}768{=}1{,}172{\approx}1{,}225$ \\
    \bottomrule
  \end{tabular}
\end{table}

\begin{figure}[htbp]
  \centering
  \includegraphics[width=\textwidth]{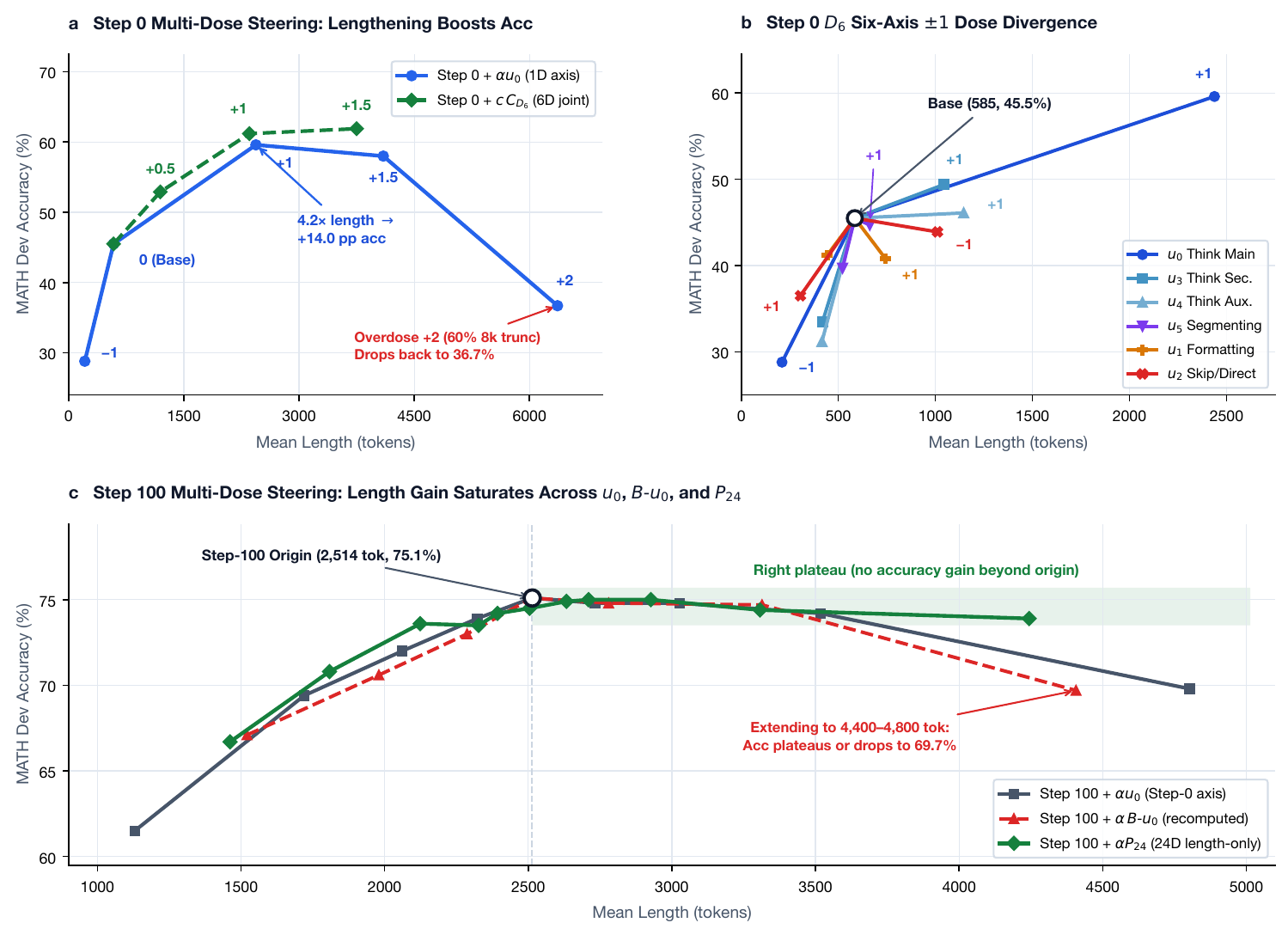}
  \caption{\textbf{Length steering loses its accuracy leverage by step~100 because RL has already saturated the useful length regime (\texttt{Qwen3-0.6B}, MATH dev, 8k cap).} \textbf{(a)} Before training, \nothink{} holds the base model in an under-thinking regime where shifting along $u_0$ ($59.6\%$ at $+1$) or joint $\CD$ ($61.2\%$ at $+1$, $61.9\%$ at $+1.5$) unlocks large accuracy gains until $+2\,u_0$ runs into 8k truncation (Table~\ref{tab:b12}). \textbf{(b)} Individual $\pm 1$ axis sweeps confirm that these step-0 gains come selectively from think-more axes rather than arbitrary $\Dsub$ perturbations (Table~\ref{tab:b5}). \textbf{(c)} By step~100, further lengthening along $u_0$, the recomputed top axis $B\text{-}u_0$, or the 24D length-subspace direction $P_{24}$ yields little or no gain and hurts accuracy at large doses, directly explaining why later RL gains cannot come from further lengthening.}
  \label{fig:b1}
\end{figure}

\clearpage
\paragraph{Length-matched convex hull and multi-stage steering.}
To rule out pure-length gains, we compute the upper convex hull of all tested reward-free length controls (template switching, $u_0$ steering, and truncation). Step-100 RL beats this frontier at every tested length point, by $\mathbf{7.3\text{--}13.9}$ pp at 1,000 tokens, $\mathbf{10.5\text{--}13.9}$ pp at 1,500 tokens, and $\mathbf{8.4\text{--}10.0}$ pp at 2,500 tokens, demonstrating an upward shift of the full accuracy--length curve (sweeps in Table~\ref{tab:b12}, Figure~\ref{fig:b1}).

\begin{table}[htbp]
  \centering
  \caption{Cross-stage training-free steering sweeps on \texttt{Qwen3-0.6B} (MATH dev, 8k cap; Figures~\ref{fig:recovery}c, \ref{fig:b1}). Reported peaks depend on the swept dose grid. $B\text{-}u_0$ is the recomputed step-100 top axis, and $P_{24}$ denotes a 1D reward-free length direction from a 24D cross-checkpoint length subspace (with negative-reward components removed); in an independent evaluation batch for $P_{24}$, small fluctuations around the origin ($2{,}505$ tokens / $74.46\%$; $+1$: $2{,}927$ / $74.98\%$; $+4$: $4{,}244$ / $73.85\%$) are consistent with little or no additional accuracy. $\bar{E}_{\mathrm{lock}}$ is the mean 40-step locked update across seeds, evaluated on dev.}
  \label{tab:b12}
  \scriptsize
  \setlength{\tabcolsep}{2.8pt}
  \renewcommand{\arraystretch}{1.15}
  \begin{tabularx}{\textwidth}{@{}>{\raggedright\arraybackslash}p{2.25cm}>{\raggedright\arraybackslash}p{1.65cm}>{\raggedright\arraybackslash}X>{\raggedright\arraybackslash}X>{\raggedright\arraybackslash}X>{\raggedright\arraybackslash}X@{}}
    \toprule
    Checkpoint & Axis \& Range & Shortest (Len / Acc) & Origin ($\alpha=0$) & Peak (Len / Acc) & Longest (Len / Acc) \\
    \midrule
    Step~0 (Base) & $u_0$, $[-1, +2]$ & $210$ / $28.8\%$ & $585$ / $45.5\%$ & $2{,}438$ ($4.17\times$) / $59.6\%$ & $6{,}365$ / $36.7\%$ ($60\%$ tr.) \\
    \addlinespace[2pt]
    Step~20 (A; 3-seed) & $u_0$, $[-1, +1]$ & $698/46.6\%$ ($582/45.0\%$) & $2{,}449/66.6\%$ ($2{,}110/65.3\%$) & $2{,}449/66.6\%$ ($3{,}257/66.4\%$) & $5{,}579/52.0\%$ ($4{,}921/57.3\%$) \\
    \addlinespace[2pt]
    Step~40 locked ($\bar{E}_{\mathrm{lock}}$) & $\bar{C}_{D_6}$, $[0, 1.25]$ & --- (origin) & $870$ / $56.5\%$ & $2{,}695$ ($3.10\times$) / $\mathbf{70.7\%}$ & $4{,}229$ / $66.4\%$ ($29\%$ tr.) \\
    \addlinespace[2pt]
    Step~100 (Seed~A) & $u_0$ / $B\text{-}u_0$ / $P_{24}$ & $1{,}130$ / $61.5\%$ & $2{,}514$ / $\mathbf{75.1\%}$ & Origin ($P_{24}$ flat $\pm 0.6$) & $4{,}802$ / $69.8\%$ ($B\text{-}u_0$: $69.7\%$) \\
    \bottomrule
  \end{tabularx}
\end{table}

\begin{table}[htbp]
  \centering
  \caption{MATH 32-token direct-answer format control (\texttt{0.6B}, dev $1{,}024 \times 4$; \S\ref{sec:net-gain}, Table~\ref{tab:controls}), showing that RL adds $5.5\text{--}5.8$ pp even after prefilling \boxedtok{} or format-axis steering closes the format gap.}
  \label{tab:b13}
  \footnotesize
  \resizebox{\textwidth}{!}{%
  \begin{tabular}{@{}lccccc@{}}
    \toprule
    Intervention & Accuracy & Box Rate & Acc $\mid$ Box & Mean Len (tok) & Truncation \\
    \midrule
    Pre-RL base & $3.64\%$ & $23.68\%$ & $15.36\%$ & 27.68 & $76.95\%$ \\
    Prefill \boxedtok{} (format baseline) & $8.11\%$ & $99.41\%$ & $8.15\%$ & 5.26 & $0.90\%$ \\
    Format-axis steering ($-0.005$) & $8.25\%$ & $97.36\%$ & $8.48\%$ & 9.40 & $2.78\%$ \\
    Format-aligned + locked $D_6^{\mathrm{dir}}$ RL (step 45) & $13.55\%$ & $99.68\%$ & $13.59\%$ & ${\sim}8.3$ & $<1\%$ \\
    Unconstrained GRPO (step 60 / 120) & $13.77\% / 13.87\%$ & $99.88\% / 99.71\%$ & $13.79\% / 13.91\%$ & $8.24 / 8.39$ & $0.20\% / 0.90\%$ \\
    \bottomrule
  \end{tabular}}
\end{table}

\begin{table}[htbp]
  \centering
  \caption{\texttt{Qwen3-8B} GSM8K scratchpad-free mental arithmetic (blind test $1{,}319 \times 4$, 64-tok cap, 3 seeds; \S\ref{sec:net-gain}, Table~\ref{tab:controls}), showing a $9.71$ pp mean RL gain even in $3.5$-token numeric-only answers.}
  \label{tab:b14}
  \scriptsize
  \setlength{\tabcolsep}{3.5pt}
  \resizebox{\textwidth}{!}{%
  \begin{tabular}{@{}lcccl@{}}
    \toprule
    Model & Default (Acc / Len / Trunc) & Prefill \texttt{"\#\#\#\# "} (Acc / Len) & Pure Digits / Parse & Notes \\
    \midrule
    \texttt{Qwen3-8B} base & $17.63\% / 20.1 / 21.8\%$ & $17.68\% / 4.89$ & $93.58\% / 100\%$ & Prefill adds $0.05$ pp \\
    \texttt{8B} GRPO S1/S2/S3 (step 60) & $27.54 / 27.62 / 26.76\%$ (5.4, $0\%$) & $27.31 / 27.67 / 27.20\%$ (3.45--3.50) & $99.17\text{--}99.43\% / 100\%$ & Mean $\mathbf{27.39\%}$ ($\mathbf{+9.71}$ pp) \\
    \bottomrule
  \end{tabular}}
\end{table}

\clearpage
\subsection{Supporting Data for \S\ref{sec:coverage}--\S\ref{sec:sharpening}: Large-\texorpdfstring{$k$}{k} Trajectories, Judge Calibration, and Sharpening Tables}
\label{app:b4}

\begin{table}[htbp]
  \centering
  \caption{Empirical evaluation trajectories on 512 held-out problems with $n=64$ samples per problem (\nothink{}, 8k cap, Seed~A; \S\ref{sec:coverage}, Table~\ref{tab:passk}). Most of full RL's \passk{64} gain already appears by step~20 ($+4.30$ of $+5.08$ pp), whereas single-sample accuracy (\texttt{avg@64}) continues to rise through step~100.}
  \label{tab:b15a}
  \scriptsize
  \setlength{\tabcolsep}{4pt}
  \renewcommand{\arraystretch}{1.32}
  \begin{tabular}{@{}lcccccc@{}}
    \toprule
    Model / Step & \texttt{avg@64} & \passk{16} & \passk{64} & L4--5 @64 & $\Delta$\passk{64} [95\% CI] & Len (Trunc) \\
    \midrule
    Pre-RL base & $42.78$ & $79.87$ & $88.09$ & $80.0$ & --- & 577 ($0.0\%$) \\
    Full RL step 10 & $50.94$ & $85.66$ & $91.80$ & $86.0$ & $+3.71$ & 1,013 \\
    Full RL step 20 & $65.73$ & $89.02$ & $92.38$ & $87.2$ & $+4.30$ & 2,508 \\
    Full RL step 30 & $69.34$ & $88.58$ & $92.58$ & $87.6$ & $+4.49$ & 2,978 \\
    Full RL step 40 & $70.08$ & $90.15$ & $93.36$ & $89.2$ & $+5.27$ & 2,494 \\
    Full RL step 60 & $71.86$ & $90.55$ & $\mathbf{93.75}$ & $90.4$ & $\mathbf{+5.66}$ & 2,564 \\
    Full RL step 80 & $74.07$ & $90.01$ & $92.77$ & $89.6$ & $+4.69$ & 2,821 \\
    Full RL step 100 & $\mathbf{75.30}$ & $90.29$ & $93.16$ & $90.4$ & $+5.08\,[2.54, 7.62]$ & 2,552 ($7.2\%$) \\
    Full RL step 100 (indep.) & $74.52$ & $89.53$ & $91.99$ & $88.0$ & $+3.91\,[1.56, 6.25]$ & 2,734 ($6.2\%$) \\
    $\Eperp^{0\to 10}$ & $46.00$ & $81.34$ & $87.70$ & $78.4$ & $-0.39$ & 702 ($0.0\%$) \\
    $\Eperp^{0\to 20}$ & $50.31$ & $84.16$ & $90.62$ & $84.4$ & $+2.54\,[0.39, 4.69]$ & 1,022 ($0.8\%$) \\
    $\Eperp^{0\to 40}$ (same) & $52.13$ & $84.61$ & $90.23$ & $83.2$ & $+2.15\,[0.00, 4.49]$ & 969 ($0.5\%$) \\
    $\Eperp^{0\to 40}$ (indep.) & $52.45$ & $85.09$ & $91.21$ & $84.8$ & $+3.12\,[0.98, 5.27]$ & 982 ($0.5\%$) \\
    $\Eperp^{0\to 60}$ & $52.97$ & $84.53$ & $89.84$ & $82.8$ & $+1.76\,[-0.20, 3.91]$ & 942 ($0.2\%$) \\
    $\Eperp^{0\to 100}$ & $56.28$ & $84.77$ & $89.45$ & $82.4$ & $+1.37\,[-0.98, 3.71]$ & 951 ($0.1\%$) \\
    $\CD^{0\to 20}$ & $57.04$ & $88.05$ & $\mathbf{93.55}$ & $88.8$ & $\mathbf{+5.47}\,[3.12, 8.01]$ & 1,676 ($4.3\%$) \\
    $\CD^{0\to 40}$ & $60.81$ & $88.17$ & $92.77$ & $87.2$ & $+4.69$ & 2,365 ($10.9\%$) \\
    $\CD^{0\to 100}$ & $62.95$ & $87.79$ & $92.38$ & $87.2$ & $+4.30$ & 2,781 ($15.1\%$) \\
    Locked $\Dsub$ step 40 (A) & $56.81$ & $86.97$ & $90.82$ & $85.6$ & $+2.73\,[0.78, 4.88]$ & 1,069 ($0.4\%$) \\
    \bottomrule
  \end{tabular}
\end{table}

\begin{table}[htbp]
  \centering
  \caption{Pairwise bootstrap 95\% CIs for large-$k$ \texttt{Level 4--5} \passk{k} ($n=256$) and hard-problem \cotpassk{192} (disjoint samples \texttt{64--255}, Seed~A; Table~\ref{tab:passk}). Rows labeled ``(8k)'' post-hoc rescore the 16k samples by counting only outputs of at most $8{,}192$ tokens as hits (rather than regenerating under an 8k cap): Base \think{} 8k gives @1 $53.54$, @256 $90.4$, CoT $46.0\%$; RL \think{} 8k gives @1 $64.56$, @256 $90.8$, CoT $56.0\%$; Native \think{} RL 8k gives @1 $58.91$, @256 $90.4$, CoT $46.0\%$. RL 40 \nothink{} achieves $54.0\%$ Hard-50 \cotpassk{192} ($79.3\%$ valid rate). On 512 held-out problems (\think{}/16k, $n=4$), \textbf{Native \think{} RL step~100} achieves $75.49\%$ accuracy (mean length $4{,}302$, $0.0\%$ truncation). The Overlap column reports Hard-50 \cotpassk{192} counts solved by both models, only the first, and only the second.}
  \label{tab:b15b}
  \scriptsize
  \setlength{\tabcolsep}{2.6pt}
  \begin{tabular}{@{}lcccc@{}}
    \toprule
    Pairwise Contrast & L4--5 @1 & L4--5 @256 & Hard CoT@192 & Overlap \\
    \midrule
    Base \think{} $-$ Base \nothink{} & $+35.5\,[32.3, 38.8]$ & $+3.2\,[0.4, 6.4]$ & $\mathbf{+34\,[18, 50]}$ & 5 / 20 / 3 \\
    RL 40 \nothink{} $-$ Base \nothink{} & $+30.8\,[27.4, 34.0]$ & $+2.4\,[-0.8, 5.6]$ & $\mathbf{+38\,[24, 52]}$ & --- \\
    RL 100 \nothink{} $-$ Base \nothink{} & $+37.0\,[33.4, 40.6]$ & $+3.2\,[0.0, 6.4]$ & $\mathbf{+36\,[20, 52]}$ & --- \\
    RL 100 \nothink{} $-$ Base \think{} & $+1.5\,[-0.5, 3.4]$ & $\mathbf{0.0\,[-1.6, 1.6]}$ & $\mathbf{+2\,[-6, 10]}$ & 23 / 3 / 2 \\
    RL 100 \nothink{} $-$ Base \think{} (8k) & $+9.3\,[7.3, 11.5]$ & $+2.4\,[0.4, 4.8]$ & $+6\,[-2, 14]$ & --- \\
    $\CD^{0\to 100}$ $-$ Base \nothink{} & $+21.6\,[19.0, 24.5]$ & $+1.6\,[-1.6, 4.8]$ & $\mathbf{+40\,[26, 54]}$ & --- \\
    RL 100 \nothink{} $-$ $\CD^{0\to 100}$ & $\mathbf{+15.4\,[13.3, 17.5]}$ & $+1.6\,[-0.8, 4.0]$ & $-4\,[-14, 6]$ & 24 / 2 / 4 \\
    RL 100 \think{} $-$ Base \think{} (Ctrl~1) & $\mathbf{+7.2\,[5.5, 9.0]}$ & $\mathbf{+0.4\,[-1.2, 2.0]}$ & $+10\,[0, 20]$ ($p{=}.125$) & 24 / 6 / 1 \\
    RL 100 \think{} $-$ Base \think{} (8k) & $\mathbf{+11.0\,[9.2, 12.9]}$ & $+0.4\,[-1.6, 2.4]$ & $+10\,[0, 20]$ & --- \\
    RL 100 \think{} $-$ RL 100 \nothink{} & $+5.7\,[4.5, 7.0]$ & $+0.4\,[0.0, 1.2]$ & $+8\,[2, 16]$ & 26 / 4 / 0 \\
    Native \think{} RL 100 $-$ Base \think{} & $\mathbf{+2.6\,[0.8, 4.2]}$ & $\mathbf{-0.4\,[-2.4, 1.6]}$ & $+2\,[-12, 14]$ ($p{=}1.0$) & 20 / 6 / 5 \\
    Native \think{} RL 100 $-$ Base \think{} (8k) & $\mathbf{+5.4\,[3.6, 7.2]}$ & $\mathbf{0.0\,[-2.4, 2.4]}$ & $0\,[-12, 12]$ ($p{=}1.0$) & 18 / 5 / 5 \\
    Native \think{} RL 100 $-$ RL 100 \think{} & $-4.6\,[-6.6, -2.8]$ & $-0.8\,[-2.8, 1.2]$ & $-8\,[-20, 4]$ ($p{=}.34$) & 23 / 3 / 7 \\
    \bottomrule
  \end{tabular}
\end{table}

\paragraph{CoT judge calibration.}
To verify that the low \cotpassk{192} of base \nothink{} reflects genuinely missing reasoning chains rather than judge bias against shorter outputs, we calibrate our CoT judge (\texttt{Qwen3.8-27B} in thinking mode, 3-vote majority, 12,288-token cap) on 150 answer-correct Level 1--3 responses per evaluation condition and cross-check it against a second judge. It rates $81.3\%$ valid for base \nothink{}, $91.3\%$ for RL 100 \nothink{}, and $96.0\%$ for base \think{}; against \texttt{Qwen3.5-35B-A3B} on 16 trajectories (48 votes), the two judges agree by majority vote on 13/16 trajectories, and manual inspection confirms all 7 invalid rulings, confirming reliable validity scoring.

\paragraph{Hard-50 sensitivity to voting rules and step-40 vs.\ step-100 coverage.}
To test whether our Hard-50 coverage conclusions depend on permissive majority voting, we re-evaluate under unanimous 3/3 validity agreement. The main ordering stays intact: base \nothink{} \cotpassk{192} falls to $10.0\%$ (5 problems), whereas base \think{} remains at $50.0\%$ (25); Native \think{} RL 100 reaches $46.0\%$ (23), RL 40 and RL 100 \nothink{} remain tied at $52.0\%$ (26 each), and RL 100 \think{} stays at $60.0\%$ (30). Comparing step~40 and step~100 under 3-vote majority likewise shows 27 solved problems at step~40 and 26 at step~100 (26 each under unanimous 3/3 agreement), while single-sample hit frequency rises over the same interval (Hard-50 answer-only \passk{1}: $11.36\% \to 13.79\%$; \cotpassk{1}: $9.01\% \to 9.57\%$). Thus, later training increases hit frequency on already-accessible problems rather than expanding the solved set, robustly across voting rules.

\paragraph{Greedy lag and mild late sharpening in isolated $\Eperp$.}
Positive token-level temperature scaling leaves the greedy ($\arg\max$) trajectory strictly invariant ($0$ pp greedy change). Empirically, isolated $\Eperp$ improves $T=1$ sampling well before greedy decoding: $\Eperp^{0\to 20}$ changes accuracy by $+7.5$ pp at $T=1$ versus $-1.4$ pp greedy, $\Eperp^{0\to 40}$ yields $+9.4$ versus $+2.5$ pp, and by step~100 ($\Eperp^{0\to 100}$) the greedy gain reaches $+4.7$ pp (CI $[0.6, 8.8]$ pp), while $T=1$ accuracy ($56.28\%$) exceeds the pre-RL greedy baseline ($51.17\%$) and retains $+11.9$ of $+13.5$ pp at calibrated temperature $T^*=1.20$. Together, these observations indicate that early $\Eperp$ elevates non-dominant correct paths before shifting the greedy mode, while steps 20--100 introduce mild late sharpening: logit dispersion rises ($1.033 \to 1.110$), the within-prompt base log-probability slope flips ($-0.111 \to +0.105$), and wrong-mode hits on 0/32-base-correct prompts tick up ($0.19 \to 0.23$ over steps 60--100).

\paragraph{Counterfactual sharpening bounds on $\Eperp^{0\to 100}$.}
Because different sharpening definitions attribute different shares of the $13.5$ pp total sampling gain on $\Eperp^{0\to 100}$, we bracket sharpening with three counterfactual estimators (definition-dependent bounds rather than a confidence interval for one estimator). Temperature calibration to the pre-RL entropy threshold ($T^* = 1.20$) explains only $1.6$ pp ($12\%$, lower bound, leaving $11.9$ pp retained) because it tests uniform logit scaling alone; non-accuracy exponential tilting on the base pool gives $6.9$ pp (a coarse midpoint across pools); and an answer-level power law $P_0(a)^\alpha$ matched to the observed answer entropy of $1.19$ predicts $10.2$ pp ($75\%$). The wide span reflects their assumptions: the power-law upper bound overstates sharpening because it assumes all answer-entropy reduction is mechanical sharpening, forcing wrong-mode inflation not observed on \texttt{mode\_wrong\_with\_support}.

\paragraph{Majority-vote RL setup.}
For majority-vote RL, we sample 8 pre-RL rollouts per training prompt, yielding 5,734 prompts with mode count $\ge 2$ ($62.7\%$ ground-truth agreement) and 666 zero-advantage prompts. Training runs on 8-GPU DDP (versus 2-GPU DDP for the primary runs) with identical global batch size, hyperparameters, seed, and data order.

\begin{table}[htbp]
  \centering
  \caption{Teacher-forcing (TF) entropy, logit dispersion ratio, and calibrated temperature $T^*$ (restoring $\HTF$ to at least the pre-RL threshold on the 8,192-trajectory TF pool), together with $T=1 \to T^*$ sampling accuracy gains on the 512 held-out problems (\S\ref{sec:sharpening}, Table~\ref{tab:sharpening}).}
  \label{tab:b16a}
  \scriptsize
  \setlength{\tabcolsep}{2.6pt}
  \renewcommand{\arraystretch}{1.32}
  \begin{tabular}{@{}lcccccc@{}}
    \toprule
    Model / Step & $\HTF(T{=}1)$ & $\Delta\HTF$ & Logit Ratio & $T^*$ & $T{=}1\!\to\!T^*$ Gain (pp) & Temp.\ Sens. \\
    \midrule
    Pre-RL base & $0.3438$ & --- & $1.000$ & $1.00$ & $42.78\%$ (base) & --- \\
    RL 10 / 20 & $.336$ / $.328$ & $-.008$ / $-.016$ & $1.000$ / $1.014$ & $1.05$ / $1.05$ & $+8.2\!\to\!7.5$ / $+23.0\!\to\!23.1$ & $8\%$ / $0\%$ \\
    RL 40 / 60 & $.300$ / $.287$ & $-.044$ / $-.057$ & $1.041$ / $1.061$ & $1.15$ / $1.15$ & $+27.3\!\to\!26.5$ / $+29.1\!\to\!28.3$ & $3\%$ / $3\%$ \\
    RL 100 & $.278$ & $-.066$ & $1.071$ & $1.20$ & $+32.5\!\to\!32.2$ & $1\%$ \\
    $\Eperp^{0\to 10}$ / $\Eperp^{0\to 20}$ & $.339$ / $.325$ & $-.005$ / $-.019$ & $1.011$ / $1.033$ & $1.05$ / $1.05$ & $+3.2\!\to\!2.4$ / $+7.5\!\to\!6.9$ & $27\%$ / $9\%$ \\
    $\Eperp^{0\to 40}$ (same / indep.) & $.300$ / $.299$ & $-.044$ / $-.045$ & $1.067$ / $1.066$ & $1.15$ / $1.15$ & $+9.4\!\to\!7.9$ / $+9.7\!\to\!7.4$ & $16\%$ / $24\%$ \\
    $\Eperp^{0\to 60}$ / $\Eperp^{0\to 100}$ & $.288$ / $.273$ & $-.056$ / $-.071$ & $1.091$ / $\mathbf{1.110}$ & $1.15$ / $1.20$ & $+10.2\!\to\!9.4$ / $+13.5\!\to\!\mathbf{11.9}$ & $8\%$ / $12\%$ \\
    $\CD^{0\to 20}$ / $\CD^{0\to 40}$ & $.349$ / $.350$ & $+.005$ / $+.006$ & $.980$ / $.971$ & $1.00$ / $1.00$ & $+14.3$ / $+18.0$ & $0\%$ / $0\%$ \\
    $\CD^{0\to 100}$ & $.355$ & $+.011$ & $.965$ & $1.00$ & $+20.2$ & $0\%$ \\
    Maj.\ 20 / 40 & $.336$ / $.324$ & $-.008$ / $-.020$ & $1.010$ / $1.028$ & $1.05$ / $1.05$ & $+15.3\!\to\!14.6$ / $+20.0\!\to\!19.4$ & $5\%$ / $3\%$ \\
    Maj.\ 60 / 100 & $.337$ / $.329$ & $-.007$ / $-.015$ & $1.028$ / $1.037$ & $1.05$ / $1.05$ & $+18.7\!\to\!18.0$ / $+16.9\!\to\!16.3$ & $4\%$ / $3\%$ \\
    \bottomrule
  \end{tabular}
\end{table}

\begin{table}[htbp]
  \centering
  \caption{Within-prompt log-probability regressions, \texttt{Hard-Flip} (\texttt{HF}) win rates (95\% CI width ${\sim}\pm 0.02$), and base-pool exponential tilt counterfactuals (8,192-trajectory TF pool; \S\ref{sec:sharpening}). Here $\beta_{\mathrm{std}}(lp_0)$ and $\beta_{\mathrm{std}}(\mathrm{acc})$ are standardized within-prompt regression slopes of per-token mean log-probability shift $\Delta\bar{\ell}$ on base log-probability $lp_0$ and correctness $\mathrm{acc}$; \texttt{Pool HF} is the fraction of within-prompt pairs with $lp_0(\text{wrong}) > lp_0(\text{right})$ where the correct trajectory receives a larger log-probability increase ($>0.5$ opposes pure prior sharpening); \texttt{Base Tilt} is the accuracy shift from exponentially tilting the base trajectory pool by the fitted $lp_0$ slope alone; and \texttt{Indep.\ HF} evaluates \texttt{HF} on an independent raw base-sampling subpool.}
  \label{tab:b16b}
  \scriptsize
  \setlength{\tabcolsep}{4pt}
  \renewcommand{\arraystretch}{1.32}
  \begin{tabular}{@{}lccccc@{}}
    \toprule
    Model / Step & $\beta_{\mathrm{std}}(lp_0)$ & $\beta_{\mathrm{std}}(\mathrm{acc})$ & Pool \texttt{HF} & Base Tilt (pp) & Indep.\ \texttt{HF} \\
    \midrule
    RL 10 / 20 & $-.27$ / $-.23$ & $+.15$ / $+.15$ & $.83$ / $.84$ & --- / $+3.3$ & --- / $.72$ \\
    RL 40 / 60 & $-.18$ / $-.17$ & $+.14$ / $+.13$ & $.84$ / $.84$ & $+2.9$ / --- & $.73$ / --- \\
    RL 100 & $-.15$ & $+.15$ & $.84$ & $+3.8$ & $.65$ \\
    $\Eperp^{0\to 10}$ / $\Eperp^{0\to 20}$ & $-.151$ / $-.111$ & $+.113$ / $+.131$ & $.809$ / $.818$ & $+2.8$ / $+5.1$ & $.64$ / $.61$ \\
    $\Eperp^{0\to 40}$ (same / indep.) & $-.011$ / $+.012$ & $+.121$ / $+.122$ & $.801$ / $.797$ & $+4.9$ / --- & $.52$ / --- \\
    $\Eperp^{0\to 60}$ / $\Eperp^{0\to 100}$ & $-.004$ / $\mathbf{+.105}$ & $+.117$ / $+.141$ & $.802$ / $.778$ & $+4.5$ / $\mathbf{+6.9}$ & $.51$ / $\mathbf{.45}$ \\
    $\CD^{0\to 20}$ / $\CD^{0\to 40}$ & $-.25$ / $-.20$ & $+.15$ / $+.14$ & $.83$ / $.82$ & $+1.9$ / $+1.5$ & $.68$ / $.70$ \\
    $\CD^{0\to 100}$ & $-.17$ & $+.14$ & $.82$ & $+1.6$ & $.66$ \\
    Maj.\ 20 / 40 & $-.31$ / $-.27$ & $+.14$ / $+.14$ & $.83$ / $.84$ & $+2.2$ / $+2.6$ & $.71$ / $.68$ \\
    Maj.\ 60 / 100 & $-.26$ / $-.17$ & $+.15$ / $+.16$ & $.84$ / $.83$ & $+3.5$ / $+4.7$ & $.64$ / $.62$ \\
    \bottomrule
  \end{tabular}
\end{table}

\begin{table}[htbp]
  \centering
  \caption{Base-polarity stratification (grouping problems by whether the base model's most frequent answer is correct) for majority-vote RL, ground-truth RL, and $\Eperp$ (512 held-out, Seed~A; \S\ref{sec:sharpening}, Figure~\ref{fig:sharpening}). Parentheses show base-mode hit rates (base: $0.60/0.28/0.22/0.98$).}
  \label{tab:b17a}
  \scriptsize
  \setlength{\tabcolsep}{5pt}
  \renewcommand{\arraystretch}{1.32}
  \begin{tabular}{@{}lcccc@{}}
    \toprule
    Model / Step & \texttt{mode\_correct} (280) & \texttt{mode\_wrong} (121) & \texttt{0/32 base-corr.} (82) & \texttt{all\_correct} (29) \\
    \midrule
    Majority RL 20 & $+18.6$ ($\to .79$) & $+19.5$ ($\to .22$) & $+4.6$ ($\to .22$) & $+1.2$ \\
    Majority RL 40 & $+24.0$ ($\to .84$) & $+25.5$ ($\to .22$) & $+6.3$ ($\to .24$) & $+1.8$ \\
    Majority RL 60 & $+23.7$ ($\to .84$) & $+22.2$ ($\to .25$) & $+3.6$ ($\to .29$) & $+2.0$ \\
    \textbf{Majority RL 100} & $+22.3$ ($\to .83$) & $+18.8$ ($\mathbf{\to .30}$) & $+2.1$ ($\mathbf{\to .36}$) & $+2.2$ \\
    GT RL 20 & $+25.0$ ($\to .85$) & $+33.2$ ($\to .14$) & $+9.8$ ($\to .10$) & $+1.6$ \\
    GT RL 40 & $+28.2$ (---) & $+42.1$ ($\to .11$) & $+13.2$ ($\to .10$) & --- \\
    GT RL 60 & $+29.5$ ($\to .90$) & $+45.3$ ($\to .11$) & $+15.3$ ($\to .11$) & $+1.2$ \\
    GT RL 100 & $+32.7$ ($\to .93$) & $+49.9$ ($\to .11$) & $+18.4$ ($\to .13$) & $+2.3$ \\
    $\Eperp^{0\to 20}$ & --- & $+8.5$ ($\to .24$) & --- ($\to .19$) & --- \\
    $\Eperp^{0\to 40}$ & $+12.0$ ($\to .72$) & $+11.1$ ($\to .25$) & $+2.3$ ($\to .18$) & $+1.1$ \\
    $\Eperp^{0\to 60}$ & $+12.8$ ($\to .73$) & $+12.6$ ($\to .25$) & $+2.4$ ($\to .19$) & $+1.0$ \\
    $\Eperp^{0\to 100}$ & $+17.3$ ($\to .78$) & $\mathbf{+15.8}$ ($\to .27$) & $+3.1$ ($\to .23$) & $+1.2$ \\
    \bottomrule
  \end{tabular}
\end{table}

\begin{table}[htbp]
  \centering
  \caption{Paired \passk{32} McNemar tests, answer-level entropy, power-law sharpening upper bounds, and training-window lengths (4,096-token cap; \S\ref{sec:sharpening}). RL~40 and $\Eperp^{0\to 20}$ greedy gains are $+16.4$ and $-1.4$ pp.}
  \label{tab:b17b}
  \scriptsize
  \setlength{\tabcolsep}{5pt}
  \renewcommand{\arraystretch}{1.32}
  \begin{tabular}{@{}lcccc@{}}
    \toprule
    Model / Step & McNemar@32 (Model/Base, $p$) & Answer Entropy & Power-Law Upper Bound & Train Len (Trunc) \\
    \midrule
    Majority RL 20 & 34 / 9 ($2\times 10^{-4}$) & $1.26$ & $+9.3$ pp & 1,045 ($4.1\%$) \\
    Majority RL 40 & 31 / 9 ($7\times 10^{-4}$) & $1.01$ & $+12.4$ pp & ${\sim}1{,}419$ ($5.4\%$) \\
    Majority RL 60 & 19 / 15 ($0.61$) & $0.93$ & $+13.3$ pp & 1,419 ($5.4\%$) \\
    \textbf{Majority RL 100} & $\mathbf{9 / 34\;(2\times 10^{-4})}$ & $0.73$ & $+15.3$ pp & 1,397 ($0.8\%$) \\
    GT RL 20 & 38 / 8 ($9\times 10^{-6}$) & $0.75$ & $+15.2$ pp & 1,409 ($15.7\%$) \\
    GT RL 40 & ($\Delta$\passk{32} $= +7.0$ pp) & --- & --- & --- \\
    GT RL 60 & 43 / 5 ($1\times 10^{-8}$) & $0.58$ & $+16.6$ pp & 1,846 ($22.5\%$) \\
    GT RL 100 & 45 / 7 ($7\times 10^{-8}$) & $0.54$ & $+17.0$ pp & 2,260 ($27.4\%$) \\
    $\Eperp^{0\to 20}$ & ($\Delta$\passk{32} $= +2.3$ pp) & --- & --- & --- \\
    $\Eperp^{0\to 40}$ & 26 / 9 ($0.006$) & $1.44$ & $+6.6$ pp & --- \\
    $\Eperp^{0\to 60}$ & 25 / 16 ($0.21$) & $1.39$ & $+7.4$ pp & --- \\
    $\Eperp^{0\to 100}$ & 24 / 12 ($0.065$) & $1.19$ & $+10.2$ pp & --- \\
    \bottomrule
  \end{tabular}
\end{table}

\clearpage
\section{Training Dynamics and Mixed-Precision Details (\S\ref{sec:dynamics})}
\label{app:c}

\subsection{Gradient Reproducibility and Optimizer State Across Checkpoints}
\label{app:c1}

\paragraph{Sampling, directional Fisher information, and cross-term denoising protocol (\S\ref{sec:step20}, Figure~\ref{fig:dynamics}).}
We evaluate step~0 and steps 10, 20, 30, 40, 60, 80, and 100 across all 3 seeds (22 checkpoints total). At each checkpoint, we sample 16 rollouts per prompt on the first 512 dev problems (\nothink{}, 4,096-token cap). Directional Fisher information along $u_0$ is $u_0^\top F_t u_0$. We estimate the largest Ritz value $\lambda_1$ with 10 Lanczos steps initialized at $u_0$. We greedily partition the 512 problems into 8 disjoint subsets of 64 problems balanced by token count. Between steps 10 and 20, same-index subset overlaps are 58, 49, and 62 problems across seeds; within-checkpoint $t$-tests and leave-one-subset jackknife SEs remain exact. Writing $g = \sum_{i=1}^8 g^{(i)}$, the norm decomposition $\|g_R\|_2^2 = \sum_i \|g_R^{(i)}\|_2^2 + \sum_{i\ne j}\langle g_R^{(i)}, g_R^{(j)}\rangle$ yields the cross-subset reproducible energy fraction. For any subspace or unit direction $S$ (e.g., $u_0$ or $\Dsub$), we write $\eta_S(g) = \|P_S g\|_2^2 / \|g\|_2^2$ for the fraction of gradient energy lying in $S$. Denoised cosines and energy fractions replace squared norms with cross-subset sums and are reported when reproducible energy significantly exceeds 0. Finally, $u_0^\top g_R$ $t$-statistics use the 8-subset variance, where $|t|>2.4$ corresponds to two-sided $p<0.05$ at $\mathrm{df}=7$.

\begin{table}[htbp]
  \centering
  \caption{Shared reward gradients weaken from step~10 to step~20 while directional Fisher information $u_0^\top F_t u_0$ and 4k truncation rise across all three seeds ($512 \times 16$, 4,096-token cap; \S\ref{sec:step20}, Figure~\ref{fig:dynamics}b). All arrow columns report step~$10\to 20$ transitions. At step~0: reproducible energy $16.9\%$ (cross-subset numerator $58.7$, $\|g_R\|_2^2 = 346.9$), 4k truncation $0.07\%$, $u_0^\top F_0 u_0 = 1.98\times 10^5$, $\eta_{u_0}(g_L)=46.5\%$. The Cross Num.\ column reports the unnormalized cross-subset sum $\sum_{i \ne j}\langle g_R^{(i)}, g_R^{(j)}\rangle$ ($^\dagger$approximate from reported $\widehat{s}_R$ and $\|g_R\|_2^2$), showing that $\widehat{s}_R$ falls because the shared numerator drops sharply, not merely because $\|g_R\|_2^2$ grows $1.55\text{--}1.83\times$.}
  \label{tab:c1}
  \scriptsize
  \setlength{\tabcolsep}{3.2pt}
  \renewcommand{\arraystretch}{1.15}
  \resizebox{\textwidth}{!}{%
  \begin{tabular}{@{}cccccccc@{}}
    \toprule
    Seed & $\widehat{s}_R$ & Cross Num. & 4k Trunc & S10 $u_0^\top g_R$ ($t$) & S20 $u_0^\top g_R$ (rel.; $t$) & $u_0^\top F_t u_0$ ($\times 10^5$) & $\eta_{u_0}(g_L)$ \\
    \midrule
    A & $20.7\% \to 1.5\%$ & $120.5 \to 15.9$ & $3.9\% \to 22.4\%$ & $+4.5$ & $-0.10$ ($-0.4$) & $2.53 \to 3.59$ & $40.0\% \to 20.4\%$ \\
    B & $11.1\% \to 3.6\%$ & $63.6 \to 34.1^\dagger$ & $3.4\% \to 15.5\%$ & $+4.9$ & $+0.01$ ($0.0$) & $3.02 \to 4.27$ & $38.5\% \to 21.2\%$ \\
    C & $22.0\% \to 1.9\%$ & $131.6 \to 18.0$ & $2.1\% \to 15.7\%$ & $+4.2$ & $+0.50$ ($+2.2$) & $2.61 \to 3.81$ & $35.5\% \to 18.8\%$ \\
    \bottomrule
  \end{tabular}}
\end{table}

\begin{table}[htbp]
  \centering
  \caption{Seed~A unconstrained GRPO trajectory across steps 0--100 (Figures~\ref{fig:overview}c, \ref{fig:dynamics}a), showing that mean length and $u_0$ displacement largely stabilize after step~20--30 while dev accuracy and corrupted-prefix recovery ($h^-$) keep rising through step~100.}
  \label{tab:c2}
  \footnotesize
  \resizebox{\textwidth}{!}{%
  \begin{tabular}{@{}lcccccccc@{}}
    \toprule
    Step & 0 (Base) & 10 & 20 & 30 & 40 & 60 & 80 & 100 \\
    \midrule
    Dev accuracy & $45.5\%$ & $53.4\%$ & $66.6\%$ & $69.0\%$ & $70.2\%$ & $71.9\%$ & $72.5\%$ & $74.5\%$ \\
    Mean length (tok) & 585 & 991 & 2,449 & 2,889 & 2,429 & 2,532 & 2,810 & 2,505 \\
    BF16 $u_0$ dose & $0.00$ & $0.24$ & $0.69$ & $0.92$ & $1.00$ & $1.12$ & $1.13$ & $1.15$ \\
    Clean / corrupt ($h^+/h^-$) & $39.4/8.8\%$ & $43.9/15.5\%$ & $49.5/25.7\%$ & $50.8/29.0\%$ & $50.4/30.0\%$ & $53.1/33.2\%$ & $55.4/39.9\%$ & $57.5/46.6\%$ \\
    \bottomrule
  \end{tabular}}
\end{table}

\begin{table}[htbp]
  \centering
  \caption{Three-seed unconstrained GRPO comparison across steps 0--100, confirming across all seeds that length and $u_0$ displacement plateau early while recovery $h^-$ continues climbing through step~100. Step-100 $h^-$ uses the unified 3-seed trajectory audit (Table~\ref{tab:b11b}); in the independent paired evaluations of Tables~\ref{tab:b9} and \ref{tab:b12}, Seeds~A and C at step~100 reach $h^-=48.7\%$ and $53.8\%$, and Seed~A reaches $75.1\%$ dev accuracy. At step~20, dev accuracies across Seeds~A, B, and C are $66.6\%$, $63.5\%$, and $65.8\%$.}
  \label{tab:c3}
  \footnotesize
  \begin{tabular}{@{}cccc@{}}
    \toprule
    Step & Mean Len A / B / C (tok) & Recovery $h^-$ A / B / C & BF16 $u_0$ Dose A / B / C \\
    \midrule
    0 & 585 & $8.8\%$ & $0.00$ \\
    10 & 991 / 1,044 / 858 & $15.5\% / 15.5\% / 14.6\%$ & $0.24 / 0.19 / 0.14$ \\
    20 & 2,449 / 1,981 / 1,901 & $25.7\% / 24.5\% / 25.1\%$ & $0.69 / 0.46 / 0.44$ \\
    30 & 2,889 / 2,931 / 2,325 & $29.0\% / 30.3\% / 33.9\%$ & $0.92 / 0.71 / 0.61$ \\
    40 & 2,429 / 2,948 / 2,287 & $30.0\% / 33.6\% / 40.9\%$ & $1.00 / 0.79 / 0.67$ \\
    60 & 2,532 / 2,659 / 3,046 & $33.2\% / 42.2\% / 50.4\%$ & $1.12 / 0.86 / 0.87$ \\
    80 & 2,810 / 2,366 / 2,097 & $39.9\% / 45.0\% / 52.3\%$ & $1.13 / 0.86 / 0.82$ \\
    100 & 2,505 / 2,057 / 1,636 & $46.6\% / 42.9\% / 52.6\%$ & $1.15 / 0.76 / 0.88$ \\
    \bottomrule
  \end{tabular}
\end{table}

\begin{table}[htbp]
  \centering
  \caption{Directional Fisher information ($u_0^\top F_t u_0$), normalized ratio $(u_0^\top F_t u_0)/\lambda_{1,t}$ (where $\lambda_{1,t}$ is the largest Ritz value at step~$t$), 4k-cap truncation rate, and gradient direction measurements across all 22 checkpoints (first 512 dev $\times 16$, 4,096-token cap). The length gradient $g_L$ retains $43\%\text{--}73\%$ cross-subset reproducible energy across checkpoints. In the $u_0^\top g_R$ column, each seed block is normalized to its own step~10; the shared Step~0 (Base) row is normalized to Seed~C step~10 ($0.32$; relative to Seed~A step~10, it equals $0.44$ as plotted in Figure~\ref{fig:dynamics}b). Between steps 10 and 20, the cross-subset numerator of $\widehat{s}_R$ falls sharply across all three seeds (Table~\ref{tab:c1}) alongside rising 4k truncation ($2.1\%\text{--}3.9\% \to 15.5\%\text{--}22.4\%$).}
  \label{tab:c4}
  \scriptsize
  \setlength{\tabcolsep}{3pt}
  \resizebox{\textwidth}{!}{%
  \begin{tabular}{@{}lcccccccc@{}}
    \toprule
    Checkpoint & $u_0^\top F_t u_0$ ($\times 10^5$) & $(u_0^\top F_t u_0)/\lambda_{1,t}$ & 4k Trunc & $g_R$ Reprod. & $u_0^\top g_R$ (rel.\ 10; $t$) & $\eta_{\Dsub}(g_R)$ & $\cos(g_R,g_L)$ & $\eta_{u_0}(g_L)$ \\
    \midrule
    Step 0 (Base) & $1.98$ & $0.93$ & $0.1\%$ & $16.9\%$ & $0.32$ ($1.9$) & $6.9\%$ & $+0.14$ & $46.5\%$ \\
    Step 10 (Seed C) & $2.61$ & $0.86$ & $2.1\%$ & $\mathbf{22.0\%}$ & $1.00$ ($4.2$) & $22.8\%$ & $+0.28$ & $35.5\%$ \\
    Step 20 (Seed C) & $3.81$ & $0.73$ & $15.7\%$ & $1.9\%$ & $0.50$ ($2.2$) & $2.5\%$ & $-0.09$ & $18.8\%$ \\
    Step 30 (Seed C) & $4.75$ & $0.69$ & $22.1\%$ & $0.6\%$ & $0.57$ ($3.8$) & $2.9\%$ & $-0.16$ & $15.9\%$ \\
    Step 40 (Seed C) & $5.04$ & $0.70$ & $19.8\%$ & $-1.2\%$ & $-0.12$ ($-0.6$) & $1.9\%$ & $-0.16$ & $15.5\%$ \\
    Step 60 (Seed C) & $6.34$ & $0.58$ & $27.3\%$ & $8.6\%$ & $-0.69$ ($-2.5$) & $5.3\%$ & $-0.32$ & $18.9\%$ \\
    Step 80 (Seed C) & $4.69$ & $0.52$ & $15.4\%$ & $1.4\%$ & $0.34$ ($0.9$) & $1.9\%$ & $-0.00$ & $17.1\%$ \\
    Step 100 (Seed C) & $4.28$ & $0.55$ & $9.2\%$ & $1.1\%$ & $0.48$ ($1.6$) & $1.9\%$ & $+0.08$ & $21.0\%$ \\
    Step 10 (Seed A) & $2.53$ & $0.86$ & $3.9\%$ & $\mathbf{20.7\%}$ & $1.00$ ($4.5$) & $12.1\%$ & $+0.22$ & $40.0\%$ \\
    Step 20 (Seed A) & $3.59$ & $0.68$ & $22.4\%$ & $1.5\%$ & $-0.10$ ($-0.4$) & $0.8\%$ & $-0.25$ & $20.4\%$ \\
    Step 30 (Seed A) & $3.87$ & $0.56$ & $27.4\%$ & $2.4\%$ & $-0.50$ ($-1.5$) & $1.1\%$ & $-0.37$ & $15.6\%$ \\
    Step 40 (Seed A) & $3.51$ & $0.51$ & $21.8\%$ & $-2.0\%$ & $0.22$ ($0.6$) & $0.8\%$ & $-0.24$ & $15.8\%$ \\
    Step 60 (Seed A) & $3.95$ & $0.46$ & $22.2\%$ & $-2.2\%$ & $-0.02$ ($-0.1$) & $0.3\%$ & $-0.20$ & $16.9\%$ \\
    Step 80 (Seed A) & $4.71$ & $0.45$ & $25.0\%$ & $2.8\%$ & $0.22$ ($0.7$) & $0.9\%$ & $-0.12$ & $10.1\%$ \\
    Step 100 (Seed A) & $4.71$ & $0.46$ & $20.7\%$ & $3.2\%$ & $-0.57$ ($-1.0$) & $1.5\%$ & $-0.22$ & $12.5\%$ \\
    Step 10 (Seed B) & $3.02$ & $0.80$ & $3.4\%$ & $\mathbf{11.1\%}$ & $1.00$ ($4.9$) & $12.9\%$ & $+0.14$ & $38.5\%$ \\
    Step 20 (Seed B) & $4.27$ & $0.62$ & $15.5\%$ & $3.6\%$ & $0.01$ ($0.0$) & $1.9\%$ & $-0.10$ & $21.2\%$ \\
    Step 30 (Seed B) & $5.45$ & $0.61$ & $27.2\%$ & $11.2\%$ & $-0.83$ ($-1.9$) & $2.6\%$ & $-0.43$ & $21.3\%$ \\
    Step 40 (Seed B) & $5.17$ & $0.58$ & $26.3\%$ & $2.8\%$ & $-0.58$ ($-1.3$) & $2.1\%$ & $-0.33$ & $20.2\%$ \\
    Step 60 (Seed B) & $5.31$ & $0.58$ & $22.9\%$ & $10.2\%$ & $-0.30$ ($-0.9$) & $5.4\%$ & $-0.25$ & $19.5\%$ \\
    Step 80 (Seed B) & $4.85$ & $0.57$ & $19.5\%$ & $1.9\%$ & $-0.66$ ($-2.0$) & $1.4\%$ & $-0.32$ & $16.5\%$ \\
    Step 100 (Seed B) & $4.68$ & $0.61$ & $16.3\%$ & $7.6\%$ & $0.34$ ($0.9$) & $2.1\%$ & $-0.29$ & $10.0\%$ \\
    \bottomrule
  \end{tabular}}
\end{table}

\begin{table}[htbp]
  \centering
  \caption{Denoised gradient statistics (with jackknife SEs) at high-reproducibility checkpoints, showing that systematic step-10 reward gradients align strongly with $+u_0$ and $g_L$ across all three seeds before flipping negative at step~60.}
  \label{tab:c5}
  \scriptsize
  \resizebox{\textwidth}{!}{%
  \begin{tabular}{@{}lccccc@{}}
    \toprule
    Checkpoint & Raw $\eta_{\Dsub}(g_R)$ & Raw $\cos(g_R,g_L)$ & Denoised $\eta_{\Dsub}(g_R)$ & Denoised $\cos(g_R,u_0)$ & Denoised $\cos(g_R,g_L)$ \\
    \midrule
    Step 0 (Base) & $6.9 \pm 2.4\%$ & $+0.14 \pm 0.06$ & $26 \pm 15\%$ & $+0.39 \pm 0.27$ & $+0.41 \pm 0.18$ \\
    Step 10 (Seed C) & $22.8 \pm 5.5\%$ & $+0.28 \pm 0.08$ & $\mathbf{82 \pm 9\%}$ & $\mathbf{+0.80 \pm 0.08}$ & $\mathbf{+0.81 \pm 0.12}$ \\
    Step 10 (Seed A) & $12.1 \pm 3.9\%$ & $+0.22 \pm 0.07$ & $\mathbf{45 \pm 10\%}$ & $\mathbf{+0.61 \pm 0.11}$ & $\mathbf{+0.67 \pm 0.10}$ \\
    Step 10 (Seed B) & $12.9 \pm 2.5\%$ & $+0.14 \pm 0.07$ & $\mathbf{81 \pm 22\%}$ & $\mathbf{+0.82 \pm 0.26}$ & $\mathbf{+0.71 \pm 0.21}$ \\
    Step 60 (C / B) & $5.3\pm 2.1\%$ / $5.4\pm 2.8\%$ & $-0.32$ / $-0.25$ & $43\%$ / $35\%$ & $-0.53$ / $-0.15$ & $-0.85$ / $-0.31$ \\
    \bottomrule
  \end{tabular}}
\end{table}

\begin{table}[htbp]
  \centering
  \caption{AdamW states ($m$ and normalized update $d = \hat{m}/(\sqrt{\hat{v}}+\epsilon)$) reconstructed from FSDP checkpoints for Seed~C, together with FP32 displacement along $u_0$. Even though adaptive scaling reduces the single-step cosine alignment of $-d$ with $u_0$ to ${\sim}\tfrac{1}{5}$ that of $-m$, the cumulative FP32 projection along $u_0$ already reaches roughly $70\%$ of its step-100 magnitude by step~20.}
  \label{tab:c6}
  \footnotesize
  \begin{tabular}{@{}cccccc@{}}
    \toprule
    Step & $u_0^\top(\theta_t - \theta_0)$ (FP32) & Interval $\Delta$ & $\cos(-m, u_0)$ & $\cos(-d, u_0)$ & Dev $u_0^\top g_R$ (rel.\ 10) \\
    \midrule
    10 & $0.0090$ & $+0.0090$ & $+0.179$ & $+0.036$ & $+1.00$ \\
    20 & $0.0199$ & $+0.0110$ & $+0.214$ & $+0.044$ & $+0.50$ \\
    30 & $0.0245$ & $+0.0045$ & $+0.067$ & $+0.013$ & $+0.57$ \\
    40 & $0.0256$ & $+0.0011$ & $+0.032$ & $+0.009$ & $-0.12$ \\
    50 & $0.0279$ & $+0.0023$ & $+0.067$ & $+0.011$ & --- \\
    60 & $0.0304$ & $+0.0025$ & $+0.071$ & $+0.008$ & $-0.69$ \\
    70 & $0.0291$ & $-0.0013$ & $-0.101$ & $-0.011$ & --- \\
    80 & $0.0276$ & $-0.0015$ & $-0.022$ & $-0.004$ & $+0.34$ \\
    90 & $0.0267$ & $-0.0009$ & $+0.066$ & $+0.008$ & --- \\
    100 & $0.0283$ & $+0.0016$ & $+0.087$ & $+0.012$ & $+0.48$ \\
    \bottomrule
  \end{tabular}
\end{table}

\subsection{Phase~II Spectral Tail Audit and Periodic BF16 Master-Weight Sync Ablation}
\label{app:c2}

\paragraph{Three-seed full synchronization.}
We run the periodic BF16 master-weight synchronization ablation on three seeds over steps 21--60 (Table~\ref{tab:c7}). In the full-sync arms, we round all FP32 master weights onto the BF16 grid ($w \leftarrow Q(w)$) every 20 or 10 steps. Before resuming AdamW, we run up to 20 iterations of $\Dsub$ coordinate calibration to restore the window's executed $\Dsub$ displacement to within $10^{-4}$.

\paragraph{Seed-A masked synchronization on crossed vs.\ not-yet-crossed weights.}
To separate coordinates that have not yet crossed a BF16 boundary from previously crossed ones, we run two masked-sync arms on Seed~A from the same step-20 checkpoint, one for each coordinate group. Both arms synchronize every 10 steps without altering any current BF16 value or $\Dsub$ coordinate. Within each arm and at each sync step $t \in \{30, 40, 50, 60\}$, a per-step boolean mask tracks that arm's own cumulative crossing history since step~20: $\texttt{ever\_crossed}_t = \bigvee_{\tau=21}^{t} [Q(w_\tau) \neq Q(w_{20})]$. At each sync step, one arm snaps only the not-yet-crossed coordinates ($\neg\texttt{ever\_crossed}_t$) to $Q(w_t)$, while the other snaps only the already-crossed coordinates ($\texttt{ever\_crossed}_t$).

\begin{table}[!htbp]
  \centering
  \caption{Phase~II (steps 20--60) spectral tail audit, 3-seed full BF16 synchronization ablation, and Seed-A masked synchronization by per-step BF16 crossing history (\S\ref{sec:unified}, Figure~\ref{fig:dynamics}d, Table~\ref{tab:unified}). ``Blind'' pools the 512 held-out and 1,024 disjoint lockbox problems ($N=1{,}536$); ``Lock'' reports the 1,024-problem lockbox alone. Each masked synchronization left its own arm's BF16 execution weights unchanged at each sync step. On Seed~A, the not-yet-crossed-only drop was $89\%$ as large as the full-sync drop ($3.05$ vs.\ $3.44$ pp pooled).}
  \label{tab:c7}
  \scriptsize
  \setlength{\tabcolsep}{3.0pt}
  \renewcommand{\arraystretch}{1.08}
  \begin{tabularx}{\textwidth}{@{}>{\raggedright\arraybackslash}p{3.45cm}>{\raggedright\arraybackslash}p{4.95cm}>{\raggedright\arraybackslash}X@{}}
    \toprule
    Audit / Ablation & Key Measurements & Interpretation \\
    \midrule
    Parameter change fraction (steps 20--60, 2 seeds; sampled ckpts) & FP32: $99.93\% / 99.93\%$; net BF16 change: $\mathbf{18.65\% / 18.12\%}$ & ${\sim}2.0\%$ of weights differ at step~40 and later revert; $\mathbf{79.3\%}$ keep identical BF16 values across sampled checkpoints (steps 20, 40, 60). \\
    \addlinespace[2pt]
    Exact per-step crossing tracker (Seed A, steps 21--60, all 40 steps) & Ever crossed: $24.54\%$; net changed: $22.68\%$; reverted: $\mathbf{1.86\%}$ & Over all 40 steps, $\mathbf{75.46\%}$ of coordinates never cross a BF16 boundary. Under \texttt{uncrossed\_only} sync, only $14.69\%$ ever cross. \\
    \addlinespace[2pt]
    196 core matrices $r > 128$ tail (full 20$\to$60 disp.) & Cross-seed shared alignment: $\mathbf{60.52\%}$ in BF16 (energy $60.90\%$); $\mathbf{30.53\%}$ in FP32 & The high-rank tail accounts for a substantial share of cross-seed alignment at both precisions, and its share is larger in BF16 than in FP32. \\
    \addlinespace[2pt]
    Baseline: keep FP32 master weights (steps 21--60, 3 seeds) & Dev $72.00 / 72.61 / 72.58\%$; Lock $68.31 / 69.56 / 68.31\%$; Blind $\mathbf{69.81 / 70.69 / 69.29\%}$ & 3-seed means: dev $72.40\%$, lockbox $68.73\%$, pooled blind $69.93\%$. \\
    \addlinespace[2pt]
    Sync full FP32 master to BF16 every 20 steps (3 seeds) & Dev $70.80 / 72.44 / 70.43\%$; Lock $66.43 / 68.12 / 65.75\%$; Blind $\mathbf{67.77 / 69.62 / 67.74\%}$ & Blind $\mathbf{-1.55}$ \textbf{pp} ($t=-5.62, p=0.0303$); Lockbox $\mathbf{-1.96}$ \textbf{pp} ($t=-6.00, p=0.0266$). \\
    \addlinespace[2pt]
    Sync full FP32 master to BF16 every 10 steps (3 seeds) & Dev $68.48 / 69.26 / 69.90\%$; Lock $65.50 / 65.65 / 65.87\%$; Blind $\mathbf{66.37 / 66.63 / 67.89\%}$ & Blind $\mathbf{-2.97}$ \textbf{pp} ($p=0.0662$); Lockbox $\mathbf{-3.05}$ \textbf{pp} ($t=-6.93, p=0.0202$); dev $\mathbf{-3.18}$ \textbf{pp} ($p=0.0064$). \\
    \addlinespace[2pt]
    \textbf{Mask: sync not-yet-crossed only} (Seed A, ev.\ 10) & Dev $\mathbf{69.09\%}$; Hold $68.90\%$, Lock $65.70\%$; Blind $\mathbf{66.76\%}$ & Blind $\mathbf{-3.05}$ \textbf{pp} vs.\ Seed-A baseline ($69.81\%$); Lockbox $-2.61$ pp ($68.31\% \to 65.70\%$). \\
    \addlinespace[2pt]
    \textbf{Mask: sync crossed only} (Seed A, ev.\ 10; keeps not-yet-crossed) & Dev $\mathbf{72.80\%}$; Hold $71.92\%$, Lock $69.65\%$; Blind $\mathbf{70.41\%}$ & Blind $\mathbf{+0.60}$ \textbf{pp} vs.\ baseline. Paired \texttt{crossed}$-$\texttt{uncrossed}: Lockbox $\mathbf{+3.96\,[+2.64, +5.27]}$ \textbf{pp} ($t=5.92$, $N{=}1{,}024$); Blind $\mathbf{+3.65\,[+2.56, +4.75]}$ \textbf{pp} ($t=6.53$, $N{=}1{,}536$); dev $+3.71\,[+2.32, +5.10]$. \\
    \bottomrule
  \end{tabularx}
\end{table}

\begin{table}[!htbp]
  \centering
  \caption{Unified mixed-precision geometry of concentrated early shifts (\S\ref{sec:deadzone}) and diffuse late updates (\S\ref{sec:unified}).}
  \label{tab:unified}
  \scriptsize
  \renewcommand{\arraystretch}{1.08}
  \begin{tabularx}{\textwidth}{@{}>{\raggedright\arraybackslash}p{2.6cm}>{\raggedright\arraybackslash}X>{\raggedright\arraybackslash}X@{}}
    \toprule
    Dimension & Concentrated Leading Fisher ($\CD$, prominent in Phase~I) & Diffuse Orthogonal ($\Eperp$, prominent in Phase~II) \\
    \midrule
    Support \& spectrum & Half of $u_0$ energy on ${\sim}10^{-5}$ weights & SVD cross-seed alignment tail beyond rank $r>128$ (full steps 20--60 disp.): $60.52\%$ (BF16) / $30.53\%$ (FP32) \\
    BF16 grid behavior & Concentrated subtraction largely survives BF16 rounding ($\mL \to 1$) & $75.5\%$ never cross a grid boundary; $1.9\%$ cross and revert (Seed~A per-step audit) \\
    Engineering consequence & FP32-only locking leaves an opposite-sign projection after rounding (Figure~\ref{fig:deadzone}b) & Full sync lowers 3-seed mean accuracy by $1.55\text{--}2.97$ pp pooled ($1.96\text{--}3.05$ pp lockbox, $p \le 0.027$); on Seed~A, syncing only not-yet-crossed vs.\ already-crossed coordinates changes pooled blind accuracy by $-3.05$ pp vs.\ $+0.60$ pp (lockbox paired $+3.96$ pp) \\
    \bottomrule
  \end{tabularx}
\end{table}

\FloatBarrier
\section{Cross-Scale, Online Damping, and Cross-Family Replication Details}
\label{app:d}

\paragraph{Online $\Dsub$ gradient damping.}
We extend step-0 $\Dsub$ damping, $\tilde{g} = g - (1-\gamma)P_{\Dsub} g$, to 40-step online GRPO (Table~\ref{tab:d2}). Mild damping ($\gamma = 0.25$) saves length modestly ($2{,}258$ vs.\ $2{,}448$ dev tokens) at limited accuracy cost ($67.3\%$ vs.\ $69.3\%$ on dev; $68.6\%$ vs.\ $70.9\%$ on held-out). Stronger damping ($\gamma = 0.10$) cuts dev length by $20\%\text{--}50\%$ ($1{,}274\text{--}1{,}974$ tokens) at a cost of $6\text{--}11$ pp in accuracy, indicating that strongly suppressing $\Dsub$ forfeits early mode-release gains.

\paragraph{Equal-LR cross-family controls.}
To rule out learning-rate mismatch as the reason native \texttt{Llama-3.2-1B-Instruct} plateaus early, we compare both families at matched learning rates ($\mathrm{lr}=2\times 10^{-6}$ and $4\times 10^{-6}$, Table~\ref{tab:d2}). Even at $\mathrm{lr}=4\times 10^{-6}$, native Llama plateaus near $30\%$ across steps 20--40 (and across steps 20--60 at $2\times 10^{-6}$), whereas \texttt{Qwen3-0.6B} climbs to $62\%\text{--}71\%$ by step~40, pointing back to the absence of a latent long-CoT Fisher mode rather than step size.

\paragraph{Cross-scale and cross-family BF16 splits.}
To test how the balance between mode release and orthogonal refinement shifts when template suppression is weak, we evaluate \texttt{Qwen3-8B} and template-free \texttt{Qwen2.5-1.5B} (Table~\ref{tab:d1}). Template-free \texttt{Qwen2.5-1.5B} keeps held-out length nearly flat ($568 \to 550\text{--}574$ tokens, $\eta_{\Dsub}(g_L)=7.22\%$), and \texttt{Qwen3-8B} lengthens only mildly ($792 \to 885$ tokens); in both models, $\Eperp$ accounts for a larger share of total accuracy gain than $\CD$, because weaker template suppression and lower $\Dsub$ length sensitivity leave limited room for length-mode arbitrage.

\begin{table}[htbp]
  \centering
  \caption{Half-ULP audit, BF16 $2 \times 2$ split, and 8-subset gradient decomposition on \texttt{Qwen3-8B} (1 seed) and \texttt{Qwen2.5-1.5B} (2 seeds; MATH, \S\ref{sec:net-gain}). On \texttt{Qwen3-8B}, $CE$ ($83.89\%$) sits $1.36$ pp below GRPO-60 ($85.25\%$) on held-out and $0.17$ pp below on dev; the held-out gap is about $1.2$ times the estimated sampling standard error ($\mathrm{SE}\approx 1.15$ pp on $512\times 4$).}
  \label{tab:d1}
  \scriptsize
  \setlength{\tabcolsep}{3.5pt}
  \renewcommand{\arraystretch}{1.16}
  \begin{tabularx}{\textwidth}{@{}>{\raggedright\arraybackslash}p{2.65cm}>{\centering\arraybackslash}p{2.55cm}>{\centering\arraybackslash}p{2.55cm}>{\raggedright\arraybackslash}X@{}}
    \toprule
    Model / Config & Dev 1,024 (Len) & Held-out 512 (Len) & Half-ULP Audit / 8-Subset Gradient Metrics \\
    \midrule
    \texttt{Qwen3-8B} base & $81.81\%$ (824) & $82.03\%$ (792) & Step 0: $\eta_{\Dsub}(g_L)=37.89\%$, $\cos(g_R^\perp,g_L^\perp)=-0.038$ \\
    \texttt{Qwen3-8B} GRPO 40 / 60 & $84.42 / 85.72\%$ (889/927) & $83.69 / 85.25\%$ (854/885) & Steps 20/40: $\eta_{\Dsub}(g_L)=33.58/41.57\%$, $\cos=-0.067$ / $-0.134$ \\
    \texttt{Qwen3-8B} FP32-only zero & $82.54\%$ (717) & $81.59\%$ (711) & Active $6.72\%$, $f=0.152\text{--}0.162$, $\mL=0.731\text{--}0.782$ ($-0.60\times$ FP32; $-2.71\times\cexec$) \\
    \texttt{Qwen3-8B} BF16 $C/\Eperp/CE$ & $83.52 / 85.42 / 85.55\%$ & $83.30 / 83.79 / 83.89\%$ & BF16 placement meets target coordinates within tolerance; avoids reversed offset \\
    \midrule
    \texttt{Qwen2.5-1.5B} base & $46.61\%$ (576) & $46.58\%$ (568) & Step 0: $\eta_{\Dsub}(g_L)=7.22\%$, $\cos(g_R^\perp,g_L^\perp)=-0.044$ \\
    \texttt{Qwen2.5-1.5B} GRPO 40 (A/B) & $52.05 / 51.81\%$ (577/561) & $51.46 / 51.46\%$ (574/550) & Steps 10/20/40: $\eta_{\Dsub}(g_L)=8.20/5.92/4.03\%$, $\cos=+.004$ / $-.059$ / $-.129$ \\
    \texttt{Qwen2.5-1.5B} FP32-only (A) & $43.85\%$ (587) & $42.48\%$ (601) & Drops below base ($-2.76$ / $-4.10$ pp) \\
    \texttt{Qwen2.5-1.5B} BF16 $C/\Eperp/CE$ (A) & $49.00 / 50.17 / 51.88\%$ & $48.93 / 49.85 / 51.37\%$ & Online locked $\Dsub$ (A): dev $50.83\%$, held-out $48.58\%$ \\
    \texttt{Qwen2.5-1.5B} BF16 $C/\Eperp/CE$ (B) & $47.88 / 50.76 / 51.42\%$ & $47.90 / 50.98 / 51.76\%$ & Online locked $\Dsub$ (B): dev $51.64\%$, held-out $50.88\%$ \\
    \bottomrule
  \end{tabularx}
\end{table}

\begin{table}[htbp]
  \centering
  \caption{Online $\Dsub$ gradient damping (\S\ref{sec:selection-response}) and equal-LR cross-family controls (\S\ref{sec:puzzles}).}
  \label{tab:d2}
  \scriptsize
  \setlength{\tabcolsep}{3.5pt}
  \resizebox{\textwidth}{!}{%
  \begin{tabular}{@{}lccl@{}}
    \toprule
    Configuration & Dev 1,024 (Len) & Held-out 512 (Len) & Auxiliary Metrics \\
    \midrule
    Online damped $\Dsub$: $\gamma=0.10$, step 40 (A/B) & $60.08 / 63.11\%$ (1,274/1,974) & $59.42 / 63.18\%$ (1,353/2,087) & Trunc $0.71/3.20\%$; $h^+=49.22/51.17\%$ \\
    Online damped $\Dsub$: $\gamma=0.25$, step 40 (A) & $67.29\%$ (2,258) & $68.55\%$ (2,358) & Trunc $8.96\%$; $h^+=53.24\%$ \\
    Equal-LR \texttt{Llama-3.2-1B}: $\mathrm{lr}=2\text{e-}6$ (20/40/60) & $30.13 / 30.49 / 30.79\%$ & $30.13 / 28.66 / 29.25\%$ & Base $11.65\%$; $h^+=25.00/25.78/24.35\%$ \\
    Equal-LR \texttt{Llama-3.2-1B}: $\mathrm{lr}=4\text{e-}6$ (20/40) & $30.32 / 30.25\%$ (515/507) & $27.64 / 30.22\%$ (489/500) & $h^+=24.74/27.20\%$ \\
    Equal-LR \texttt{Qwen3-0.6B}: $\mathrm{lr}=2\text{e-}6$, step 40 (A/B) & $62.45 / 63.16\%$ (1,750/1,888) & $62.45 / 61.77\%$ (1,763/1,909) & Base $45.58\%$; $h^+=48.96/48.45\%$ \\
    Equal-LR \texttt{Qwen3-0.6B}: $\mathrm{lr}=4\text{e-}6$, step 40 (A) & $69.31\%$ (2,448) & $70.85\%$ (2,473) & $h^+=52.33\%$ \\
    \bottomrule
  \end{tabular}}
\end{table}

\end{document}